%% file: main.tex
\pdfoutput=1
\documentclass[11pt]{article}
\usepackage[final]{acl} %preprint final review
\usepackage{times}
\usepackage{latexsym}

\usepackage{hyperref}
\usepackage{url}
\usepackage{booktabs}
\usepackage{graphicx}
\usepackage{subcaption} % For subfigures
\usepackage{pdfpages} 
\usepackage{verbatim}
\usepackage{kotex} %korean
\usepackage{tabularray}
\usepackage{stfloats}

\usepackage[table]{xcolor}
\usepackage{contour}
\contourlength{0.1pt} 
\usepackage{colortbl}
\definecolor{OISkyBlue}{RGB}{86,180,233}
\definecolor{OIBlue}{RGB}{0,114,178}
\definecolor{OIBluishGreen}{RGB}{0,158,115}
\definecolor{OIOrange}{RGB}{230,159,0}
\definecolor{OIYellow}{RGB}{240,228,66}
\definecolor{OIVermillion}{RGB}{213,94,0}
\definecolor{OIReddishPurple}{RGB}{204,121,167}
\definecolor{OIGrey}{gray}{0.85}
\definecolor{darkblue}{rgb}{0, 0, 0.5}
\hypersetup{colorlinks=true, citecolor=darkblue, linkcolor=darkblue, urlcolor=darkblue}
\usepackage{adjustbox}
\usepackage{caption}
\usepackage{tikz}

\usepackage{graphicx}
\definecolor{OliveGreen}{rgb}{0,0.6,0}

\usepackage{linguex}

\title{The Grammar of Transformers:\\A Systematic Review of Interpretability Research on Syntactic Knowledge in Language Models} 

\author{Nora Graichen$^1$, 
  Iria de-Dios-Flores$^{1*}$, %\thanks{Equal senior authorship.},
  Gemma Boleda$^{1,2*}$ \\ %\textsuperscript{\thefootnote}
  Universitat Pompeu Fabra$^1$, ICREA$^2$\\
  {\small\texttt{\{firstname.lastname\}@upf.edu}
  }
}
\begin{document}
%\linenumbers
\maketitle
\vspace{-1em}
\begin{center}

\end{center}

\begin{abstract}
We present a systematic review of 337 articles evaluating the syntactic abilities of Transformer-based language models (TLMs), reporting on over 3,000 datapoints spanning a wide range of syntactic phenomena, languages, models, and methods. We take the data to collectively show that TLMs encode a non-trivial amount of syntactic knowledge. Behavioral evidence shows strong performance on formal syntactic phenomena, but weaker and more variable performance on phenomena at the syntax-semantics interface. Performance is also consistently lower for languages with less digital support. Probing and mechanistic studies further support the presence of syntactic knowledge in TLMs. Yet, because most work remains observational and methodologically heterogeneous, insight into the detailed computational mechanisms underlying syntactic processing remains limited. At the same time, the literature remains heavily concentrated on English and BERT-like models. We discuss the implications of our results and provide recommendations for future research.

  %We present a systematic review of 337 articles evaluating the syntactic abilities of Transformer-based language models (TLMs), reporting on over 3,000 datapoints from a wide range of syntactic phenomena, languages, models, and methods. We take the data to collectively show that TLMs do encode a non-trivial amount of syntactic knowledge. The state of the art presents a healthy variety of methods and data, but an over-focus on a single language (English), a single model (BERT), observational (as opposed to interventional) methods, and phenomena that are easy to get at (like part of speech and agreement). Behavioral results also suggest that TLMs capture these form-oriented phenomena well, but show more variable and weaker performance on phenomena at the syntax-semantics interface, like binding or filler-gap dependencies. Performance is also worse for languages with less digital support. %%Overall, results suggest a  generally positive trend in mechanistic evidence. 
  % We provide a thorough discussion of the results and recommendations for the future.
  %%Moving forward, we recommend the field address the uncovered gaps, in particular 
  % and including better reporting practices and a move towards 
  %regarding phenomena, languages, and combining observational and experimental methods.
\end{abstract}

\section{Introduction}
\label{sec:intro}
%Transformer-based language models (\textbf{TLMs})are driving a global shift in AI. %, producing fluent text across diverse languages. 
%A growing body of research has examined how modern TLMs acquire, represent, and deploy linguistic knowledge \cite{rogers_primer_2020,zhou_linguistic_2025, brinkmann_large_2025,kryvosheieva_controlled_2025}; yet, their grasp of natural language syntax remains only partially understood \cite{agarwal:etal:2025}. Meanwhile, the very active field has developed along multiple directions. In response to this increasingly broad and fragmented literature, the present article provides a systematic review of research on syntactic knowledge in TLMs, with the aim of charting the current state of the art.

%Given the sizable literature on the topic, we conduct a \textbf{systematic review}, gathering relevant studies using transparent procedures for search and inclusion \citep{page:etal:2021} and providing quantitative insights.
\def\thefootnote{}\footnotetext{* Equal senior authorship.}\def\thefootnote{\arabic{footnote}}
It has been observed at least since \citet{Gulordava:etal:2018} that language models (LMs), though trained simply on a token prediction task, seem to acquire a non-trivial amount of syntactic knowledge.
In recent years, a growing body of research has focused on understanding what precisely is learned and how it is used at processing time \cite[e.g.,][]{rogers_primer_2020,lasri_does_2022, brinkmann_large_2025}.
However, even if some trends have emerged, there is yet no general picture about syntactic knowledge in LMs.
We therefore think that it is a good moment to stop and appraise the extant literature.
In this article, we do precisely that through a \textbf{systematic review} \citep{page:etal:2021}, a type of review that (to our knowledge) is not commonly used in computational linguistics (CL) and NLP. %\footnote{Computational Linguistics and Natural Language Processing.}
Systematic reviews differ from common narrative reviews in that they include transparent search, selection, and annotation criteria, and they provide quantitative evidence for the identified patterns.
We focus on Transformer-based language models (TLMs) because they are the dominant model as of 2026.

The specific goals for this review are twofold: (i) mapping the landscape of research on syntactic knowledge in TLMs, including what has been done and how, as well as what remains to be done (Section~\ref{sec:State_of_the_field}); and (ii) summarizing what is currently known about syntactic knowledge in TLMs (Section~\ref{sec:Syntactic_Knowledge_TLMs}).

Our work partially builds on six previous narrative reviews with overlapping content: %\id{narrative?}
\citet{Limisiewicz:Maracek:2020}, \citet{Linzen:Baroni:2020}, \citet{Kulmizev:Nivre:2021}, \citet{chang:bergen:2023}, \citet{Milliere:2024}, and \citet{lopezotal:etal:2025}.
These works found that modern LMs acquire %rich and hierarchical syntactic knowledge 
rich syntactic representations early during pretraining \citep{chang:bergen:2023}, 
localize much of this information in their middle layers, and encode at least some syntactic relations via specialized attention heads and groups of neurons \citep{lopezotal:etal:2025}.
The reviews also point out limitations of TLMs, including reliance on surface-level heuristics, lexical sensitivity, and challenges with rare constructions and subtle distinctions \citep{Kulmizev:Nivre:2021,Linzen:Baroni:2020}.
While valuable, these prior reviews are largely narrative and qualitative. In contrast, we provide a quantitative and systematic synthesis of the literature with the aim of offering a transparent mapping between the available evidence and the conclusions, an approach that is increasingly necessary given the scale and methodological heterogeneity of the knowledge accumulated in recent years.
%In our review, \id{Start here:}
Our review covers 337 %relevant 
articles and 3,074 individual datapoints across different languages, syntactic phenomena, methods and datasets. % 1015+ 134 average BliMP scores + 368 BliMP per-category scores  + non-benchmark 1557 scores
% \n{type token number of models (54/1015), syntactic phenomena(11/585).!579}
We also contribute to the community an annotated database of the reviewed studies, which is publicly available together with the analysis code at \url{https://github.com/norgrai/syntax_review}.

\section{Method}
\label{sec:method}

% add for 9 pages? A systematic review gathers and synthesizes relevant studies, providing a transparent content-to-conclusion mapping \citep{page:etal:2021}. Unlike narrative reviews, to reduce bias, it follows explicit procedures for article search, inclusion, and processing; unlike meta-analyses or meta-reviews, it does not build a statistical model with the different variables, but still synthesizes results quantitatively (and may do so qualitatively when outcome measures are not directly comparable). 

\begin{figure}[h] %htp
  \centering
  \includegraphics[trim=4.7cm 0.1cm 1cm 0.5cm, % left bottom right top width=0.7\linewidth
  clip, scale=0.39]{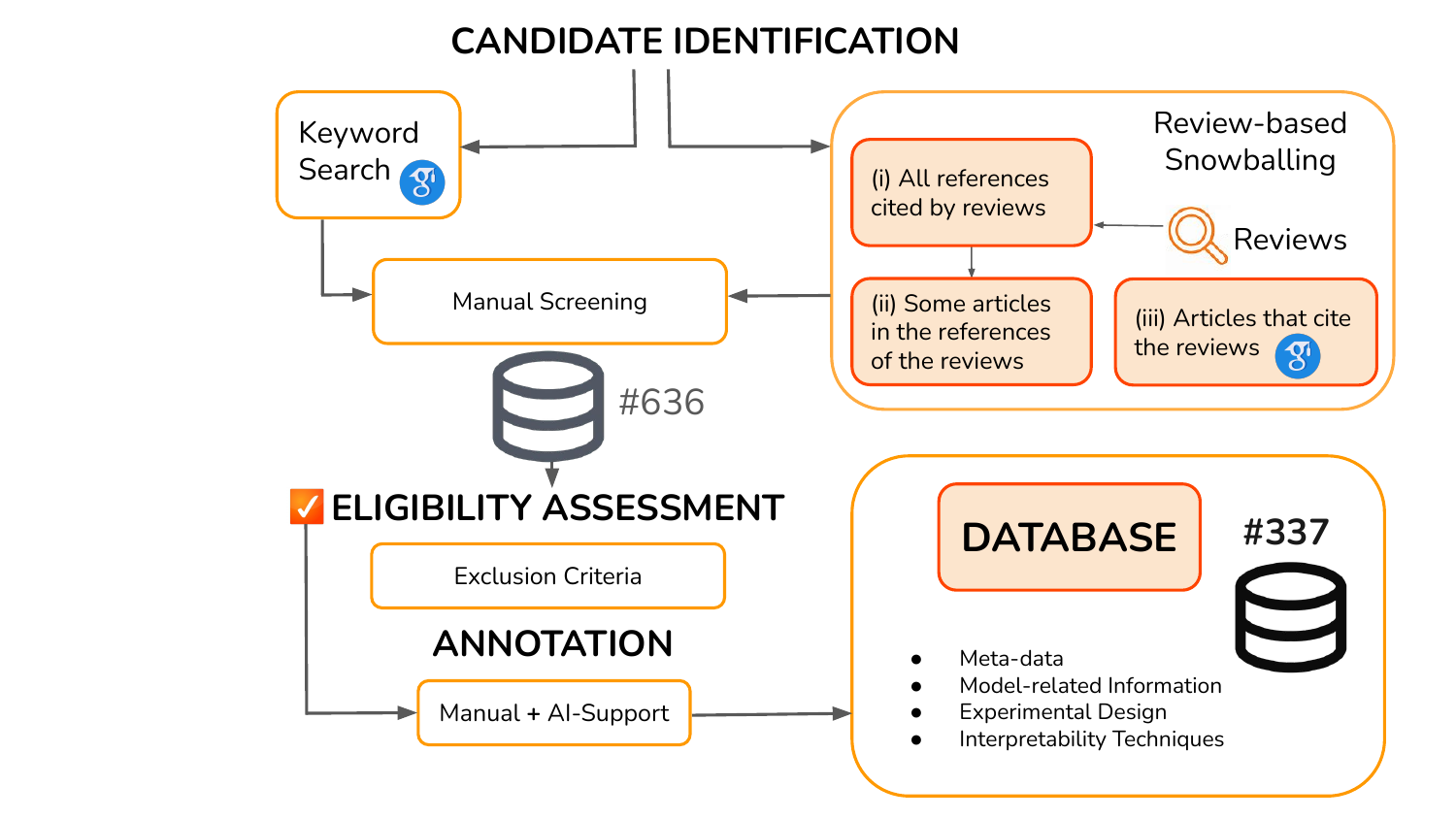}%[width=8cm]{} width=\columnwidth,
  \caption{Flowchart summarizing database creation.}
  \label{fig:source_review}
\end{figure}

Fig.~\ref{fig:source_review} summarizes the process we followed to create the database, comprising paper candidate identification, eligibility assessment, and annotation. We did data collection in summer 2025, with a cut-off date for inclusion of July 31, 2025.
%We did data collection in summer 2025 and set the cut-off date for inclusion to July 31, 2025. %\id{Here we could also somehow note that the idea is that the database is updated, the forms for paper inclusion...etc.}
Importantly, the database is designed to remain open to future extensions allowing researchers to contribute new papers through a dedicated form.\footnote{\url{https://forms.gle/sbdEYkzcHziqevcr7}}
% Data collection started in March 2025 %G: I know I asked for this, but I find it confusing and on a 2nd thought not necessary

\paragraph{Paper candidate identification.}
In the initial identification stage, we prioritized recall, deferring concerns about precision to the subsequent stage. 
Article relevance was primarily assessed based on the title; in case of doubt, we reviewed the abstract, and, if necessary, the full text.
To identify candidates, we followed a two-pronged strategy: A \textit{keyword search} in Google Scholar, using keywords that combined ``syntactic structure/knowledge'' and ``LLMs/LMs/transformer'', and a \textit{snowballing strategy} using the six aforementioned reviews on the topic to identify relevant references. %\cite{lopezotal:etal:2025, Milliere:2024, chang:bergen:2023, Kulmizev:Nivre:2021, Linzen:Baroni:2020, Limisiewicz:Maracek:2020}.
The search was conducted in English; reference snowballing was not restricted by publication language. 
In the latter, we screened (i)~all the 263 references cited by the reviews,%
\footnote{With the exception of \citet{lopezotal:etal:2025}, for which we considered only the studies listed in its dedicated section on syntax-related research in TLMs.} (ii)~some %\id{Why some?} 
of the articles that were cited in these 263 articles (pursued until further screening yielded few new relevant studies), and (iii) most of the articles that cite the reviews themselves, identified via Google Scholar.%
% \citep{chang:bergen:2023} Citat per 156 
% \citep{lopezotal:etal:2025} Citat per 2
% \citep{Milliere:2024} Citat per 32 
% \citep{Linzen:Baroni:2020} Citat per 305 
% \citep{Kulmizev:Nivre:2021} Citat per 25 
% \citep{Limisiewicz:Maracek:2020} Citat per 15 
% Since the latter amount to over 500 citations (citation count as of August 2025) on Google Scholar. From these, 
\footnote{In particular, if a review had more than 100 citations, we reviewed the top 30-40\% (sorted by relevance as determined by Google Scholar), otherwise all citations were checked.}
This yielded 636 articles. %together with their respective sources.

\paragraph{Eligibility assessment.}
This stage shifted attention to precision. We excluded:
% with the added page in the final version, if possible transform the following into an enumerated list
1)~Studies focusing on non-transformer architectures (e.g., RNNs, LSTMs) or on encoder–decoder architectures, including those designed for machine translation, so as to keep articles within the scope of our research question.
%(see last paragraph of Section~\ref{sec:intro}).
2)~Studies that, while analysing TLMs, do not empirically assess syntactic knowledge.
3)~Publications without a clear empirical component, such as position papers or surveys (the previous reviews were not included in the analysis due to the absence of an empirical component).
4)~Publications that were neither peer-reviewed journal articles, conference papers, nor archived preprints. This implies excluding blog posts and MA or PhD theses, since they often report on findings that are also published in peer-reviewed venues.  The remaining 337 articles (cited in App.~\ref{app:references}, Table~\ref{tab:ref}) were further annotated and analysed.%
\footnote{Excluded articles and exclusion reasons are documented in the database to ensure transparency and enable future studies beyond the scope of this review.}

\begin{figure*}[!ht]
  \begin{minipage}[c]{0.35\textwidth}
  \includegraphics[trim=0.2cm 0.255cm 0.1cm 0.255cm, clip, scale=0.49]{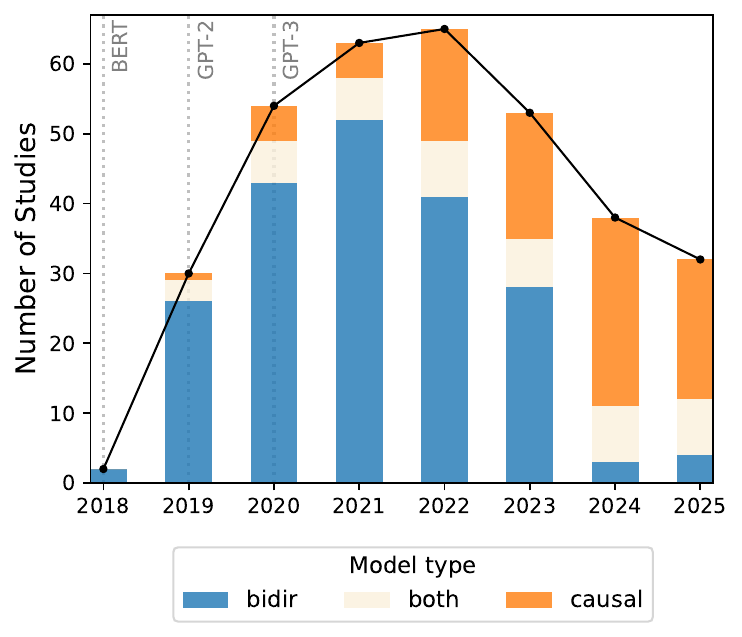}
  \end{minipage}
  \hspace{0.7cm}
  %\subfloat{\includegraphics[trim=0.1cm 0.25cm 0.1cm 0.21cm, clip, scale=0.5]{figures/model_by_family_vertical.pdf}}
  \begin{minipage}[c]{0.47\textwidth}
  \includegraphics[trim=0.25cm 0.9cm 21.9cm 2.6cm, clip, scale=0.5]{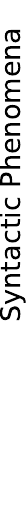}
  \includegraphics[trim=1.15cm 0.33cm 1.5cm 0.7cm, clip, scale=0.45]{figures/phenomena11_languages-1.pdf}
  \end{minipage}
  \label{fig:overview4}
  \caption{\textbf{Left:} Number of studies published each year; debut of influential models. \textbf{Right:} Counts of syntactic phenomena by  language tested (only English tested, English and other languages, and non-English).}
  \label{fig:overview}
\end{figure*}

\paragraph{Annotation.}
%The annotation focused on dimensions motivated by our research question, supporting the subsequent quantitative analyses and enabling comparison across studies.
%The annotation aimed at facilitating data analysis, ensuring transparency and enabling comparison across studies. \id{rephrasing suggestion: "The annotation focused on dimensions directly motivated by our research questions, supporting the subsequent quantitative analyses and enabling comparison across studies."}
%The resulting database comprises 33 %26 variables belonging to 5 different categories. \id{I would go straight to the point. E.g. 
We defined 32 annotation variables, organized into four categories, to systematically capture the aspects of the selected works that would enable quantitative analyses and cross-study comparisons.
%We provide an overview in what follows, and a detailed description is given in App.~\ref{Appendix:annotations}.
An overview is provided below, with a detailed description in App.~\ref{Appendix:annotations}.

%The \textit{meta-information} category comprises article meta-data such as year and URL, as well as how articles entered or exited the review process (source of each article and, if applicable, the grounds for exclusion).
The \textit{meta-data} category includes publication details and information on how articles entered or exited the review process (source of each article and, if applicable, the grounds for exclusion).
\textit{Model-related} information specifies the model name(s), type(s), fine-tuning status, and language coverage.
\textit{Experimental design} codes the syntactic phenomena investigated, languages tested, and experimental materials employed.
To establish the syntactic categories, we applied a bottom-up approach, starting from the terms used in the studies themselves and grouping them under broader labels recognizable to both the linguistics and NLP communities. The resulting 11 labels are shown in Fig.~\ref{fig:overview} (right) and described in detail in App.~\ref{Appendix:annotations}.
\textit{Interpretability techniques} include the specific method, as named in the original paper, and a general category \cite[behavioral, mechanistic, or probing, along the lines of][see next section for more information]{Milliere:2024}. Further, we annotated various dimensions for probing and mechanistic studies, covering methodological aspects, causal nature, locality and strength of evidence of syntactic awareness. 

We produced %all annotations manually except 
annotations manually for 28 of the 32 variables. For the remaining 4 (model type, syntactic phenomena, experimental materials, and method name) %, and main syntactic findings) 
we used a semi-automated AI-assisted workflow, which was sucessfully validated, to extract some information from the papers %against a manual gold standard 
(further details are provided in App.~\ref{app:ai-flow}). Final labels were determined manually. %for which we used a semi-automated AI workflow. \id{I think this phrasing still gives the impression that there was much less manual work than it really was. For instance, the syntactic phenomena was annotated mannually, even though the initial information extraction was done using the semi-automated AI workflow. ALternative phrasing: "for which we used a semi-auomated AI-assisted step/workflow to extract some information from the papers". Final labels were determined manually (not sure the second sentence is a bit misleading...)}

%\no{Additionally to the annotations, we created an AI-summary \textit{findings} that summarizes, in prose, the main insights regarding syntactic competence and processing as well as any model comparisons, which is available upon request.}

%Upon completion, the annotated information was automatically post-processed to create the final database, containing standardized variables. 
%\gb{More detailed than the 26? Clarify "more detailed" or remove and leave only "standardized"}
%\n{Python?} scripts 
%\id{commnet: maybe cite the library/package you used?} 
%\gb{No, too low-level for the main text. Do that in appendix.}

%\section{Results}
%\label{sec:results}
%We structure the results in two parts. First, we characterize the current state of the field and provide a quantitative view that confirms and refines some previously suspected trends. %a quantitative view, confirming and refining previously suspected trends. \id{a quantitative view that confirms and refines SOME previously suspected trends}
%showing that patterns previously only suspected are borne out empirically in our aggregated analysis. 
%Next, we turn to syntactic knowledge in TLMs, examining in detail how these models perform across linguistic phenomena thoroughly.

%\subsection{State of the field/ A Quantitative Overview of the Field}\label{sec:State_of_the_field} \no{also the notebook is structured like this}%\id{I don't love the title, even though I know it's the one we've been using in our conversations. Perhaps: Overview of the field? Overview of the literature? Quantitative overview of the literature?}

\section{A Quantitative Overview of the Field}\label{sec:State_of_the_field} 

Fig.~\ref{fig:overview} (left), which shows the distribution of papers by year, indicates that %with the distribution of papers per year, shows that %interest in the topic of our review took off around the time BERT was published \cite{Devlin:etal:2019}, %\id{there couldn't have been papers on TLMs before, so I think the phrasing is a bit weird. Instead: interest in the topic of our review quickly took off after the appearance of BERT}
interest in the topic of our review quickly took off after the appearance of BERT \cite{Devlin:etal:2019},
peaking in 2022 and decreasing afterwards (though note that the 2025 data only covers up to July, so the final count will likely exceed that of 2024). As expected, initially most work focused on bidirectional TLMs, but in later years, causal models have dominated. Also note that most studies focus on one model type (either bidirectional, 59\%, or causal, 28\%), with very few comparing the two (13\%). 

Beyond the information shown in Fig.~\ref{fig:overview} (left), we found that studies predominantly evaluate monolingual models (66\%, the vast majority English), rather than multilingual (20\%) or both kinds (14\%). %see Fig.~\ref{fig:general_heatmap} in App.~\ref{app:general}).
%model type and language coverage in App.~\ref{app:fig_syntacticPhenomena}
%We report additional results on methods, experimental materials, and model type in  App.~\ref{app:method}.
%Fig.~\ref{fig:overview} (right) shows that, 
Also, consistent with previous observations \cite{waldis_holmes_2024, lopezotal:etal:2025}, most of the results in our database concern a single model: BERT, with 30\% of the results. BERT plus variants account for 58\% of the datapoints, and the four overall most studied models for 62\% (BERT, GPT-2, RoBERTa and mBERT). %, see Fig.~\ref{fig:overview_app} in App.~\ref{app:general}).
%(180/609 model results in our database), 356/609
%may appear worryingly skewed at first glance, 
The over-focus on BERT is likely due to BERT’s early availability, strong baseline performance, and suitability for a wide range of research questions.  As the field matures, research is increasingly moving toward a broader and more principled selection of architectures.
    
\paragraph{Linguistic phenomena \& language diversity.}\label{sec:rq1}Fig.~\ref{fig:overview} (right) shows that the distribution of syntactic phenomena is severely imbalanced:
Most research has been conducted on formal syntax (c.f. \textit{dependency parsing and constituency}, \textit{agreement and feature-checking}, \textit{POS and other lexical properties}---with most studies in this category focusing on POS---, \textit{hierarchical structure}, and \textit{word order}).
These are phenomena that are relatively easy to get at with current NLP tools, especially for English.
Instead, phenomena at the syntax-semantics interface have received much less attention (cf.\ \textit{binding and coreference}, \textit{negation and NPIs}, \textit{ellipsis and filler–gap dependencies}); and only a handful of studies concern typological aspects. %\id{I would delete these last 2 uses of "severely". It is not so much a problem with interpretability but with linguistic and particularly psycholinguistic research in general, which is where this distribution likely comes from.}
Somewhat surprisingly, only 42 studies used benchmarks, 29 of which used BLiMP \cite{Warstadt:etal:2020}.%
\footnote{The others were the CoLA task from GLUE \cite[][11 uses]{Wang:etal:2018}, SyntaxGym \cite[][3 uses]{gauthier:etal:2020}, Holmes and HANS  \cite[][1 use each]{mccoy_right_2019, waldis_holmes_2024}.} %For more information on these benchmarks, see App.~\ref{Appendix:benchmarks}.
%\id{comment: if we have space, I'd move footnote 6 to the main text}
%gb: done; if needed, move back to footnote.
Fig.~\ref{fig:overview} (right) also showcases the overwhelming dominance of English (in purple), which we already noted regarding the models (here, the data shows the languages \textit{tested} in the articles, not the languages covered by the model).
%We report additional results on model type and language coverage in App.~\ref{app:fig_syntacticPhenomena}. 

  %\begin{figure}[!ht]
        %\centering
        %\subfloat{\includegraphics[trim=0.25cm 0.3cm 21.9cm 0.6cm, clip, scale=0.37]{figures/phenomena11_languages-1.pdf}}
        %\subfloat{\includegraphics[trim=1.15cm 0.33cm 1.5cm 0.7cm, clip, scale=0.37]{figures/phenomena11_languages-1.pdf}}
        %\caption{Counts of syntactic phenomena by evaluated language (only English tested, English and other languages, and non-English).}
        %\label{fig:RQ1phenomena}
    %\end{figure}

\begin{figure*}[!ht]
  \begin{minipage}[c]{0.43\textwidth}
  \centering
  \includegraphics[trim=0cm 0.7cm 0cm 1.5cm, clip,  scale=0.6]{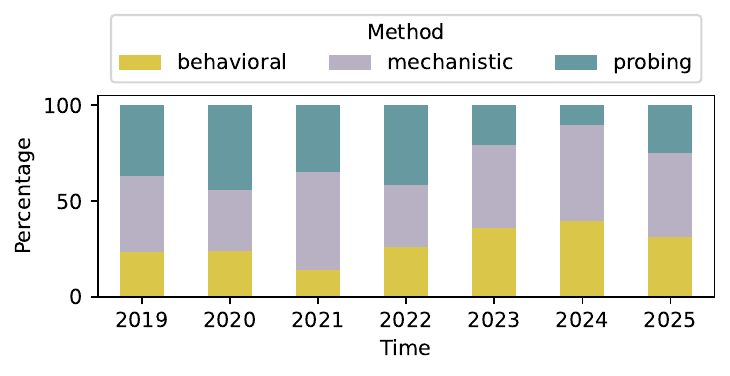}
  \includegraphics[trim=0cm 4.8cm 0cm 0.2cm, clip,  scale=0.6]{figures/method_years_percentage.pdf}
 \includegraphics[trim=0.2cm 0.3cm 0.2cm 0.2cm, clip,  scale=0.5]{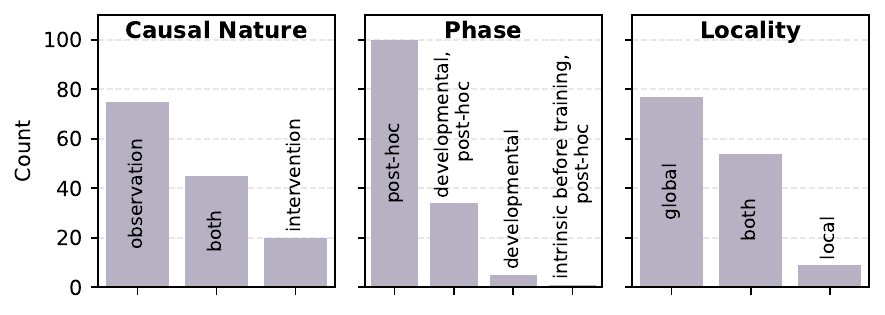}
 \end{minipage}
 \hspace{1.cm}
 \begin{minipage}[c]{0.43\textwidth}
 \centering
 \includegraphics[trim=0.245cm 0.15cm 0.235cm 0.2cm, clip, scale=0.52]{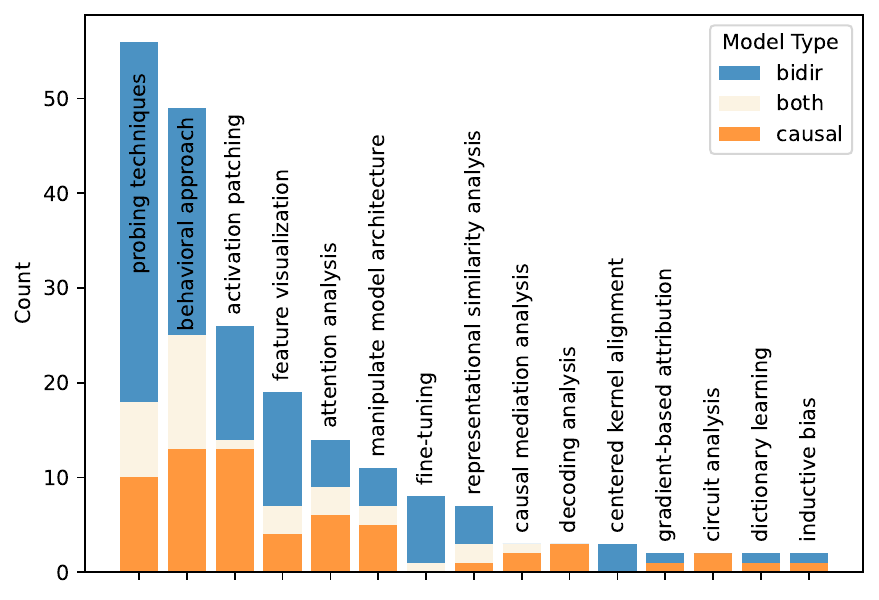}
\end{minipage}
%\vspace{1em} 
  %\subfloat{\includegraphics[trim=0cm 1cm 2.4cm 2cm, clip, scale=0.43]{figures/method_phenomena_percentage.pdf}}
  \caption{\textbf{Left:} Top: Percentage of methods over years. Bottom: mechanistic interpretability methods broken down by causal nature, training phase, and locality. \textbf{Right}: Most frequent interpretability techniques used in mechanistic papers; methods may be combined.} %\id{Legend & bars: i would order them as behab, prob, mech} 
  \label{fig:RQ2method}
\end{figure*}

\paragraph{Methods.}\label{sec:rq2} 
%We coded for the three main experimental paradigms that are currently used to study syntactic knowledge in TLMs: 
% Research about syntactic knowledge in TLMs, as in any scientific field, can be carried out with two broad approaches: observational, and interventional (or causal).
% As an example of the former, 
%While both are valuable, only intervention-based approaches can convincingly get at causation, as observational studies are subject to causation-correlation confounds.
Recall from above that we follow \citet{Milliere:2024}'s classification of interpretability methods into behavioral, probing, and mechanistic, as it provides a useful framework for organizing the literature. Fig.~\ref{fig:RQ2method} (left-top) shows a relatively balanced distribution of these methods over the years.
Behavioral methods assess models based on their %input–
output behavior at the final layer. %This includes both analysing generated outputs and comparing model-assigned probabilities for competing alternatives in controlled settings, such as minimal pairs~\cite{Linzen:etal:2016,Lau:etal:2017}. 
They include analysis of generated outputs and comparisons of model probabilities in controlled settings \cite[e.g., minimal pairs,][]{Linzen:etal:2016,Lau:etal:2017}.
The latter paradigm, highly influenced by the psycholinguistic tradition, treats the higher probability of a correct form compared to an incorrect alternative as evidence that the TLM has mastered a given syntactic construct. 
As an example, for the anaphor agreement item from the BLiMP dataset \textit{Many girls insulted \underline{themselves}/*\underline{herself}}, we would expect a well-trained model to assign higher probability to \textit{themselves} than to \textit{herself}.
%See App.~\ref{app:general}, Fig.~\ref{fig:blimp_cats} %App.~\ref{Appendix:benchmarks} 
%for further example stimuli.}
%For instance, example~\ref{ex:Brett} shows a filler-gap dependency stimulus from the BLiMP dataset \cite{Warstadt:etal:2020}.
%A model that masters filler-gap constructions is expected to assign higher probability to \textit{what} than to \textit{that} for this stimulus. 
%\ex. \label{ex:Brett} Brett knew \underline{what}/*\underline{that} many waiters find.

Probing is an interpretability method that became very popular after \citet{hewitt_structural_2019} introduced their structural probe, which is why it is in a category on its own.
It aims at detecting specific types of information in the models' internal representations, often by training auxiliary classifiers on activations from specific layers \cite{waldis_holmes_2024}. 
%%add in 9-page version:
%%
The argument goes that if, for instance, a classifier trained to recover the POS of input words from a given hidden layer is successful, then that layer encodes POS information.
%\footnote{Mechanistic methods may (and often do) incorporate behavioral or probing elements; in the database, we categorized studies investigating internal dynamics, even if including behavioral or probing tests as well, as mechanistic.}

Finally, mechanistic approaches actually cover a wide array of specific methods, see Fig.~\ref{fig:RQ2method} (right) for the most common ones. These methods may (and often do) incorporate behavioral or probing elements. In our database, we categorized studies investigating internal dynamics as mechanistic, even if they included behavioral or probing tests as well. Overall, mechanistic approaches aim at getting at internal mechanisms by inspecting model components (e.g., layers, neurons, circuits, or training dynamics) and, where possible, intervening on them to causally link representations to behavior \cite{Wiegreffe:Saphra:2024}. %
%%add in 9-page version:
%%
For instance, these methods may modify the activations of components identified as syntactically crucial and test the effects of these interventions on model behavior.
%%
%\no{As an example, \citet{zhang:etal:2025} show that the BLOOM model uses a shared mechanism for indirect object identification in English and Mandarin, alongside language-specific components for language-specific processing. This pattern aligns with broader evidence that models organize syntactic knowledge into phenomenon-specific mechanisms, with subject-verb agreement standing out as a shared cross-linguistic computation \citep{kryvosheieva:etal:2026}. Consistent with this, 
As an example, \citet{ferrando_similarity_2024} identify a  subject-verb agreement mechanism for English and Spanish in the Gemma 2B model. %both fo path patching
They substantiate their causal claim by, a.o., showing that steering the activations of Spanish inputs (for the relevant components of the model) in the direction of English activations flips their predicted verb number.
%argue that in the multi-lingual model Gemma 2B the subject number is represented as a language-independent direction in the residual stream. They provide support for their claim by showing

%Mechanistic methods often incorporate behavioral elements but focus on causal interventions rather than mere correlation. In our database, studies using only behavioral methods are categorized separately, while those exploring internal dynamics (regardless of behavioral tests included) are classified as mechanistic.
%This is where mechanistic interpretability methods come in: these methods, very varied in nature, inspect and intervene on internal components (e.g., layers, neurons, circuits, or training dynamics) to causally link representations to behavior \cite{Wiegreffe:Saphra:2024}.

%Behavioral methods offer experimental control, are simple to implement and, if the confounds are well chosen, informative.
%However, behavioral performance alone does not guarantee underlying linguistic competence, and behavior doesn't provide insight into internal mechanisms.
All methods have their pros and cons. For instance, probing offers scalability and clear feature detection but, on its own, lacks causal insight: decoding linguistic features is not equivalent to showing that the identified representations are actually used in predictions \cite[besides other issues, see][]{Belinkov:2022,Wiegreffe:Saphra:2024,agarwal:etal:2025}. %
%The different mechanistic methods again have their own pros and cons (see below for more information on this subcategory).
%
%Given that no single method is enough, thus, and that all methods have pros and cons, 
Therefore, it is reassuring that the field has been using a healthy variety of methods throughout (Fig.~\ref{fig:RQ2method}, left-top). %We see that in general the use of the different paradigms is relatively balanced across years, with no single method dominating.
%Overall, their use remains balanced. 
The only significant temporal trend is that probing methods have diminished substantially in recent years, possibly due to the growing awareness of their limitations.\footnote{For instance, following the publication of the influential criticism in \citet{Belinkov:2022}, the number of probing studies with bidirectional TLMs in our database declined sharply, from 27 studies in 2022 to just 11 in 2023.} %Fig.~\ref{fig:RQ2method_phenomena}.
However, there is one aspect that we believe should improve: There are still too few studies that can get at causal mechanisms (only 19\% use interventional methods, see App.~\ref{app:sec_3}, Fig.~\ref{fig:method_growth_stacked_trend}).
%compared to 81\% that use only probing, behavioral, or observational mechanistic methods).

%\begin{figure}[!ht]
  %\centering
  %\subfloat{\includegraphics[trim=0.245cm 0.15cm 0.235cm 0.2cm, clip, scale=0.49]{figures/MImethods_stacked-2.pdf}}
  %\caption{Most frequent interpretability techniques used in mechanistic papers; methods may be combined.}\label{fig:RQ2MI_methods} 
%\end{figure}

We next delve a bit more into mechanistic studies, largely following the taxonomy from \citet{bereska2024mechanistic}.
%the methods used, %causal nature, phase, and locality
%the causal nature of the approach, the phase of application, and the locality of the analysis.
We annotated the 140 mechanistic papers in our database along four dimensions: (i) \textit{causal nature}, indicating whether the methods used are observational, interventional, or both; (ii) \textit{phase}, indicating when in the TLM lifecycle the analysis is performed—before training (intrinsic), during training (developmental), or after training (post-hoc); (iii) \textit{locality}, indicating the scale at which the model is analyzed, from global representations (e.g., the residual stream) to local units (e.g., individual neurons); and (iv) \textit{specific methods} used, as described by the authors (with some generalization). The first three dimensions are shown in Fig.~\ref{fig:RQ2method} (left-bottom) and the most frequent methods are shown in Fig.~\ref{fig:RQ2method} (right).
We find over 80 specific methods in mechanistic studies, which we take to be a positive development since mechanistic interpretability is still a young field. %\id{Now that we have extra space, I think that probing and behavioral featuring as the two most freq mech methods deserves to be explained/clarified in the main text to avoid confusion.}
Recall that mechanistic papers often include also probing and behavioral methods, and indeed they show in Fig.~\ref{fig:RQ2method} (right) as the two most frequent approaches even within mechanistic papers. 
%The next two most frequent are activation patching (intervening on internal activations to test causal contribution, as explained above), and feature visualization (e.g. the use of PCA, t-SNE and attention heatmaps to explore internal representations).
%As observed above, however, even in mechanistic work most work relies on observational approaches.
Post-hoc analysis of fully trained models at a global level clearly dominates. However, there is a healthy amount of papers that track development, and many also check both global and local trends.
Additional results for this section can be found in App.~\ref{app:sec_3}.

\section{Syntactic Knowledge in TLMs}\label{sec:Syntactic_Knowledge_TLMs}
%Justify our approach without saying it is as a justification:)
%We originally set out to map how TLMs process syntax on the basis of the full database, however, this goal proved unrealistic, given the daunting conceptual and methodological variation across studies (see §~\ref{sec:conclusions} for discussion).
%To maintain an objective and quantitative analysis, we originally set out to map how TLMs process syntax on the basis of the full database, however, this goal proved unrealistic, given the daunting conceptual and methodological variation across studies (see §~\ref{sec:conclusions} for discussion). We therefore begin with behavioral evidence before turning to probing-based and mechanistic interpretability studies, moving from externally observable performance toward analyses of internal representations and processing mechanisms.
We now examine what these studies reveal about syntactic knowledge in TLMs, starting with behavioral evidence from papers that measure model accuracy, %providing an overview of syntactic performance across studies
and then turning to results from probing and mechanistic approaches. Additional results for this section, including complete statistical model specification, %and coefficients, 
are provided in App.~\ref{app:sec_4}. %, which enable analysis of internal representations and processing mechanisms beyond the model's output alone.

\begin{figure*}[!ht]
  \centering
  \subfloat{\includegraphics[trim=15.55cm 1.05cm 0.24cm 0.11cm, clip, scale=0.455]{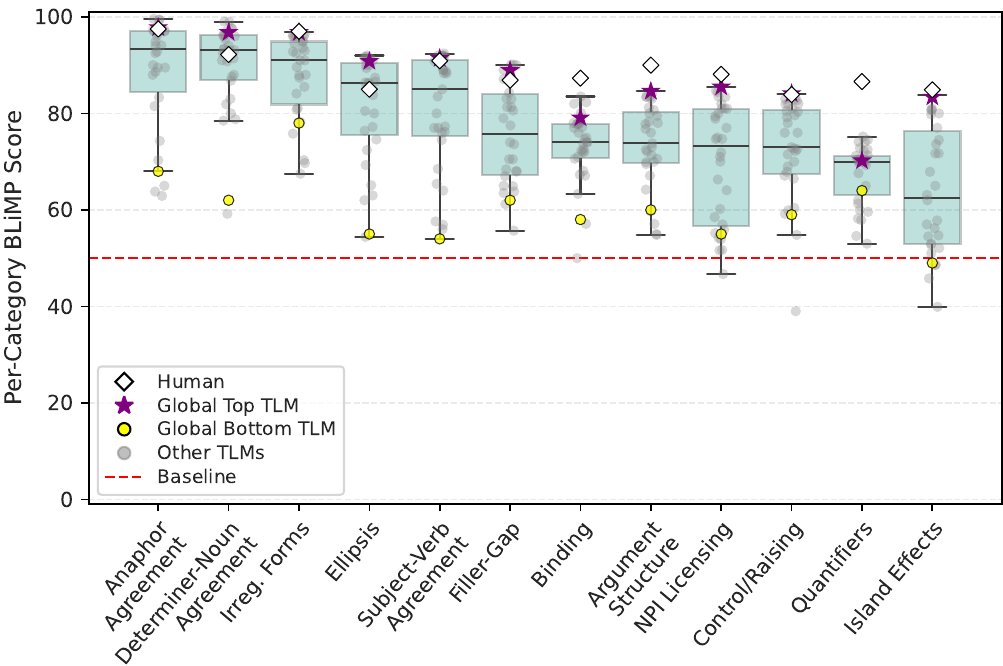}}
  \subfloat{\includegraphics[trim=0.1cm 0.13cm 0.2cm 0.2cm, clip, scale=0.42]{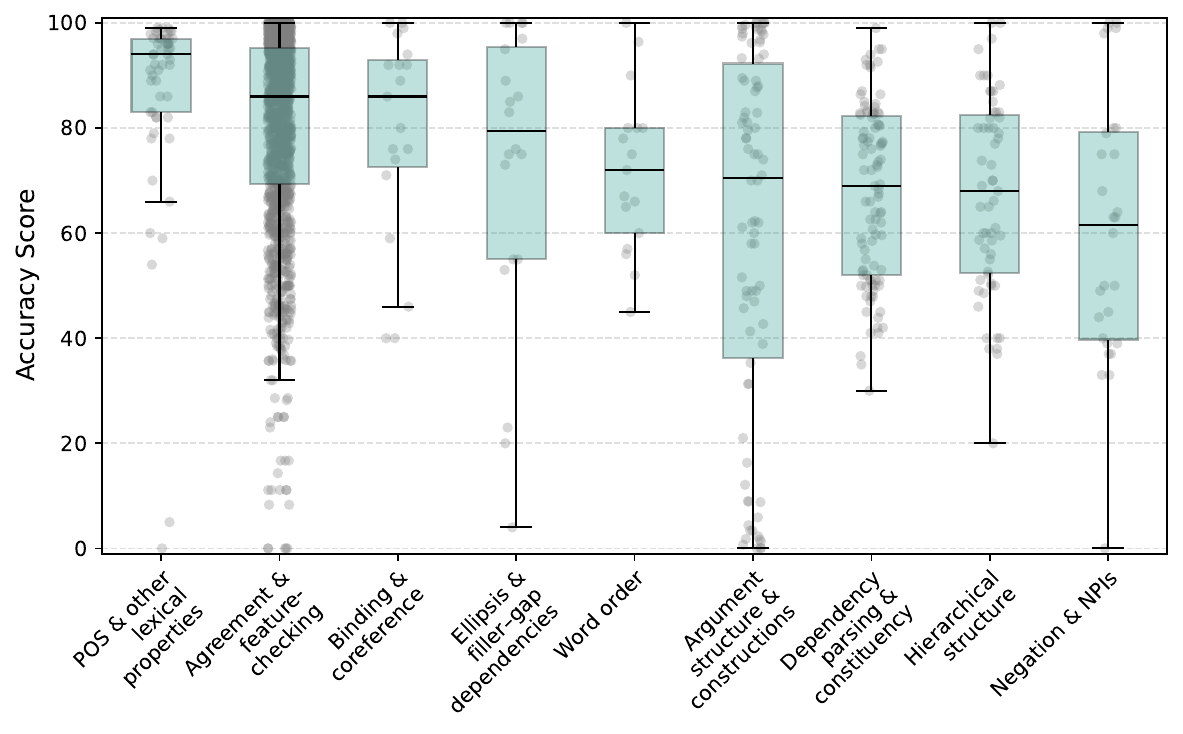}}
  \caption{\textbf{Left}: \textbf{BLiMP (English-Only)} scores by category. Human performance {\protect\textcolor{black}{$\diamond$}} from \citet{Warstadt:etal:2020}. Top {\protect\textcolor{purple!60!black}{$\star$}} (Roberta large) and bottom {\protect\contour{black}{\protect\textcolor{yellow}{\textbullet}}} (Minibert) TLM among those with complete per-category results. \textbf{Right}: \textbf{Accuracy scores (113 languages)} by linguistic phenomena, covering more than 1,500 data points from 76 studies. Note that $77\%$ of the datapoints for Agreement \& feature-checking originate from one study \citep{jumelet:etal:2025}.}\label{fig:RQ3blimp} %76.5
\end{figure*}

\subsection{Behavioral Evaluation}
%We originally set out to map what is known about how TLMs process syntax on the basis of the whole database; however, this goal proved unrealistic, given the daunting conceptual and methodological variation across studies (see Section~\ref{sec:conclusions} for discussion). Responding to these difficulties, and in order to keep our review objective and quantitative, we chose to focus on the results obtained with the BLiMP benchmark, which allows for systematic comparisons, and for which we have enough datapoints. In addition, we analyze overall accuracy scores across linguistic phenomena (excluding benchmark-specific analyses), covering more than 1,500 data points from 76 studies and spanning 113 languages.

%In the following, we first analyze BLiMP results, which provide a standardized English benchmark for minimal-pair evaluation \cite{Warstadt:etal:2020}. We then examine over 1,500 accuracy scores from 76 non-benchmark studies across 113 languages, offering a broader but more heterogeneous view of syntactic knowledge.
We first report on BLiMP benchmark results for English, which provide a controlled setup where results are comparable, and then turn to non-benchmark studies for 113 languages, which offer a broader but more heterogeneous view.

%we analyze behavioral results from the BLiMP separately, as it provides a standardized evaluation setting with sufficient datapoints for systematic comparison in English, while reflecting the typical scope and constraints of benchmark-based minimal-pair evaluations. In addition, we examine overall accuracy scores across linguistic phenomena from non-benchmark studies, covering more than 1,500 data points %from 76 studies and across 113 languages. Although these evaluations are methodologically more heterogeneous and less directly comparable, they offer a broader perspective on syntactic assessment across tasks and languages. %models, 
%\id{I'm missing from this and the following section on behavioral results an explanation of why we analyzed the results from blimp separately from other studies. I would do so at the beggining of the behavioral section}

\paragraph{BLiMP.} %We originally set out to map what is known about how TLMs process syntax on the basis of the whole database; however, this goal proved unrealistic, given the daunting conceptual and methodological variation across studies (see Section~\ref{sec:conclusions} for discussion).
%Responding to these difficulties, and in order to keep our review objective and quantitative, we chose to focus on the results obtained with the BLiMP benchmark, which allows for systematic comparisons, and for which we have enough datapoints.
The BLiMP benchmark \cite{Warstadt:etal:2020} is designed for behavioral evaluation of models through minimal pair comparison.\footnote{In addition to BLiMP, the database contains a few papers featuring BLiMP-inspired benchmarks for other languages, including CLiMP \cite{xiang:etal:2021}, SLING \cite{song:etal:2022}, and RuBLiMP \cite{taktasheva:etal:2024}. This subset is too small to analyze separately and is therefore not included in this analysis.}
% The dataset covers a broad range of linguistic phenomena, shown in Fig.~\ref{fig:RQ3blimp} (left).
Our analysis includes English BLiMP scores from 25 papers: 11 that report by-category results and 14 that report average BLiMP scores. %excluding results from papers where final BLiMP scores were not reported. %excluding results from four further papers due to methodological issues. \id{excluding results from papers where BLiMP scores were not reported + delete footnote}%\footnote{The four studies propose alternative metrics, derived from BLiMP, or do not provide final BLiMP scores.} %We report additional results on average BLiMP scores as a function of training data size,  and model type in App.~\ref{app:rq3_blimp}, observing the well-known scaling pattern in deep learning: TLMs trained on larger amounts of data tend to perform better.
%Fig.~\ref{fig:RQ3blimp} (left) displays the distribution of TLM scores divided by linguistic phenomenon.%\id{I'm confused: in the previous paragraph you introduce figure 6 left and say its 25 papers. Then you introduce it again and say it's 11.)}
Fig.~\ref{fig:RQ3blimp} (left) shows accuracies by category, with an average of 31 datapoints per category, as most articles analyze multiple models.\footnote{Only results from fully trained models are included.} %\footnote{If a paper reports scores for TLMs trained on varying sizes of training data, we include only the fully trained models in the boxplot.}
The most salient feature of these results is how well the models perform on this benchmark. Notably, the best TLM is at or above human performance for 7 out of the 11 categories, and all but 8 out of the 368 results are above the baseline. Another salient feature is the large within-phenomenon variance across TLMs. We find that this variance is mostly driven by how good each model is at syntax generally, as measured by its overall BLiMP score: Model rankings across phenomena are highly consistent.
This is illustrated in Fig.~\ref{fig:RQ3blimp} (left) by highlighting the best (purple star) and worst (yellow dot) TLMs (see also Fig.~\ref{fig:RQ3blimp_size} in App.~\ref{app:rq3_blimp} for quantitative support).
%using Kendall’s τ on the basis of model ranking per phenomena. 
In this respect, additional statistical analyses of average BLiMP scores confirm the expected scaling trend: larger TLMs and TLMs trained on more data generally achieve higher performance, with bidirectional models tending to outperform causal models (see App.~\ref{app:rq3_blimp}).

Despite the large variance, the results also clearly suggest that TLMs perform better on formal syntactic relations (e.g.\ all agreement categories have medians over 85\%) than on phenomena at the syntax-semantics interface (binding, argument structure, NPI licensing, control/raising, quantifiers and island effects all have medians below 75\%).
%However, there is one exception to the observed pattern: ellipsis is also a phenomenon at the syntax-semantics interface, and TLMs seem to excel at it, with several models matching or outperforming humans. However, this BLiMP category, as noted by \citet{Warstadt:etal:2020}, covers only specific cases of noun phrase ellipsis whose grammatical violation we suspect might be easier to detect than others. Thus, the exception could be due to the particular characteristics of BLiMP rather than ellipsis as a phenomenon being easy for models.
One exception is ellipsis: despite its reliance on interactions between syntax and semantics, several TLMs match or outperform humans on this category. However, BLiMP evaluates only specific cases of noun phrase ellipsis, whose violations may be comparatively easy to detect \citep{Warstadt:etal:2020}, suggesting this result may reflect dataset characteristics rather than ellipsis as a phenomenon being easy for models.

%More generally, note that BLiMP does not comprehensively cover syntactic phenomena, even in English, as it includes only selected categories and, within each of these categories, only cases that can be readily transformed into minimal pairs. %BLiMP’s coverage of syntactic phenomena is not comprehensive, even for English; despite its breadth, it contains only some selected categories and, for each category, it includes only cases that can be straightforwardly transformed into minimal pairs.
%Furthermore, as discussed by \citet{Kulmizev:Nivre:2021}, BLiMP aggregates diverse phenomena into composite scores with equal weighting, thereby masking important differences in difficulty and relevance. 
%Qualitative analysis of model behavior is difficult, and no single dataset is definitive; results must therefore be interpreted in light of dataset characteristics. Qualitative analysis of model behavior is difficult, and no single dataset can be expected to be definitive. It is thus important to interpret results in the light of dataset characteristics.
%Qualitative analysis of model behavior is difficult, and since no dataset is definitive, results should be interpreted with consideration of the dataset's characteristics.

More generally, despite its value for enabling cross-model comparisons across phenomena, BLiMP still provides limited coverage of syntax, even for English. It focuses on cases that can be readily expressed as minimal pairs and aggregates heterogeneous cases into equally weighted composite scores, potentially obscuring important differences in difficulty and relevance \cite{Kulmizev:Nivre:2021}. As a result, model performance should be interpreted with the characteristics and limitations of the benchmark in mind.

%Accuracy in the Wild. \id{
\paragraph{Beyond benchmarks and English.}%Out-of-Benchmark Evaluation: Behavioral results beyond Controlled Benchmarks.
%- Extend accuracy analysis beyond BLiMP by aggregating results from all behavioral and mechanistic studies in the database (covering X papers and XXX languages)
%- Observe similar performance trends as in BLiMP under our classification framework
%- Group languages into  language support categories (following Joshi 2021)
%- Statistical modeling shows that both language support and model type (causal vs. non-causal) are significant predictors of performance (appendix put a forest plot of different coef. of phenomena of statistical model)

\begin{figure}[!ht]
    \centering
    \subfloat{\includegraphics[trim=0.2cm 0.2cm 0.2cm 0.7cm, clip, scale=0.51]{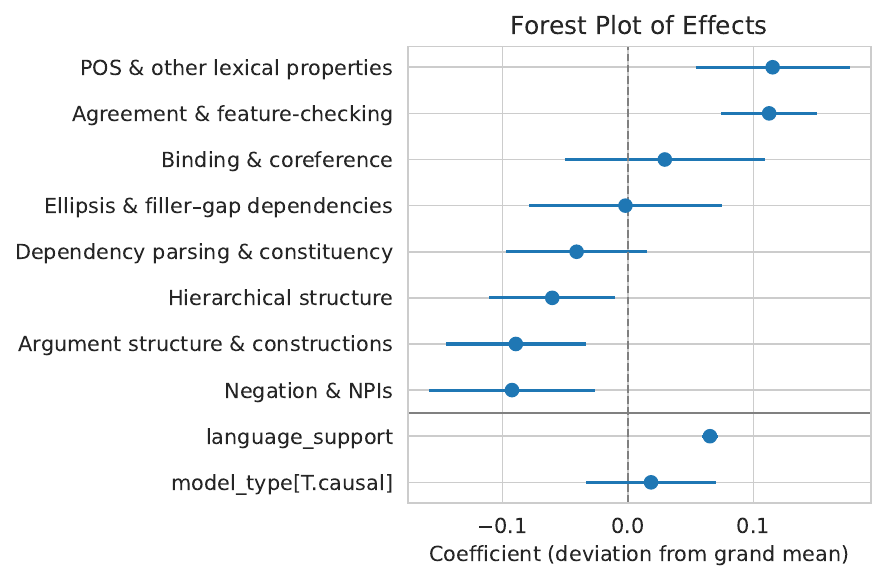}}
    \caption{%Forest plot of 
   Coefficients from a regression model predicting accuracy scores based on syntactic phenomena, model type, and language support. Horizontal lines show 95\% confidence intervals.
    }\label{fig:forestplot}
\end{figure}

We next examine all the non-benchmark accuracy scores from the database, %behavioral and mechanistic papers, %accuracy scores from all behavioral and mechanistic studies, 
yielding 1,557 datapoints %\footnote{Including the linguistic category \textit{benchmark} would yield 1773 datapoints.} 
from 76 studies spanning 113 languages. Fig.~\ref{fig:RQ3blimp} (right) presents results for our annotated categories of linguistic phenomena. %While the  picture is more varied, we observe a similar performance trend: agreement is much simpler for TLMs than the syntax-semantics interface. Binding appears to be an exception, although this may be attributable to the limited number of available data points.
The variance this time is even greater than in the BLiMP case, which is not surprising given that different studies use different materials and test different languages.
Since this makes for an unclear picture at a descriptive level, we turn to statistical modeling to investigate the factors influencing accuracy. In particular, we fit a mixed-effects regression model with syntactic phenomena, model type (causal/bidirectional), and language support as fixed effects, and model name and study ID as random effects. 
%use as predictors syntactic phenomena, model type (causal / bidirectional), and language support to see how syntactic knowledge varies across languages. %\footnote{\no{We include model and study id as random effect?}
%We group all languages into language support categories following the taxonomy of 
Language support gets at the availability of NLP resources and datasets for a given language. We take the digital support data from \citet{Joshi:etal:2020}, who define a 5-level classification ranging from low- to high-support (App.~\ref{app:rq3_non-blimp}, Fig.~\ref{fig:language_support}). %\footnote{\citet{Joshi:etal:2020} introduce a taxonomy of language digital support based on the availability of language resources, such as datasets and NLP tools, using this to categorize languages from low- to high-resource settings. We adopt this grouping as a proxy for language support.}
We find that language support and
%both model type and
linguistic phenomena categories are significant predictors of accuracy scores. 
%causal models outperform bidirectional models by 10\% points ($\beta= 0.1$, $p<0.001$). In addition, 
As could be expected, performance increases with greater language support, with each higher support category associated with a gain of $\sim7$\% points ($\beta= 0.065$, $p<0.001$). %See App.~\ref{app:rq3_non-blimp} for additional figures and the complete model specification and coefficients; the estimated intercepts for the linguistic phenomena categories, visualized in Fig.~\ref{fig:forestplot}, reveal a similar trend across linguistic categories, as in Fig.~\ref{fig:RQ3blimp} (right).
The estimated intercepts for the linguistic phenomena categories, displayed in Fig.~\ref{fig:forestplot}, confirm that \textit{POS and other lexical properties} and \textit{agreement and feature-checking} have generally higher accuracies than average, while \textit{hierarchical structure}, \textit{negation \& NPIs} and \textit{argument structure \& constructions} have lower accuracies. For the other phenomena, there is no evidence either way (see Fig.~\ref{fig:forestplot}).
This result largely confirms that formal syntactic phenomena are easier for TLMs than phenomena at the syntax-semantics interface, though the evidence is more mixed than in the case of the BLiMP benchmark. While models perform better on  \textit{binding \& coreference} than on \textit{hierarchical structure}, we suspect that this is because the latter contains more heterogeneity than \textit{binding \& coreference}, similar to the results for ellipsis we pointed out above in the BLiMP benchmark. %\gb{I think we should explain a bit more here, signalling that the evidence is a bit more mixed than in the case of English: hierarchichal structure is not synt-sem interface and is more difficult, the binding and ellipsis categories are synt-sem and are easier than others. Iria?} \id{agree: one possible explanation fo the lower score for hierarchical structure is that it contains more heterogeneity than, say, binding. It's a bit like the ellipsis case in blimp}

%upon acceptance: include forset plot here

%Although the overall picture is more varied and less clear than for BLiMP, POS and agreement (formal syntax) seems easier for TLMs %than phenomena at the syntax–semantics interface. 
%and NPI (syntax-semantics) as most difficult, but the order doesn’t follow the divide as neatly as with blimp. And I’d say that more research is needed to determine what’s going on. Nevertheless, the variation within categories indicates that difficulty does not strictly follow a surface-to-abstract hierarchy; POS tagging and related lexical-syntactic properties remain high in performance, while hierarchical structure, dependencies, and negation/NPIs show lower and more variable behavior. 
%The overall picture is more heterogeneous than for BLiMP: formal syntax phenomena such as POS tagging and agreement appear easier for TLMs than syntax–semantics interface phenomena such as NPIs, though the ordering is less clear than in BLiMP and warrants further research. Variation within categories suggests that difficulty does not strictly follow a surface-to-abstract hierarchy: POS and related lexical-syntactic tasks remain high-performing, whereas hierarchical structure, dependencies, and negation/NPIs show lower and more variable results. Binding appears to be an exception, although this may be attributable to the limited number of available data points.

%, consistent with our earlier observations. % with figures and additional details on the statistical model provided in App.~\ref{app:rq3_non-blimp}

\subsection{From Behavior to Internal Mechanisms} %/Beyond Behavioral
%We code all probing and mechanistic paper that feature the probing method in the DB, 147 in total, by whether they state in their results clearly find syntactic representations or not. Findings: probing often successful (but less if mechanistic), bring up the NLP pipeline argument? We annotate whether they state where the probe is most successful and find that out of 53 papers,  middle layers are mainly reported (but just stronger), a lot of bidir models tested,

\begin{figure}[!ht]
  \centering
  \subfloat{\includegraphics[trim=0.25cm 0.2cm 0.25cm 0.2cm, clip, scale=0.45]{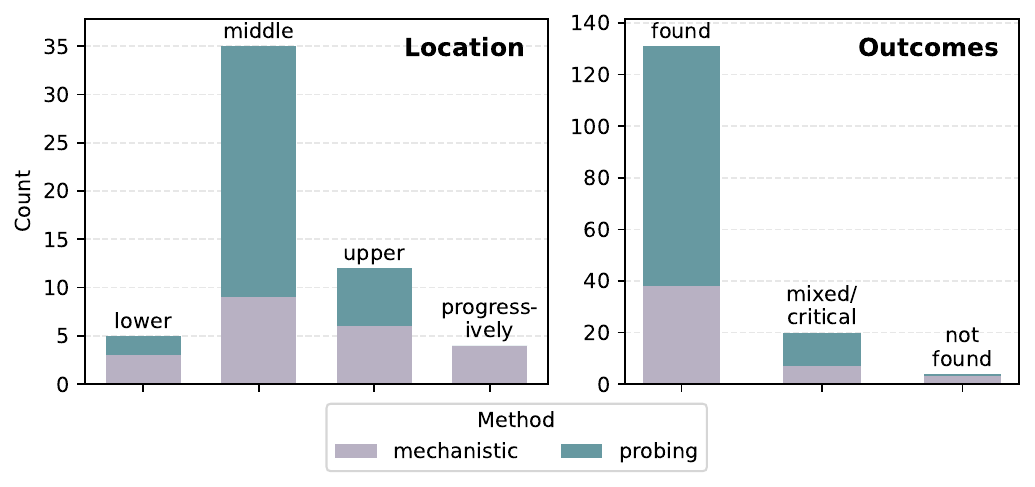}}
  \caption{Probing-based studies. \textbf{Left:} location in TLMs where probing performance is strongest. \textbf{Right:} author-reported evidence for syntactic representations in TLMs. Note that some mechanistic papers include probing methods; these are included.}
  \label{fig:RQ3probing}
\end{figure}

%\textbf{3.3. Beyond behavioral}: \textbf{Probing}
%- 147 studies analyzed using probing methods on TLMs
%- Probing studies often claim models capture syntactic information while mechanistic studies are more cautious and critical about such claims
%- Difference likely due to stronger causal standards in mechanistic approaches
%- among 53 studies reporting layer effects, middle layers perform best
%- Middle layers seen as strongest, not solely responsible for syntax
%- Most studies focus on bidirectional TLMs due to their prevalence
%-  similar ratio for found/not found across ling. phenomena
\paragraph{Probing.}
%We analyze all studies that employ probing methods (147 papers in total), which includes all papers coded as probing, plus some coded as mechanistic when they also employed probing methods.
Probing has been primarily used to localize layers where syntactic information is most readily available. 
We code the 56 probing-based studies that include this information (including mechanistic papers that also use probing)
%Similar to \citet{lopezotal:etal:2025},\footnote{\citet{lopezotal:etal:2025} report on 5 BERT model studies.} who report on the potential layerwise location of syntactic phenomena, \id{not sure the ref. to lopez otal is necessary...we could start by saying that we annotated for location, the lower, middle, upper etc. is not from them, right?}  (lower, middle, upper)
for the model locations at which studies report the strongest probing performance in detecting syntactic information. Fig.~\ref{fig:RQ3probing} (left) shows that the middle layers are most commonly identified, corroborating previous observations \cite{lopezotal:etal:2025, Skean:etal:2025}. % %yielding the highest probing performance or the strongest effects when representations are intervened on.
%\footnote{This pattern is consistent with previous reviews such as \citet{lopezotal:etal:2025}, who report similar findings across five BERT-based studies concerning the layerwise location of syntactic phenomena.}  
Despite the prominence of middle layers, we should highlight that studies typically describe them as showing stronger, rather than exclusive, encoding of syntactic information.\footnote{We also observed that the distinction between “middle” and “upper” layers is somewhat subjective and varies across studies.} %Understanding the representations in TLMs centers on how information flows and accumulates across layers, and some mechanistic studies build on this by examining progressive knowledge encoding \cite[e.g.,][]{he:etal2025}, consistent with staged interpretations of decoder-only models, which mirror a progression from surface-level cues to deeper syntactic and hierarchical processing \cite{Lad:etal:2024}.
%\id{It's not entirely clear why this follows} This aligns with a view of TLM representations as distributed across layers, motivating analyses of how information accumulates and transforms throughout the network.
%\no{This is consistent with a view of TLM representations as distributed across layers, 
This aligns with the perspective that TLM representations are distributed across layers, where information accumulates and transforms throughout the network, instead of being confined to one specific stage. In this vein, some mechanistic studies examine progressive knowledge encoding \cite[e.g.,][]{he:etal2025}, consistent with staged interpretations of causal TLMs that reflect a transition from surface-level cues to deeper syntactic and hierarchical processing \cite{Lad:etal:2024}. Unlike a strict pipeline interpretation \cite[e.g.,][]{jawahar_what_2019}, however, this staged view does not assume discrete functional modules, but rather a gradual and overlapping transformation of representations across depth \cite[e.g.,][]{niu_does_2022}.

\begin{figure*}[!ht]
  \centering
  \subfloat{\includegraphics[trim=0.25cm 0.25cm 0.2cm 0.2cm, clip, scale=0.55]{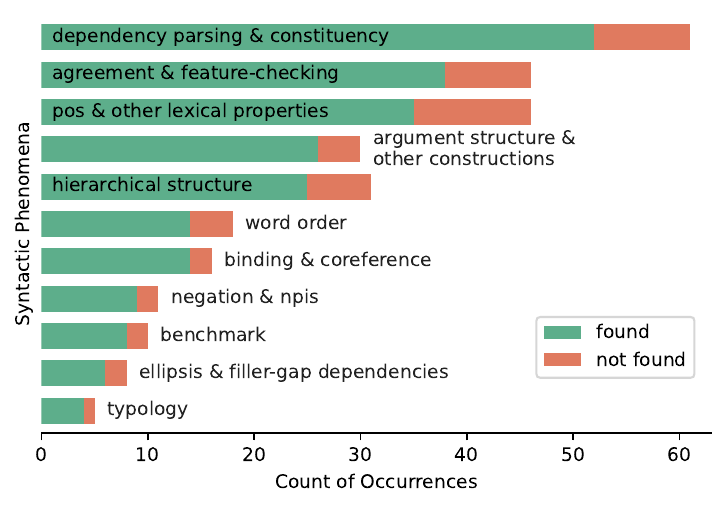}}\hspace{1.5cm}
  \subfloat{\includegraphics[trim=0.55cm 0.25cm 0.55cm 0.24cm, clip, scale=0.55]{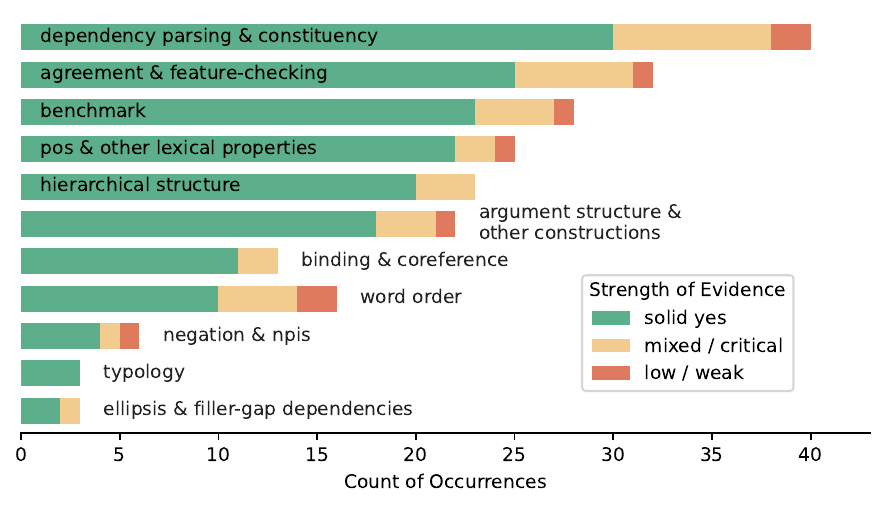}}
  \caption{Across linguistic phenomena, \textbf{left:} successful recovery of syntactic information from TLMs' representations via \textbf{probing}. \textbf{Right:} Evaluation of syntactic awareness in \textbf{mechanistic interpretability} studies.}\label{fig:RQ3probing_lingPhenomena}
\end{figure*}

Furthermore, we check 147 probing-based studies
for their outcomes: Each study was coded according to whether probing recovered syntactic information from TLM representations, yielded mixed results, or failed to recover such information. %see App.~\ref{app:probing} for further details. 
Fig.~\ref{fig:RQ3probing} (right) shows the results.
85\% of studies report successful recovery of syntactic information from model representations.
Although this high success rate may partly reflect publication bias toward positive results, this again points to substantial syntactic knowledge in TLMs. 
Fig.~\ref{fig:RQ3probing_lingPhenomena} (left) does not reveal substantial differences across syntactic phenomena in whether studies successfully recover syntactic information from model representations or yield inconclusive results. %(App.~\ref{app:probing}, Fig.~\ref{fig:RQ3probing_lingPhenomena}). 
This contrasts with behavioral results, which do show clear differences in how easily models handle different syntactic phenomena. The mismatch provides further support to the idea that successful probing is not, by itself, a reliable measure of the quality or functional use of syntactic representations in TLMs \cite{agarwal:etal:2025}.

%Building on this, some mechanistic studies specifically address progressive knowledge encoding in TLMs \cite[e.g.,][]{he:etal2025}, which align with recent work by \citet{Lad:etal:2024} on decoder-only TLMs.\footnote{\citet{Lad:etal:2024} suggest a staged interpretation of TLM behavior, which similarly mirrors a progression from surface-level cues to deeper syntactic and hierarchical processing.} %Note that studies largely concentrated on bidirectional TLMs only ($>$75\%), reflecting the imbalance identified previously (see App.~\ref{app:probing}). % %the dominance of such architectures in probing-based investigations of syntactic representations.

\paragraph{Mechanistic interpretability.}

\begin{figure}[!h]
  \centering
  %\subfloat{\includegraphics[trim=0.25cm 0.25cm 0.25cm 0.25cm, clip, scale=0.31]{figures/Counts_mech_syntactic.pdf}}\\
  \subfloat{\includegraphics[trim=0.27cm 0.285cm 0.75cm 0.25cm, clip, scale=0.45]%{figures/evidence_by_language_mech-2.pdf}
  {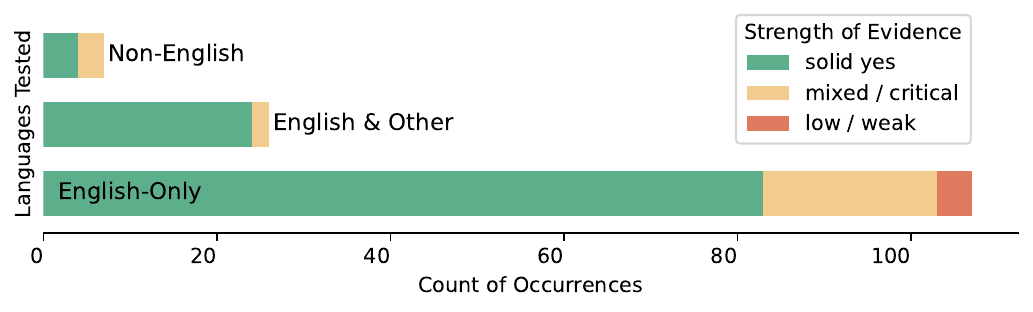}}
  \caption{Evaluation of syntactic awareness in mechanistic interpretability studies %\textbf{top:} by linguistic phenomena; \textbf{bottom:} 
  by language tested.}\label{fig:RQ3mechanistic}
\end{figure}

%\textbf{3.4. Beyond behavioral}: \textbf{Mechanistic}
% - most studies report evidence
% -skewd regarding language
%Above, we examined probing-based evidence; however, such results are 
%Probing-based evidence is inherently limited to demonstrating correlations between representations and linguistic structure. %To advance our analysis,
Our attempt to systematically synthesize what mechanistic studies reveal about syntactic mechanisms in TLMs faced a major challenge: once studies are broken down by method and phenomenon to enable meaningful comparisons, the resulting groups become too sparse and heterogeneous for a straightforward synthesis. We therefore decided to %instead 
focus on author-reported strength of evidence for the presence of syntactic knowledge, similarly to our approach with probing studies.
%We explore whether the 140 mechanistic studies in our database report to find evidence of syntactic knowledge, that is, whether 
%authors claim syntactic knowledge and processing in TLMs.
%report to find evidence of syntactic knowledge
The results, shown in Fig.~\ref{fig:RQ3mechanistic}, are positive: 79\% of the 140 mechanistic papers in our database find clear supporting evidence. At the same time, Fig.~\ref{fig:RQ3mechanistic} again underscores that the vast majority of this evidence comes from studies focusing exclusively on English. %Notably, the proportion of Non-English studies reporting mixed or critical outcomes is significantly higher than that of studies incorporating English, %showing positive results, although they still represent only a fraction of the overall literature. 
Again, as shown in Fig.~\ref{fig:RQ3probing_lingPhenomena} (right), we do not observe salient differences across phenomena. Importantly, however, phenomena at the syntax-semantics-interface have low coverage in the mechanistic literature, so this is still an open point. %(see App.~\ref{app:mechanistic}, Fig.~\ref{fig:RQ3mechanistic_app}).

\section{Discussion \& Conclusion}
\label{sec:conclusions}

\paragraph{Conducting systematic reviews in CL/NLP.}%Conducting this systematic review has also allowed us to identify several aspects that need improvement within the community. We end by proposing a few recommendations and directions for future research. 
%To our knowledge, this is among the first systematic reviews in CL and NLP. 
Systematic reviews remain relatively uncommon in CL/NLP. 
As the field of AI continues to grow at breakneck speed, they (together with meta-reviews) will become ever more important, as in other large fields like Medicine. However, conducting such reviews is challenging, and we have found some current practices in the community that make it even more challenging. We propose two main recommendations moving forward.
%The process has helped us to identify several aspects that need improvement within the community. %We conclude with some recommendations and future research directions.
%[changed above to: current practices in the community that make systematic and meta-reviews difficult to conduct. This is both more specific and more constructive]
First, studies should report complete data, to allow for meaningful comparison of results. Many papers %fully reported or accurate parameters.
included only partial data on models and methods, despite the fact that results crucially depend on factors like architecture, size, tokenization, and training data. Benchmark reporting is also often partial (e.g., only 11/25 papers include BLiMP per-category results). Additionally, documenting task framing and prompting is crucial, as these can also significantly impact outcomes (e.g., zero-shot vs. few-shot).
Second, we should collectively standardize investigations to a greater extent. Differences in how syntactic phenomena are defined fragment results and limit comparability, as even the notion of ``syntactic knowledge'' varies across studies.\footnote{For instance, \citet{waldis_holmes_2024} place several of the phenomena we have considered here outside of syntax (e.g.\ negation in reasoning, and negative polarity item licensing in semantics).}
We offer a typology of syntactic phenomena that has been built in a bottom-up fashion from the current literature, and which could be refined and expanded upon as the field grows.

Standardization is particularly important for cross-linguistic research. We have shown a huge variance in cross-linguistic performance, only partially accounted for by factors like the digital support of a language or the type of linguistic phenomenon. One of the main drivers, presumably, are the characteristics of the specific data the models are being tested on, but those can vary; compare for instance determiner-noun number agreement (easy) versus long-distance subject-object agreement (hard).
Therefore, a standardization effort needs to be made in creating multilingual benchmarks, beyond direct translations or adaptations of English datasets \cite[e.g., as in][]{baucells-etal-2025-iberobench,jumelet:etal:2025}, as we otherwise risk overlooking language-specific phenomena. %Thus, we need benchmarks that are both comparable and rich enough to account for linguistic variation.
As an example of a ``best of both worlds" approach that both standardizes and allows for variation, the Spanish variant of SyntaxGym \cite{perez-mayos:etal:2021} incorporates language-specific features like subjunctive mood and flexible word order besides English-compatible phenomena.

\paragraph{The state of the field.}
Our review reveals a rapidly developing area of research, with plenty of studies and a wide variety of methods (337 papers and more than 3,000 individual results are included in our database).
%The literature also reveals a substantial and successful transfer of theoretical and methodological insights from psycholinguistics into LLM interpretability research. This is a particularly encouraging development, as it enables more principled comparisons between human and model behavior and strengthens the connection to linguistic theory.
The literature also shows substantial transfer of theoretical and methodological insights from psycholinguistics to model interpretability, enabling more principled comparisons between human and model behavior and strengthening links to linguistic theory.

However, on the negative side, we find an over-study of a single language (69\% of the articles are on English only, and 91\% include English) and a few models (30\% of the results concern BERT; including BERT and its variants raises the total to 58\%),
which severely limits the robustness and generality of the conclusions.
In addition, while formal syntactic phenomena such as agreement, syntactic structure, word order, and morphosyntactic properties such as part of speech have been well studied,
the syntax–semantics interface remains comparatively underexplored: phenomena such as binding, coreference, negation, ellipsis, or filler-gap dependencies have received very little attention. Expanding research to cover the syntax-semantics interface is particularly necessary given that we have found that in general models perform worse at it than at more formal aspects of syntax.
%This is especially relevant given that, based on the partial data reviewed above, TLMs seem to perform worse on phenomena at the syntax-semantics interface than on formal syntactic phenomena connected to syntactic function and structure.  Based on our analysis, thus, we tentatively conclude that formal syntax including morphosyntax is easier for models than the syntax-semantics interface, and look forward to future analyses of our database expanding on ours.
%\id{I think that the paragraph above and the next one overlap a bit in terms of the information discussed (particularly when it comes to the syntax-semantics interface conclusion. It could be organized so that we shortly recap what's out there in the field and move everything else to the discussion after the question. }

At a methodological level, the field is characterized by a mix of established methods (behavioral, probing) and emerging approaches (under the broad umbrella ``mechanistic'').
Overall, however, we find an overuse of purely observational methods (81\%), at the expense of causal approaches that can link model behavior to internal mechanisms.
While we encourage the use of interventional methods, we emphasize that no single method is sufficient in isolation, and that the strongest evidence ultimately comes from combining observational and interventional paradigms \citep[e.g.,][]{simon2024a,agarwal:etal:2025,acevedo:etal:2026} and comparing different architectures \citep[e.g.,][]{domenichelli:etal:2026,kryvosheieva:etal:2026}.

Furthermore, the field has mostly focused on analyzing trained models at a global level; it should pay more attention to the emergence of syntactic knowledge during training and to the connections between local and global representations and mechanisms.

\paragraph{Syntactic knowledge in TLMs}
As for what TLMs learn about syntax through the language modeling task, the evidence suggests: a whole lot. 
For instance, TLMs perform quite well on the English BLiMP benchmark (average score of 72\%, against a 50\% random baseline), and between 79 and 85\% of the analyzed studies identify syntactic knowledge in TLMs. % (recall Figs.~\ref{fig:RQ3probing} and ~\ref{fig:RQ3mechanistic}).
Even accounting for the possible effect of publication bias (by which we mean here the general tendency for positive findings to be published more often than negative ones), we take the literature to show that TLMs do encode a non-trivial amount of syntactic knowledge.

As for factors influencing performance, as mentioned above, we find that models perform better at formal syntactic phenomena than phenomena close to the syntax–semantics interface. As could be expected, larger models and models trained with more data tend to perform better; and performance degrades for lesser-resourced languages. Importantly, we also find that models that perform well on some syntactic phenomena tend to perform well across others, suggesting a general syntactic ability rather than phenomenon-specific strengths.
As for internal mechanisms, probing data supports the view that much of syntactic processing occurs in the middle layers \cite{lopezotal:etal:2025}. However, there is still limited mechanistic understanding of why such representations emerge or how they support downstream computation.
%Given the heterogeneity of the field, we leave to future work the summarization of other mechanistics insights.
Given the current heterogeneity of the field, and that mechanistic methodologies and reporting practices are still evolving, we leave a broader synthesis of mechanistic insights to future work.

To sum up, the current literature supports the view that TLMs possess remarkable syntactic capabilities. However, understanding how this knowledge is acquired and used, as well as how it relates to universality and variation in language, will require more work in both the analysis of existing literature and new studies. We encourage greater methodological harmonization, more explicit theoretical commitments, and a broader empirical landscape. Some open topics are whether TLMs implement near-symbolic mechanisms, as hypothesized by \citet{boleda:2025}; when and how TLMs fall back on superficial statistical patterns \cite[e.g.\ performance on agreement, binding, or scope often degrades for rare or nonce lexical items,][]{lasri_does_2022}; %maudslay_syntactic_2021,
or whether improvements with model scale reflect genuine generalization or massive memorization of specific instances \cite{Lake:Baroni:2017}. We hope that our review, database, and annotation standard contribute to moving the field forward.
\section*{Acknowledgments}
We thank the members of the UPF COLT group for feedback. Our work was funded by the Catalan government (AGAUR grant SGR 2021 00470). NG also received the support of a fellowship from Fundación Ramón Areces. GB also received the support of grant PID2020-112602GB-I00/MICIN/AEI/10.13039/501100011033, funded by the Ministerio de Ciencia e Innovación and the Agencia Estatal de Investigación (Spain).
This paper reflects the authors’ view only, and the funding agencies are not responsible for any use that may be made of the information it contains.

\section*{Limitations}
\label{sec:limits}
% - focus not only on Transformer-based but also recurrent architectures
% - extend findings of other blimp benchmarks? iterate over papers that cite them (feasibility); future work to extend to other languages
% - DB is ongoing/growing. We identified relevant studies after finalizing analysis. The database is suppose to be interactive for the field and extended.
For feasibility, this study focuses on Transformer-based language models and textual input, thus our findings may not fully generalize to recurrent or other neural architectures. Also, we conducted the keyword search in English, which likely led to under-retrieval of publications written in languages other than English.
Note, however, that this was only one of the 3 strategies we used to identify relevant literature.
%Also for feasibility, our answer to RQ3 is based only on one benchmark (BLiMP) and one language (English). Future analyses can use the \textit{findings} variable of our database to get at a broader understanding of the patterns.
% While we build on insights from the BLiMP benchmark, we do not systematically extend or compare results across all related benchmarks or their subsequent follow-up studies; doing so is left for future work, especially including extensions to languages beyond English. \gb{Leaving it commented out because imo it is subsumed by the previous line}
%\no{We originally set out to map what is known about how TLMs process syntax on the basis of the whole database; however, this goal proved unrealistic, given the daunting conceptual and methodological variation across studies (see Section~\ref{sec:conclusions} for discussion). Responding to these difficulties, and in order to keep our review objective and quantitative, we chose to focus on the results obtained with the BLiMP benchmark and behavioral accuracy scores, which allows for systematic comparisons, and for which we have enough datapoints across different languages.}
Finally, our database is necessarily incomplete and evolving; the resource is intended to remain interactive and expandable as the field progresses. We see this work as a foundation that invites continued contributions and updates from the research community.

\bibliography{nora-import,SyntaxInLLMs}

\appendix
% \section{Appendix}

\section{References}\label{app:references}
Table~\ref{tab:ref} lists the studies included and analyzed in our database. The entries are organized first by publication year and then sorted alphabetically by the authors’ names.
\input{table_refrences}

\section{Annotation dimensions}\label{Appendix:annotations}
This section gives more details on the annotation dimensions used in the database and lists the specific \textsc{column name}. Additional information on variable levels and coding schemes are publicly available in the repository’s README. 

\subsection{Meta-Data} %
%\gb{transform the main itemize into} 
%\begin{verbatim}
%\paragraph
%\end{verbatim}
%\gb{this way we make more compact (subsubnesting doesn't go well with 2 columns)} 
%\texttt{id, short\_ref, url, year, title, abstract}
%\textsf{id, short\_ref, url, year, title, abstract}

We publish the main information about the study (\textsc{id, short\_ref, url, year, title, abstract}), and information of its review process:
\paragraph{Source for the article:} \textsc{review}, indicates where the article was sourced from with chronological line of adding papers. Tracks the order of inclusion from different sources. 
%\begin{itemize}
    %\item If it comes from the main review papers, it is labeled as such (A--F).
    %\item Otherwise, it states the spreadsheet tab (X, M, or additional search S).
%\end{itemize}
\paragraph{Exclusion grounds:} \textsc{exclusion\_grounds},  %(closed list)Number 0 
indicates the reason for exclusion.  %and letters to indicate reasons for exclusion:  A -- architecture: not transformer (e.g., LSTM/RNN); B -- task: MT-task; C -- publication type: excluding MA/PhD theses, any article without some empirical component); D -- do not empirically assess or analyse syntactic structure or syntactic knowledge (\textit{Can we learn something about how LMs learn and treat syntax?}). 

\subsection{Model-related information}

\paragraph{Model type:} \textsc{model\_type}, categorized as:
\begin{itemize}
        \item \textbf{Bidirectional:} At least one bidirectional model is featured (e.g., BERT, LSTMs).
        \item \textbf{Causal:} At least one causal model is featured (e.g., GPT, LSTMs).
        \item \textbf{Both:} Both model types are featured (e.g., comparative experiments with BERT, GPT, and LSTM).
        \item \textbf{Other:} Cases with neither causal nor bidirectional models, either excluded or meeting special inclusion criteria, e.g., specific analysis of encoder component of a Transformer architecture. %\id{comment: clarify this last point or give an example as otherwise it is quite confusing}
\end{itemize}
    
\paragraph{Model name and post-processing:} \textsc{model\_name\_comment}, for the comment string, as specific as possible it is stated in the paper.  \textsc{model\_name}, we post-process this string into a list of dictionaries, regard the following example: 
    \begin{enumerate}
        \item Name as reported in the paper, e.g., \textit{bert\_base\_uncased}. 
        \item Name enriched with post processing script: e.g., \textit{family} =``BERT'' plus mention of size, BERT-base.
        \item Final example: list(\{\textit{string} =``bert\_base\_uncased'', \textit{family} =``BERT'', \textit{size} = base, \textit{case} = uncased\}).
    \end{enumerate}

\paragraph{Fine-tuned:} \textsc{fine\_tuned}, indicates whether fine-tuning was performed. %Instructed or domain-specific models are excluded, but models with continual pretraining are included (e.g., RuBert, GalBert). \gb{Explain why} \n{discuss, since this is not the case for 51 instruct models: Instructed or domain-specific models are excluded}

\paragraph{Language coverage:}\textsc{model\_language\_coverage}
    \begin{itemize}
        \item \textbf{Monolingual:} One language from scratch or continually pretrained (e.g., GPT2, BERT, ruBERT). %\gb{also rubert adn galbert?}
        \item \textbf{Multilingual:} Trained on multiple languages simultaneously from scratch (e.g., mBERT, GPT3).
        \item \textbf{Both:} Studies comparing monolingual and multilingual models.
    \end{itemize}
    
%\subsection*{Training data} \gb{Training data is part of model information, no?}
%\paragraph{Size (open list):} Recorded as reported; if unavailable, coded as ``unavailable data''. For well-documented models, information may be filled via post-processing. Represented as a dictionary, e.g., \{\texttt{mymodel: 83KB, mymodel2: 87KB}\}.

%\paragraph{Type (open list):} Recorded as reported (e.g., online corpora, acquisition data). For well-documented models, data may also be obtained via post-processing.

\subsection{Experimental design}
\paragraph{Syntactic phenomena:}

\renewcommand{\arraystretch}{1.2}
\begin{table*}[t]
\centering
\small
\begin{tabular}{|p{3cm} p{12.5cm}|}
\hline
\textbf{Phenomena} & \textbf{Description} \\
\hline

\rowcolor{white!45} Benchmarks  %OIReddishPurple
& Aggregated evaluation suites combining multiple syntactic phenomena (e.g.\ BLiMP, CoLA, SyntaxGym), often reporting composite performance metrics rather than phenomenon-specific analyses. \\

\rowcolor{cyan!35}Dependency parsing \& constituency 
& Structural representations of sentence form, including dependency relations, constituency trees, and syntactic functions (e.g., subject vs.\ object), without targeting hierarchical depth or processing complexity. \\

\rowcolor{blue!30}Hierarchical structure
& Properties of abstract hierarchical organization independent of specific formalisms, including tree depth, recursion, center embedding, attachment ambiguity, and structural complexity. \\

\rowcolor{blue!15}Word order 
& Constraints on the linear ordering of words and phrases, including canonical and non-canonical orders, scrambling, and  variation in surface order. \\ 

\rowcolor{OIYellow!15}Agreement \& feature-checking 
& Morphosyntactic dependencies involving grammatical features such as number, gender, person, and tense, typically requiring local or semi-local syntactic constraints. \\

\rowcolor{OIYellow!35}POS \& other lexical properties 
& Lexical-level categories and features, such as part of speech, verbal subcategorization, or lexical aspect; excludes tasks like named entity recognition that are primarily semantic or discourse-driven. \\

\rowcolor{orange!25}Negation \& NPIs 
& Licensing and interpretation of negation and negative polarity items, involving interactions between syntax, semantics, and scope. \\

\rowcolor{OIBluishGreen!15} Binding \& coreference 
& Constraints on anaphora and antecedent resolution, including reflexives, pronouns, and coreference relations, typically requiring hierarchical and discourse-sensitive representations. \\

\rowcolor{OIBluishGreen!35}Ellipsis \& filler--gap \mbox{\textnormal{dependencies}}
& Long-distance dependencies such as wh-movement, gap filling, and ellipsis resolution, requiring sensitivity to abstract structure beyond linear proximity. \\
%\rowcolor{OIBluishGreen!35} dependencies & sensitivity to abstract structure beyond linear proximity. \\

\rowcolor{brown!25}Typology 
& Cross-linguistic variation in syntactic patterns and constraints, including agreement systems, and constructional differences across languages. \\

\rowcolor{red!20}Argument structure \& constructions 
& Systematic syntactic patterns linking form and meaning, including argument realization, alternations (e.g.\ active/passive, dative alternation), relative clauses, and other constructional schemas at the syntax--semantics interface. \\

\hline
\end{tabular}
\caption{The 11 Linguistic phenomena of our survey with brief definitions, which we annotated and organized from general Benchmark, to those focusing on syntactic structure and surface-level form (blue-purple tones) to form-and-function-related categories. %Linguistic phenomena of our survey, with brief definitions.
}
\label{tab:phenomena}
\end{table*}
%\gb{I think it'd be best to put a table with the linguistic phenomena + explanation for each of them. Also provide a brief explanation with the motivation + rationale for the specific division of phenomena that we went for (is anything there that is controversial? if so, justify --- e.g.\ without an explanation I don't understand why constituency is classified differently from hierarchical structure).}

\begin{itemize}
        %\item \textbf{Even more general:} \textit{Added later via automatic matching} \gb{deprecated?}: 
        %\begin{enumerate}
        %\item Morphosyntax, 
        %\item Syntax-Semantic Interface, 
        %\item Naturalistic (anything goes?), 
        %\item General benchmark,
        %\item Misc? Overarching?
        %\end{enumerate}
       
        \item \textbf{General:} \textsc{syntactic\_phenomena}, bottom-up annotation, starting from general to more specific cases. See Table~\ref{tab:phenomena}.
        
\item \textbf{Specific:} \textsc{specific\_phenomena}, as referred to by authors.
\end{itemize}

\paragraph{Languages tested and post-processing:} \textsc{languages\_tested\_comment}, for the comment string, as specific as possible it is stated in the paper. \textsc{languages\_tested}, we post-process this string into a list of all languages used in the experiments. \textsc{languages\_tested\_eng\_included}, indicates whether English was tested. \textsc{languages\_tested\_monolingual\_eng}, indicates whether only English was tested. \textsc{languages\_tested\_lang\_count}, number of tested languages.
\paragraph{Fine-tuning materials:} \textsc{fine\_tuning\_materials}, resources used for fine-tuning in the experiments.
\paragraph{Experimental materials:} \textsc{experimental\_materials\_comment}, materials as referenced by authors and summarized through AI-based workflow. \textsc{experimental\_materials}, annotated in post-processing to indicate whether they are naturalistic language or controlled stimuli, both, or ``unknown'' (if under-specified/referring to other's paper dataset).
    
\subsection{Interpretability techniques}
\paragraph{General:} \textsc{interp\_category}, categorized as:
    \begin{itemize}
        \item \textbf{Behavioral:} Methods based on last layer input--output behavior.
        \item \textbf{Probing:} Methods training a classifier on top of a given TLM layer.
        \item \textbf{Mechanistic:} Methods investigating internal dynamics, may incorporate behavioral or probing elements.
    \end{itemize}
\paragraph{Specific:} \textsc{interp\_method}, specific techniques description, named as in the original paper.

\paragraph{Probing:} \textsc{probing}, classification of results to claim evidence for syntactic representations in the TLM.
\paragraph{Location:} \textsc{probing\_location}, specific location article states where performance peaks within the model.

\paragraph{Mechanistic Interpretability:} We use the taxonomy introduced by \citet{bereska2024mechanistic} to characterize 4 dimensions. We annotated each mechanistic paper, 140 in total, along these dimensions: %, including method, causal nature, analysis phase, and locality.
    \begin{itemize}
        \item \textbf{Method:} \textsc{mechanistic\_method}, string as general as possible as it is stated in the paper.
        \item \textbf{Causal Nature:} \textsc{mechanistic\_causal nature}, ranges from purely observational to interventional approaches.
        \item \textbf{Phase:} \textsc{mechanistic\_phase}, the phase distinguishes post-hoc from intrinsic approaches applied during model development.
        \item \textbf{Locality:} \textsc{mechanistic\_locality}, describes the level of analysis, ranging from individual components (e.g., neurons, attention heads) to full model architectures (e.g., residual stream).
    \end{itemize}
    
\paragraph{Syntactic Awareness:} \textsc{Mechanistic\_Syntactic evidence/ awareness} captures whether mechanistic interpretability studies claim evidence of syntactic awareness of TLMs.

%\subsection{Findings, withhold?}
%\paragraph{\no{Main findings on syntactic knowledge:}\textsc{main\_findings}, summarizes the findings of studies on how TLMs handle syntax, and their syntactic knowledge.}
%\gb{I would prefer not to include it in the paper, because that's what we said we would do in our response to the reviewer. We can include it in the repository as an additional piece of information and document it there.}

\section{AI-Based Workflow for Information Extraction}\label{app:ai-flow}

\begin{table*}[!ht]
  \centering
  \begin{tabular}{|p{4.5cm}|p{10.5cm}|}
    \hline
    % Model-related information: 
    \textbf{Category} & \textbf{Question}\\
    \hline 
    \textbf{Model information}
                      &\\
    \textsc{model\_type}%name/size, fine-tuned
                      &\textit{\small Which model do they test and which languages, do they do finetuning?}\\ 
    \hline
    \textbf{Experimental design}
                      &\\
    \textsc{syntactic\_phenomena}
                      &\textit{\small What are the ``Syntactic phenomena, the Specific: list set as referred to by authors''}\\ 
    \textsc{experimental\_materials}
                      &\textit{\small What is the ``Experimental materials: naturalistic language, controlled stimuli''} \\
                      %``syntactic knowledge''
    \hline
    \textbf{Interpretability}&\\
     \textbf{techniques}&\\% Interpretability techniques
    \textsc{interp\_method} & \textit{\small What is the Specific method name as referred to by the authors?}  \\
    %\hline
    %\textbf{Findings}&\\
    %\textsc{main\_findings} on & \textit{\small What are the main findings, especially on syntactic knowledge in the LM?} \\
    %syntactic knowledge & \\
    \hline
  \end{tabular}
  \caption{Questions used for AI-based information extraction.}
  \label{tab:ai_questions} 
\end{table*}

During the review process, we developed a semi-automated strategy combining NotebookLM \citep{google:2023:notebooklm} and ChatGPT \citep{chatgpt5:2025} for structured information extraction on 4 variables, \textsc{model\_type}, \textsc{syntactic\_phenomena}, \textsc{experimental\_materials}, and method name (\textsc{interp\_method}), listed in Table~\ref{tab:ai_questions}.  %\no{and additionally to the annotations another field \textsc{main\_findings},\footnote{We created an AI-summary \textit{findings} that summarizes, in prose, the main insights regarding syntactic competence and processing as well as any model comparisons, which is available upon request.} \gb{again, I would include this info in the repo only, to stick to what we promised} listed in Table~\ref{tab:ai_questions}.}

\begin{figure*}[ht]
    \centering
    % First subfigure
    \begin{subfigure}{0.8\textwidth}
        \centering
        \fbox{\includegraphics[width=\linewidth]{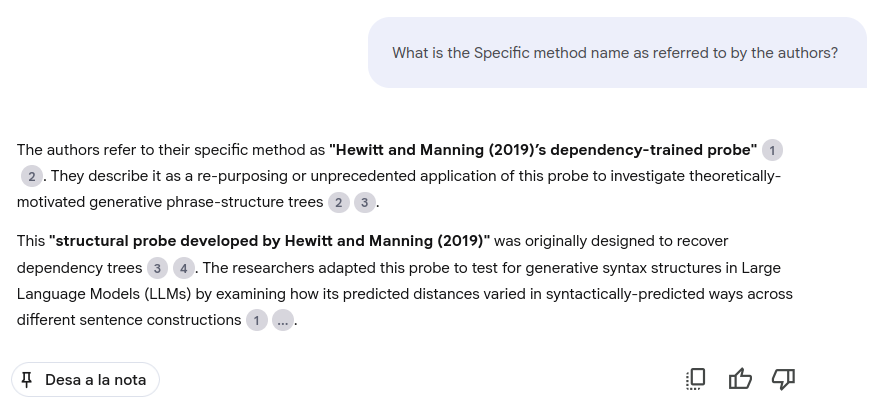}}
        %\caption{}
        \label{fig:subfig1}
    \end{subfigure}
    \hfill
    % Second subfigure
    \begin{subfigure}{0.8\textwidth}
        \centering
        \fbox{\includegraphics[width=\linewidth]{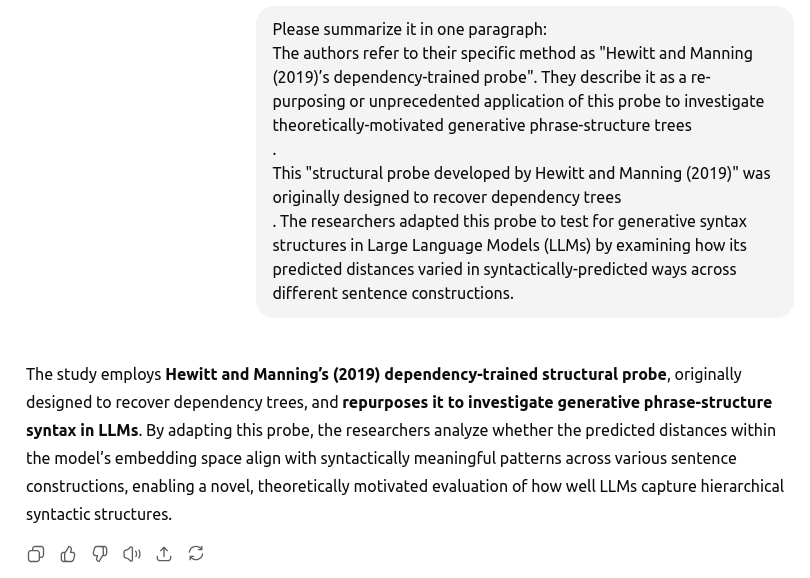}}
        %\caption{}
        \label{fig:subfig2}
    \end{subfigure}
    \caption{Illustration of the AI-based workflow. \textbf{Top:} NotebookLLM \citep{google:2023:notebooklm} interaction depicting how the model assists with information extraction; numbers refer to the original source in the paper.  \textbf{Bottom:} Example of a ChatGPT \citep{chatgpt5:2025} session showing a query and the generated response.}
    \label{fig:notebookllm_gpt}
\end{figure*}

Each paper was first analysed using NotebookLM \cite{google:2023:notebooklm} by posing a series of targeted questions 
%(listed in Table~\ref{tab:ai_questions} in App.~\ref{app:ai-flow}) 
to extract relevant details for these categories. The responses generated by NotebookLM were then summarized into a coherent paragraph using ChatGPT \cite{chatgpt5:2025},%
\footnote{August 2025 release, incorporating GPT-4 and GPT-5 models.} 
which we subsequently reviewed and used to populate the database. %We specifically chose to prompt for paragraph-style summaries for the \textsc{main\_findings} field, to maintain compact, integrated information for the database, rather than fragmented notes. 
% add an example with the answers from NotebookLM and the paragraph by ChatGPT. Also, why did you need the paragraph? why were the answers not enough?
See Fig.~\ref{fig:notebookllm_gpt} for an example. The workflow was iteratively refined during the course of the review process until it stabilized; starting at paper ID 120, this approach was applied without further adaptation. %\no{Following the validation of our approach, as described in the next section, we decided to withhold the \textsc{main\_findings} field of the database, exclusively with LLM-generated output.}

\subsection{Validation of the Approach}
%\subsection{Inter-annotator agreement} \gb{I would put this information at the beginning of the previous subsection ("Validation of the approach") --- We don't compute inter-annotator agreement because it's not adequate in this case; we do the following types of validation}
The annotations in the database did not involve subjective judgments or interpretation, but rather the extraction of explicitly stated factual information from each paper (e.g., the names of the models tested or the languages investigated). For this reason, we do not believe that multiple annotation or inter-annotator agreement would meaningfully improve reliability. We conducted the following types of validation.

The AI-assisted workflow was validated differently depending on the variable. For the \textsc{model\_type} variable, NotebookLM was used only to locate the relevant section of each paper, after which the information was manually extracted and verified. Likewise, although the \textsc{syntactic\_phenomena} variable was initially populated using the AI-assisted workflow, all entries were manually checked and corrected during the annotation of the general syntactic phenomena. These two variables therefore underwent complete manual validation.
%The LLM-based validation was performed only for the \textsc{main\_findings} variable. Because the use of LLM-generated summaries remains contentious, this variable has been omitted from the publicly released dataset; a version including it is available upon request.}\gb{so you don't want to release it?}

For the \textsc{experimental\_material} and \textsc{interp\_method} variables, we manually inspected a random sample of 50 papers annotated using the AI-assisted workflow. Annotations were scored as 0 (incorrect), 0.5 (correct but incomplete or superficial), or 1 (correct). No annotation received a score of 0. The resulting accuracy was 97\% for \textsc{experimental\_material} and 93\% for \textsc{method}, and all identified issues were corrected in the final database.

\subsection{Limitations of Workflow}
Some linguistic phenomena referenced in articles pertained to related work rather than the studies' own experiments, which can introduce ambiguity in AI-assisted summaries.
% syntactic phenomenas might refer to related work rather than what was tested in experimental work. 
During the manual annotation of the general syntactic phenomena, we identified and corrected a few such cases, but minor residual errors %\id{rephrase: errors? rather than ambiguity} 
may remain. %Though we opened 70\% of the paper for the annotation. 

%\subsection{Illustrative Figures}
%We provide screenshots from the workflow and interactions with the models in Fig.~\ref{fig:notebookllm_gpt}. %and Fig.~\ref{fig:subfig3}.

\section{Additional Results to Section 3}\label{app:sec_3}

  \begin{figure*}[!ht]
        \centering
        \subfloat{\includegraphics[trim=0.1cm 0.1cm 0.1cm 0.1cm, clip, scale=0.35]{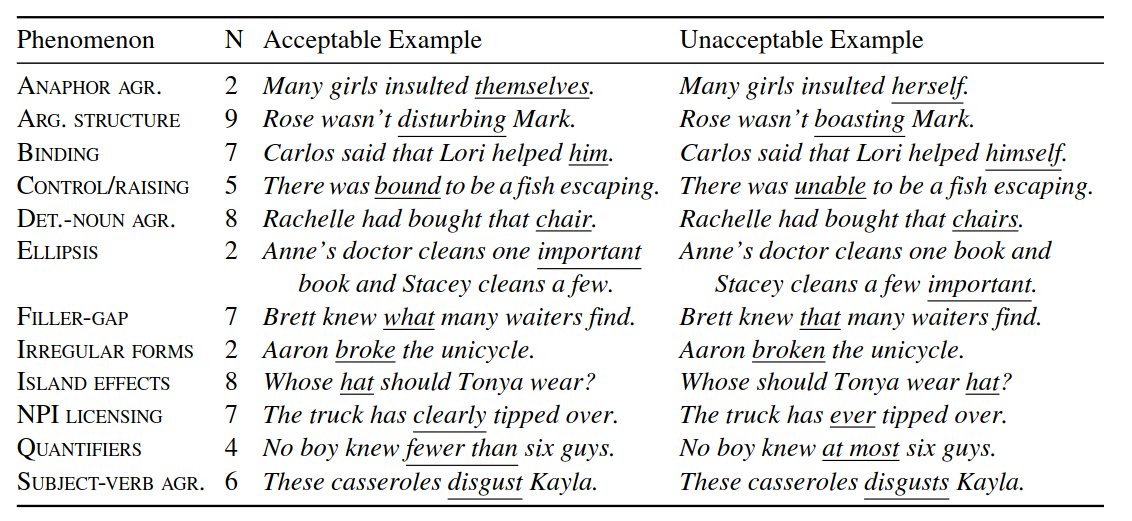}}
        \caption{Table from \citet{Warstadt:etal:2020}; minimal pairs from each of the 12 linguistic phenomenon categories covered by BLiMP. Column \textit{N} records the number of different subphenomena exemplifying the relevant general phenomenon, 67 in total.}
        \label{fig:blimp_cats}
    \end{figure*}

\subsection{Additional Results to the Overview} \label{app:general}
\begin{figure}[!ht]
        \centering
        \subfloat{\includegraphics[trim=0cm 0cm 0cm 0cm, clip, scale=0.69]{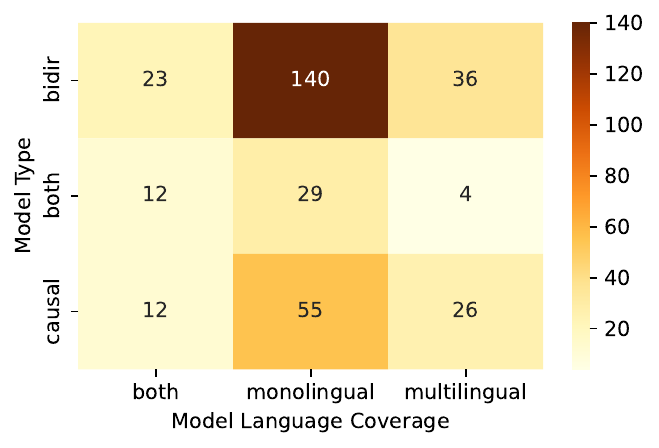}}
        \caption{\small Count distribution by model type and language coverage.}
        \label{fig:general_heatmap}
\end{figure}

We plot the coverage of model type and TLMs language coverage in Fig.~\ref{fig:general_heatmap}. Most work focuses on a single model paradigm, primarily bidirectional models (59\%) or causal models (28\%), with only a small fraction directly comparing both (13\%); similarly, evaluations largely target monolingual TLMs (66\%), while multilingual (20\%) and both monolingual–multilingual settings (14\%) remain underexplored.

\begin{figure}[!ht]
  \centering
  \subfloat{\includegraphics[trim=0.25cm 0.25cm 0.1cm 0.21cm, clip, scale=0.55]{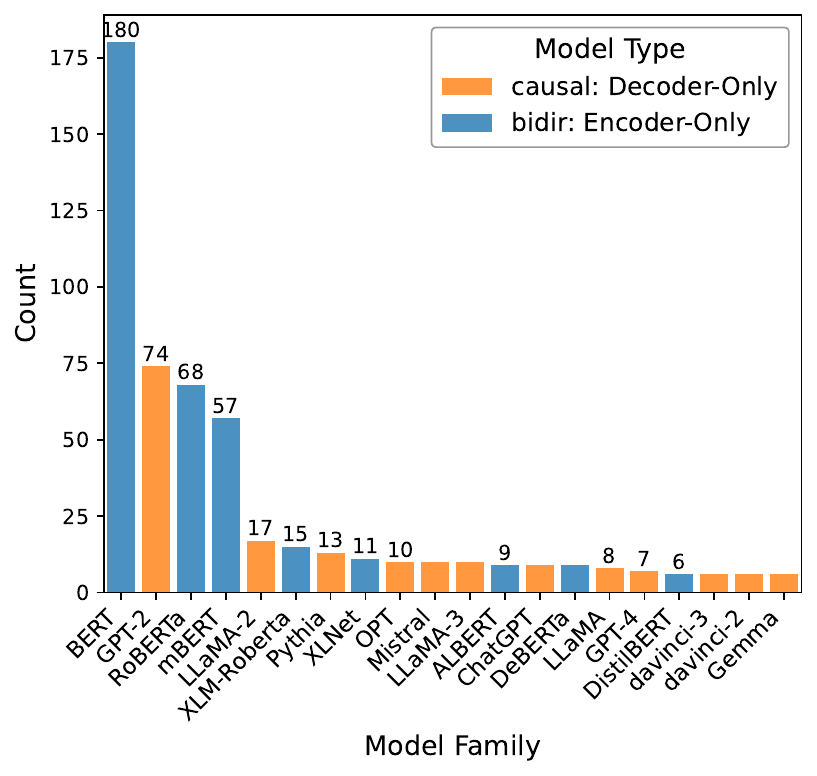}}
  \caption{The 20 most frequently used TLMs. TLMs are counted once per study and grouped by family or model name when available (e.g., one BERT count for studies comparing BERT-base/large/(un)cased). Overall, the database includes 54 unique TLMs and 609 non-duplicate instances across studies, as illustrated above.
  } 
  %TLMs are counted once per study and grouped by their family or specific model name when available (e.g., one BERT count for a study that compares BERT-base/large/(un)cased). Overall, the database features 54 different TLM names and 609 instances without duplicates per study, exemplified previously.}
  \label{fig:overview_app}
\end{figure}

We identified 54 unique TLMs and 609 non-duplicate model instances in the database; models were counted once per study and grouped by family or model name when appropriate (e.g., studies comparing BERT-base, BERT-large, or cased/uncased variants contributed a single BERT count). Fig.~\ref{fig:overview_app} shows that most of the results in our database focus on a single model, BERT, which accounts for 30\% of all TLMs results. BERT and its variants together represent 58\% of the datapoints, while the four most extensively studied models account for 62\% overall.

\subsection{Additional Results to Linguistic Phenomena}\label{app:fig_syntacticPhenomena}
\begin{figure}[ht]
        \centering
        \subfloat{\includegraphics[trim=0cm 0cm 0cm 0cm, clip, scale=0.75]{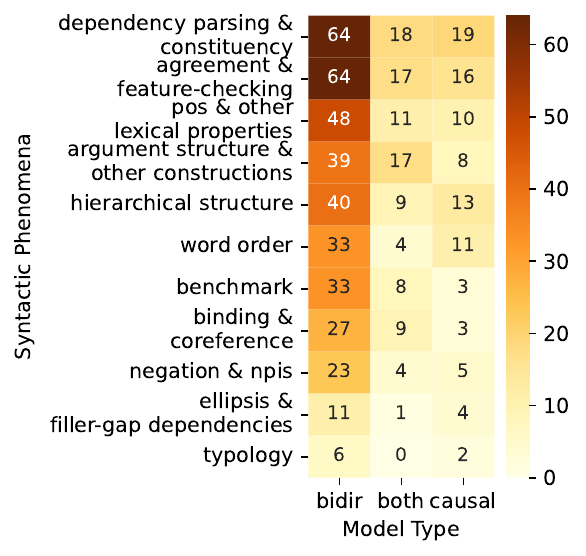}}
        \caption{Heatmap of counts illustrating model type and syntactic phenomena.}
        \label{fig:RQ1phenomena_heatmap}
\end{figure}

\begin{figure*}[!ht]
    \centering
    %\subfloat{\includegraphics[scale=0.43]{figures/method_years.png}}\\
    \subfloat{\includegraphics[trim=0cm 0cm 0 0cm, clip, scale=0.5]{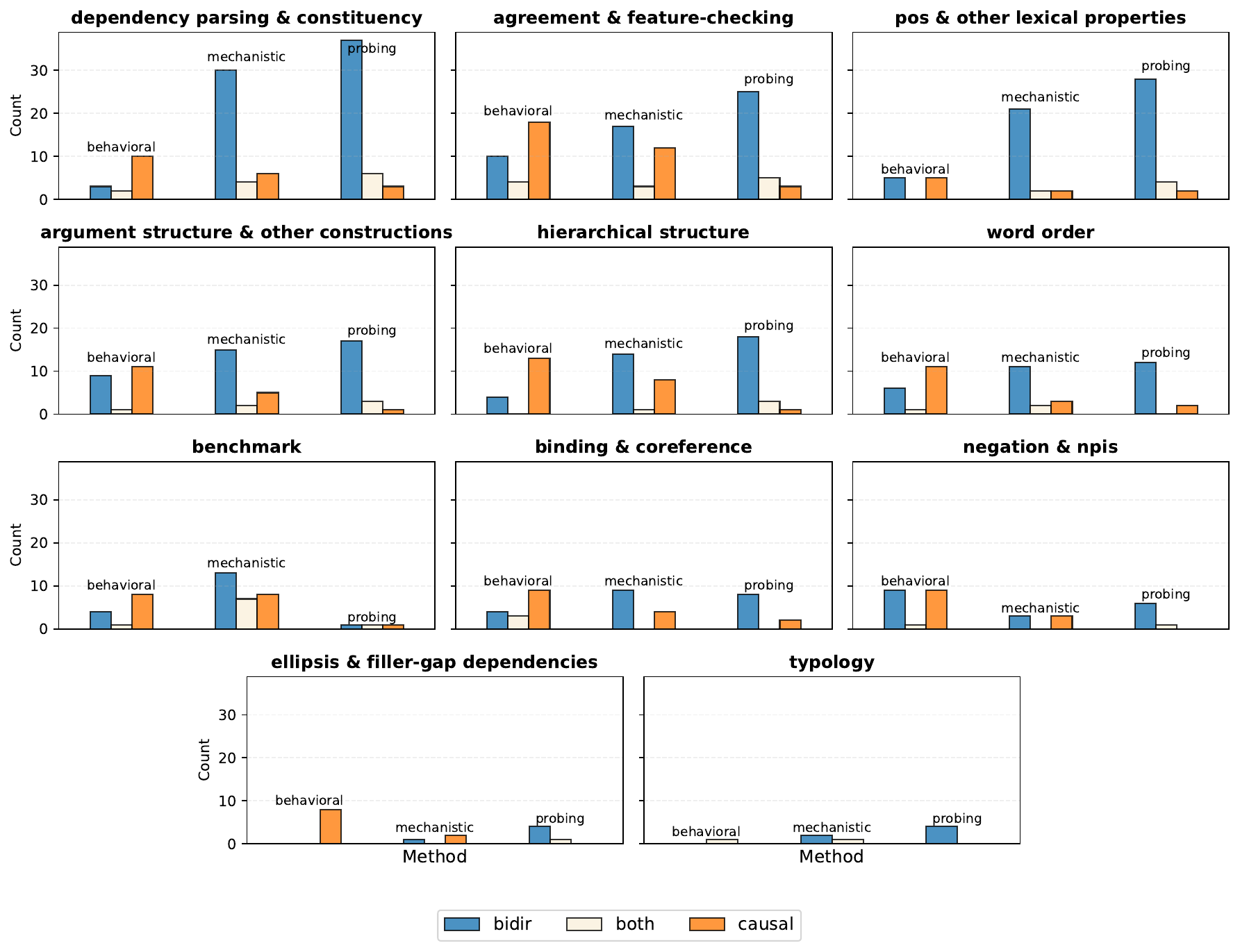}}
    \caption{Distribution of studies in our database across syntactic phenomena, model type (bidirectional, causal, both TLMs types) and interpretability method, ordered by overall counts. 
    %Placeholder, just present syntactic phenomena trend? \textbf{Top:} . \textbf{Bottom:} .
    }\label{fig:phenomXmethodXmodel}
\end{figure*}

\paragraph{Model type \& language.} Regarding coverage across TLMs, the distribution varies notably across syntactic phenomena, with bidirectional TLMs dominating most phenomena (ranging from ${\sim}60$\% to 73\%) evaluation, while causal and both model types were more often tested on phenomena like \textit{POS and other lexical properties} and \textit{dependency parsing and constituency} (${<}20$\%). See Fig.~\ref{fig:phenomXmethodXmodel} for a distribution of studies across syntactic phenomena, model type, and interpretability method. 
When looking at the composition of phenomena within each model type, Fig.~\ref{fig:RQ1phenomena_heatmap}, \textit{agreement and feature-checking} and
\textit{dependency parsing and constituency} constitute the majority across all TLM tasks. Phenomena such as \textit{argument structure and other constructions} are more prominent in causal TLMs. %, whereas phenomena such as \textit{dependency parsing and constituency} and \textit{pos} are more prominent in \texttt{both} and \texttt{causal} model type. conversely 

\begin{figure}[ht]
        \centering
        \subfloat{\includegraphics[trim=0.2cm 0.2cm 0.2cm 0.2cm, clip, scale=0.6]{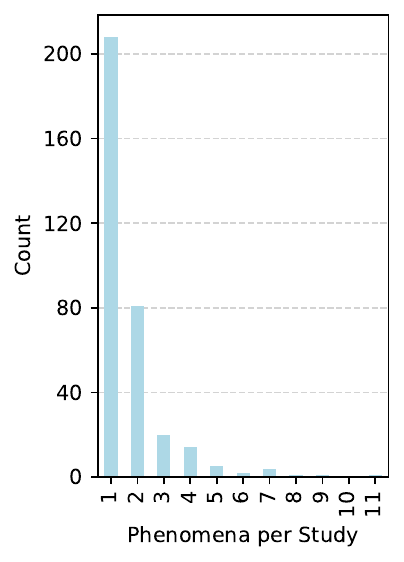}}
        \caption{Counts of syntactic phenomena investigated per study.}
        \label{fig:RQ1phenomena_counts}
    \end{figure}

Moreover, 62\% of the articles focus on only one syntactic category, with 24\% testing two, Fig.~\ref{fig:RQ1phenomena_counts}.
This reflects a common emphasis on individual syntactic phenomena, which should not be interpreted as a limitation in itself. Rather, progress in the field benefits from combining fine-grained, phenomenon-specific analyses with more holistic approaches, as there is no single or definitive test for syntactic competence.

\begin{figure}[ht]
        \centering
        \subfloat{\includegraphics[trim=0cm 0.6cm 0cm 0cm, clip, scale=0.75]{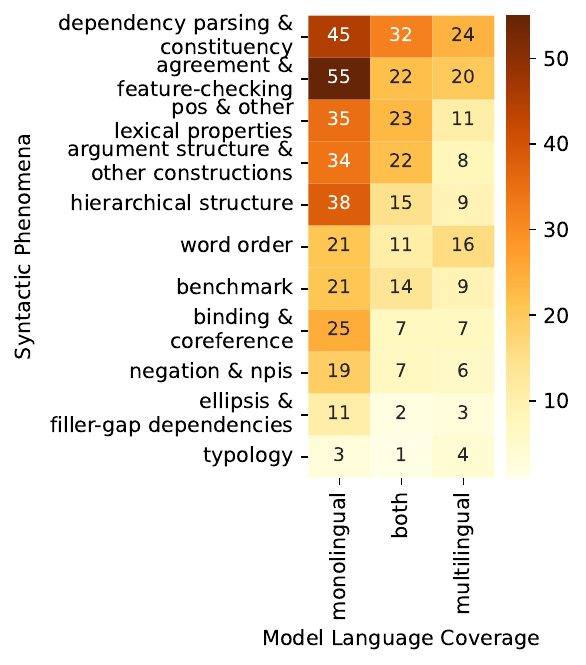}}
        \caption{Counts heatmap of model language coverage and syntactic phenomena.}
        \label{fig:RQ1phenomena_heatmap_lang}
\end{figure}

Across syntactic phenomena, monolingual TLMs dominate coverage in most categories (typically around ${\sim}50$–65\%), while %multilingual TLMs have notable presence in phenomena like typology (42.9%) and miscellaneous (27.6%)
studies testing both and multilingual TLMs tend to contribute moderately (usually 15–30\%) depending on the phenomenon, see Fig.~\ref{fig:RQ1phenomena_heatmap_lang}.

\subsection{Additional Results to Methods}\label{app:method}
\begin{figure}[ht]
        \centering
        \subfloat{\includegraphics[trim=0.27cm 0.09cm 0.2cm 0.99cm, clip, scale=0.53]{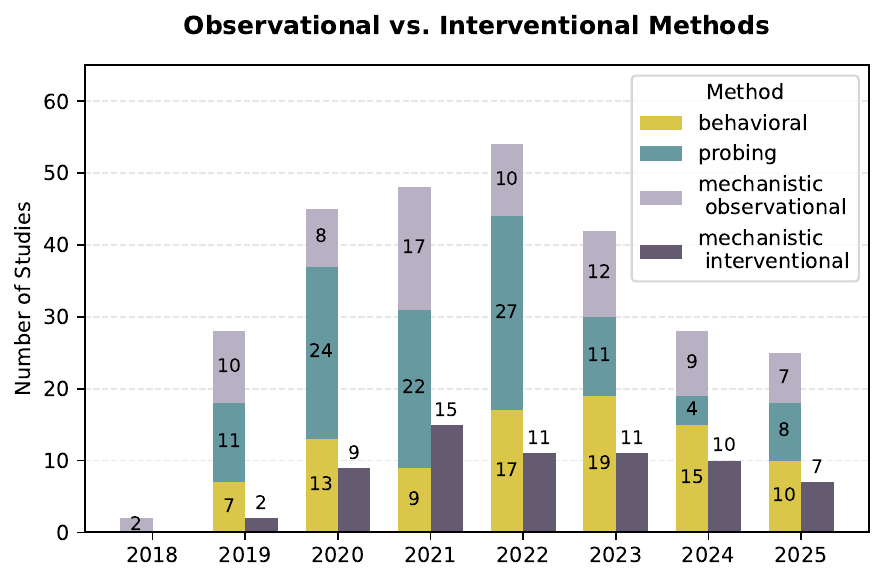}}
        \caption{Counts of methods over years, grouped by observational and interventional approaches.}
        \label{fig:method_growth_stacked_trend}
\end{figure}

Fig.~\ref{fig:method_growth_stacked_trend} shows the counts of methods over the years, categorized according to whether they are observational or interventional. Notably, a significant temporal trend is that probing methods have diminished substantially in recent years, possibly due to the growing awareness of their limitations. For instance, following the publication of the influential criticism in \citet{Belinkov:2022}, the number of probing studies with bidirectional TLMs in our database declined sharply, from 27 studies in 2022 to just 11 in 2023.

\begin{figure}[h!]
  \centering
  \includegraphics[trim=0cm 1cm 2.4cm 1.9cm, clip, scale=0.43]{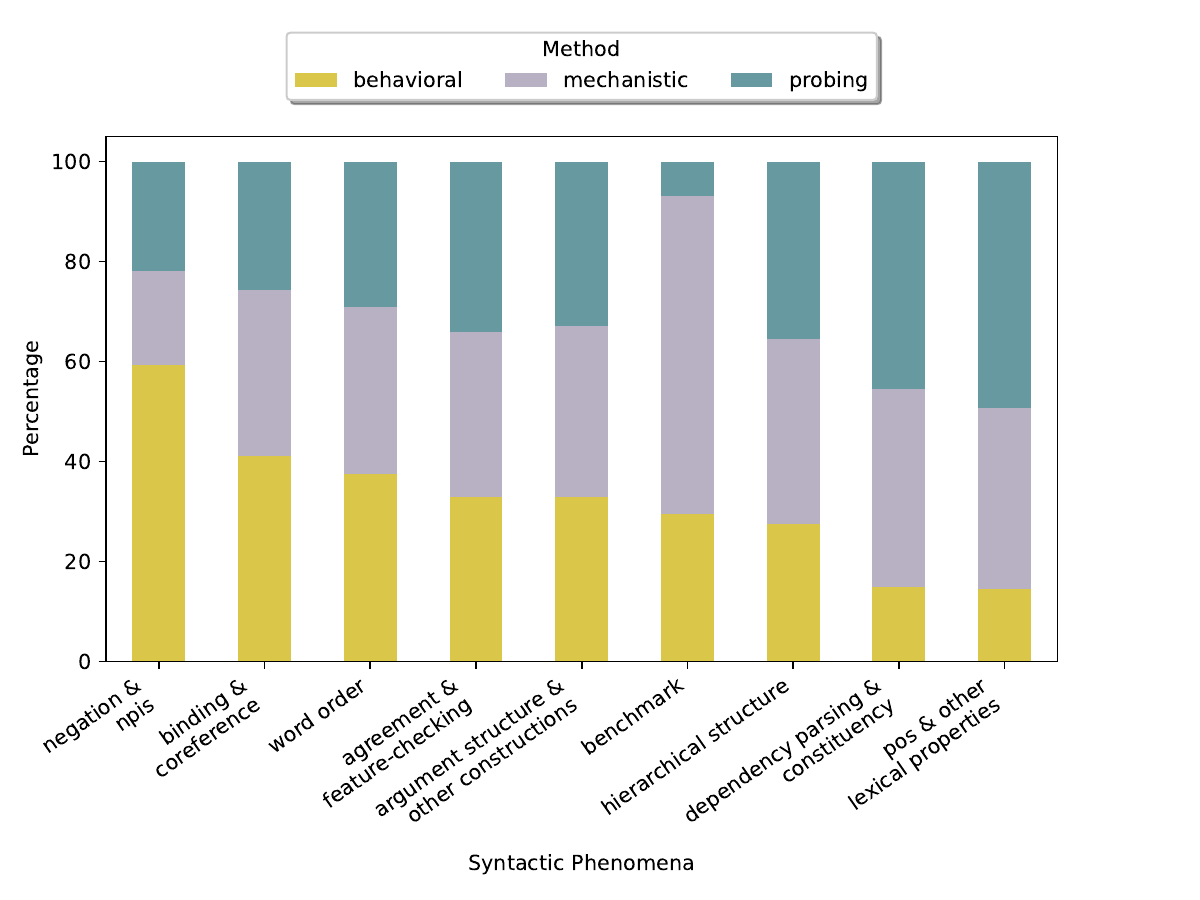}
    \includegraphics[trim=0cm 4.8cm 0cm 0.2cm, clip,  scale=0.6]{figures/method_years_percentage.pdf}
  \includegraphics[trim=0cm 0.2cm 0cm 0cm, clip,  scale=0.5]{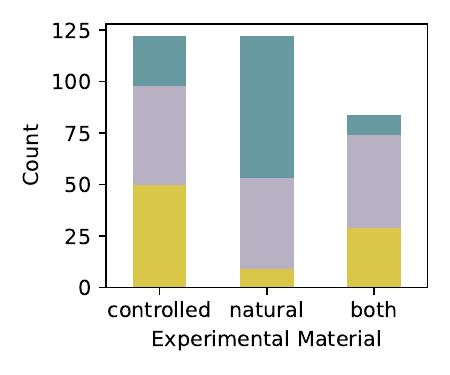}\hfill
  \includegraphics[trim=0cm 0.2cm 0cm 0cm, clip,  scale=0.5]  {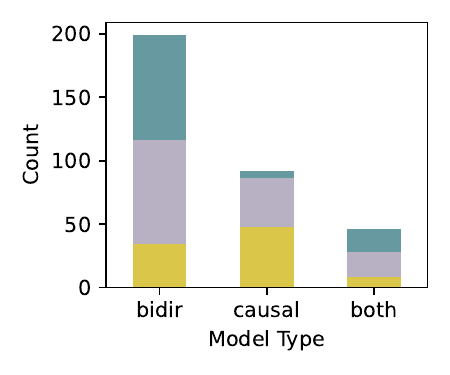}
  \caption{\textbf{Top:} Percentages of methods across syntactic phenomena; only phenomena with more than 20 instances are included. Linguistic phenomena are sorted by percentage of behavioral method. \textbf{Bottom:} Counts of experimental materials and model types by method (\textit{unknown} experimental material with fewer than 10 counts excluded).}
  \label{fig:RQ2method_phenomena}
\end{figure}

\paragraph{Linguistic phenomena.} %Fig.~\ref{fig:RQ2method_phenomena} shows a largely balanced distribution of methods across syntactic phenomena, with three notable exceptions: behavioral methods dominate syntax–semantics interface phenomena (negation/NPIs; binding/coreference), mechanistic methods dominate benchmark-based evaluations, and probing methods dominate dependency/constituency parsing and POS-level phenomena, likely reflecting early methodological trends.
%As for syntactic phenomena, we again find a quite balanced distribution of methods (see Fig.~\ref{fig:RQ2method_phenomena}), with three exceptions. 
Fig.~\ref{fig:RQ2method_phenomena} shows a largely balanced distribution of methods across syntactic phenomena, with three notable exceptions. First, behavioral methods dominate for the phenomena that are most clearly at the syntax-semantics interface: \textit{negation \& NPIs} and \textit{binding and coreference}. Second, mechanistic methods dominate in \textit{benchmark}-based evaluations. Third, probing methods dominate in \textit{dependency parsing \& constituency} and \textit{POS \& other lexical properties}; %\gb{I'm improvising here, we should check:} ALSO EASIER TO MAKE A CLASSIFIER TASK THERE THAN FOR E.G. BINDING OR ARGUMENT STRUCTUREs}
for the latter, again we suspect this is an artifact of time, since these were among the first phenomena to be checked frequently with probing methods in earlier years. Moreover, tasks related to POS tagging and lexical properties are generally easier to formulate as classification problems than phenomena such as binding or argument structures, which may also contribute to this pattern.

\paragraph{Experimental material.} With respect to experimental materials, instead, clear differences emerge (Fig.~\ref{fig:RQ2method_phenomena}, bottom-left). Probing approaches (dark green) occur more frequently with naturalistic than with controlled data, particularly when targeting lexical or morphosyntactic properties such as POS (see Fig.~\ref{fig:RQ2method_phenomena}, top); however, this pairing reflects an empirical co-occurrence rather than a methodological dependency, as the easy availability of e.g.\ POS annotations makes naturalistic data a convenient, but not necessary, choice for probing.
Probing methods are instead almost absent in studies with causal TLMs (Fig.~\ref{fig:RQ2method_phenomena}, bottom-right); we suspect that this is a historical accident, as most studies of causal TLMs have been carried out in 2024-25, when probing methods have become less common (note, however, that as mentioned above many mechanistic methods include some probing component).

Behavioral methods (lime yellow) pattern most frequently with controlled data, which makes sense given that many existing psycholinguistic datasets can be adapted to test TLMs, as behavioral methods need carefully designed data to test for a given property. Thus, the frequent pairing of behavioral methods with controlled data reflects the adaptation of psycholinguistic paradigms to test models, satisfying the need for carefully designed data to test specific properties.

\section{Additional Results to Section 4}\label{app:sec_4}
\subsection{Additional Results to BLiMP}\label{app:rq3_blimp}

\begin{figure*}
  \centering
  \subfloat{%
  \includegraphics[trim=0.25cm 0.1cm 0.1cm 0.1cm, clip, scale=0.45]{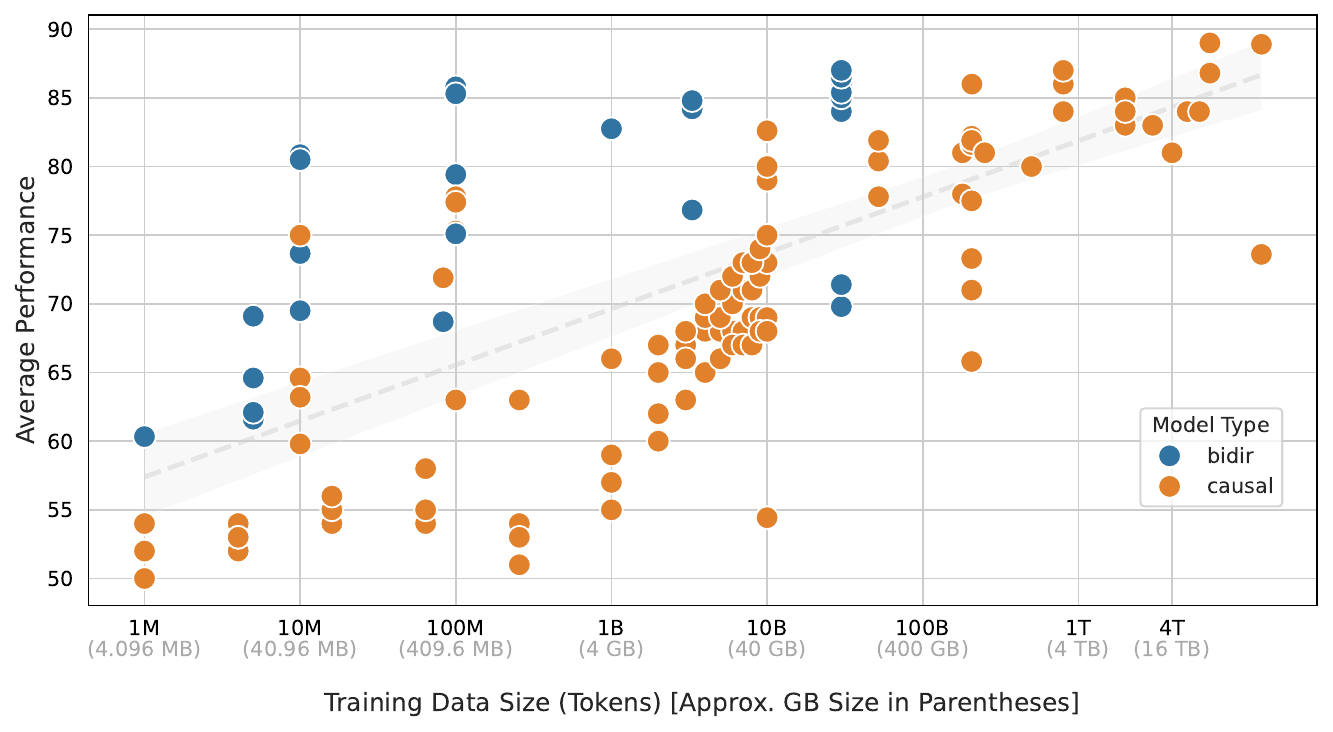}}
  \subfloat{%
  \raisebox{0.5cm}{%
  \includegraphics[trim=0.2cm 0.1cm 0.2cm 0.1cm, clip, scale=0.5]{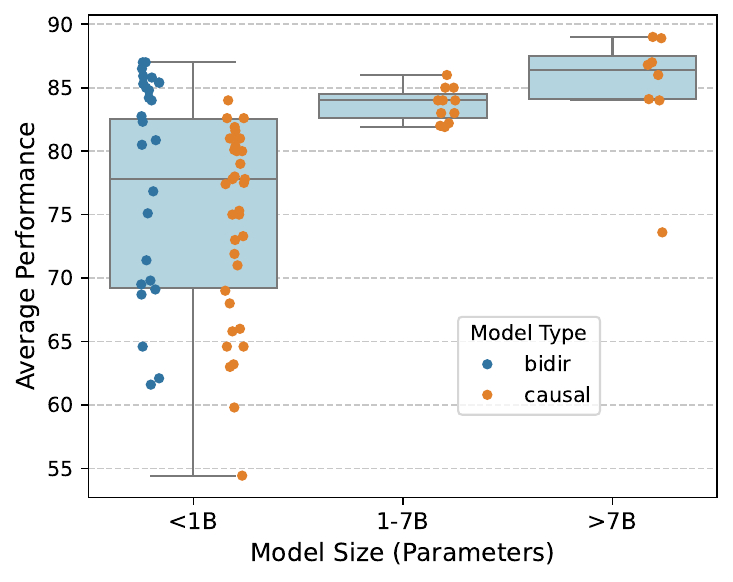}
  }}
  %\subfloat{\includegraphics[trim=0.25cm 0.6cm 0.25cm 0.2cm, clip, scale=0.49]{figures/Combined_Blimp_Analysis.pdf}}\\
  \caption{\textbf{Left:} BLiMP \cite{Warstadt:etal:2020} average scores plotted against the number of training tokens (log scale). \textbf{Right:} BLiMP  average scores shown by model size and model type. If a paper reports scores for TLMs trained on varying sizes of training data, we include only the fully trained models in the boxplot.}
  \label{fig:RQ3blimp4_training}
\end{figure*}

Fig.~\ref{fig:RQ3blimp4_training} (left) shows the average BLiMP scores as a function of training data size, color-coded by model type. The first thing to note is how good models are according to this benchmark: the overall BLiMP score is 72\% (random baseline: 50\%).
We also observe the well-known scaling pattern in deep learning: TLMs trained on larger amounts of data tend to perform better (linearly in log scale), as do larger TLMs (see also Fig.~\ref{fig:RQ3blimp4_training}, left). 

As for the difference between model types, note that, for a given size or amount of training data, bidirectional models tend to outperform causal models \cite[as also observed in][]{waldis_holmes_2024}. Since BLiMP uses sentence-level comparisons, this cannot solely be due to the fact that bidirectional models, but not causal models, have access to the whole context of the stimulus when they process the critical fragment (e.g.\ \textit{themselves/herself} in the first example in Fig.~\ref{fig:blimp_cats}). We hypothesize instead that bidirectional models reach better sentence representations due to the bidirectional nature of the training task. However, for larger amounts of training data and model sizes ($>$30B training tokens and $>$1B parameters) only causal models have been examined in the literature we cover (Fig.~\ref{fig:RQ3blimp4_training}, right), such that we cannot know whether the trend continues at these scales. %\id{Maybe better to say that bidir do not come at those sizes? I guess it's not that they've not been studied, but that they cannot be studied because they don't exist. Confirm.} \no{there is DeBERTa-v3-XXLarge (~1.5B), but we do not have blimp results}

\begin{table}[t]
\centering
\small
\begin{tabular}{p{4cm} c c}
\toprule
 & coeff.  & SE\\ %$\sigma$
\midrule
Intercept & 77.06*** & (1.91) \\
Model Type (causal)& -8.9*** & (2.16) \\
Training Data Size (in billions) & 0.002*** & (0.00) \\
\midrule
Observations & 134 &\\
%$R^2$        & 0.198 &\\
%Adj. $R^2$   & 0.186 &\\
Log-Likelihood: & -492 &\\
BIC   &  1012 &\\
AIC   & 995 &\\               
\bottomrule
\end{tabular}
\vspace{1mm}
\footnotesize Note: * $p < 0.05$, ** $p < 0.01$, *** $p < 0.001$.
%\end{tablenotes}
\caption{Mixed-effects regression model predicting average BLiMP accuracy scores. Standard errors in parentheses.}
\label{tab:regression_blimp}
\end{table}

\paragraph{Statistical modeling.} 
%In particular, we fit a mixed effects regression model with syntactic phenomena, model type (causal / bidirectional), and language support (to see how syntactic knowledge varies across languages) as fixed effects and model and study id as random effects.
We fit a mixed-effects regression model predicting average BLiMP accuracy scores (134 data points in total), with model type (causal/bidirectional) and training data size as fixed effects and model name and study ID as random effects. Full model specifications and estimates are reported in Table~\ref{tab:regression_blimp}.  Although the explained variance is modest (log-likelihood $= -494$, $AIC=997$, $BIC=1012$), both model type and training data size were significant predictors: causal models performed worse ($\beta=-9.1$, $SE=2.08$, $p<0.001$), and training data size showed a small positive effect ($\beta= 0.002$, $SE=0.0$, $p<0.001$).\footnote{The intercept ($\beta=77.57$, $SE=1.83$, $p<0.001$) represents the expected score for a bidirectional TLMs at zero training data (represented in billions of tokens).}

\begin{figure}[!ht]
  %\centering
  %\begin{subfigure}[b]{0.25\textwidth}
    %\raisebox{2cm}[0pt][0pt]{%
      %\includegraphics[trim=0.2cm 0.1cm 0.2cm 0.1cm, clip, scale=0.5]{figures/Blimp_scores_size.pdf}
    %}
  %\end{subfigure}
  %\hspace{2.1cm}
    %\begin{subfigure}[b]{0.49\textwidth}
    \includegraphics[trim=0.31cm 0.4cm 0.3cm 0.2cm, clip, scale=0.4]{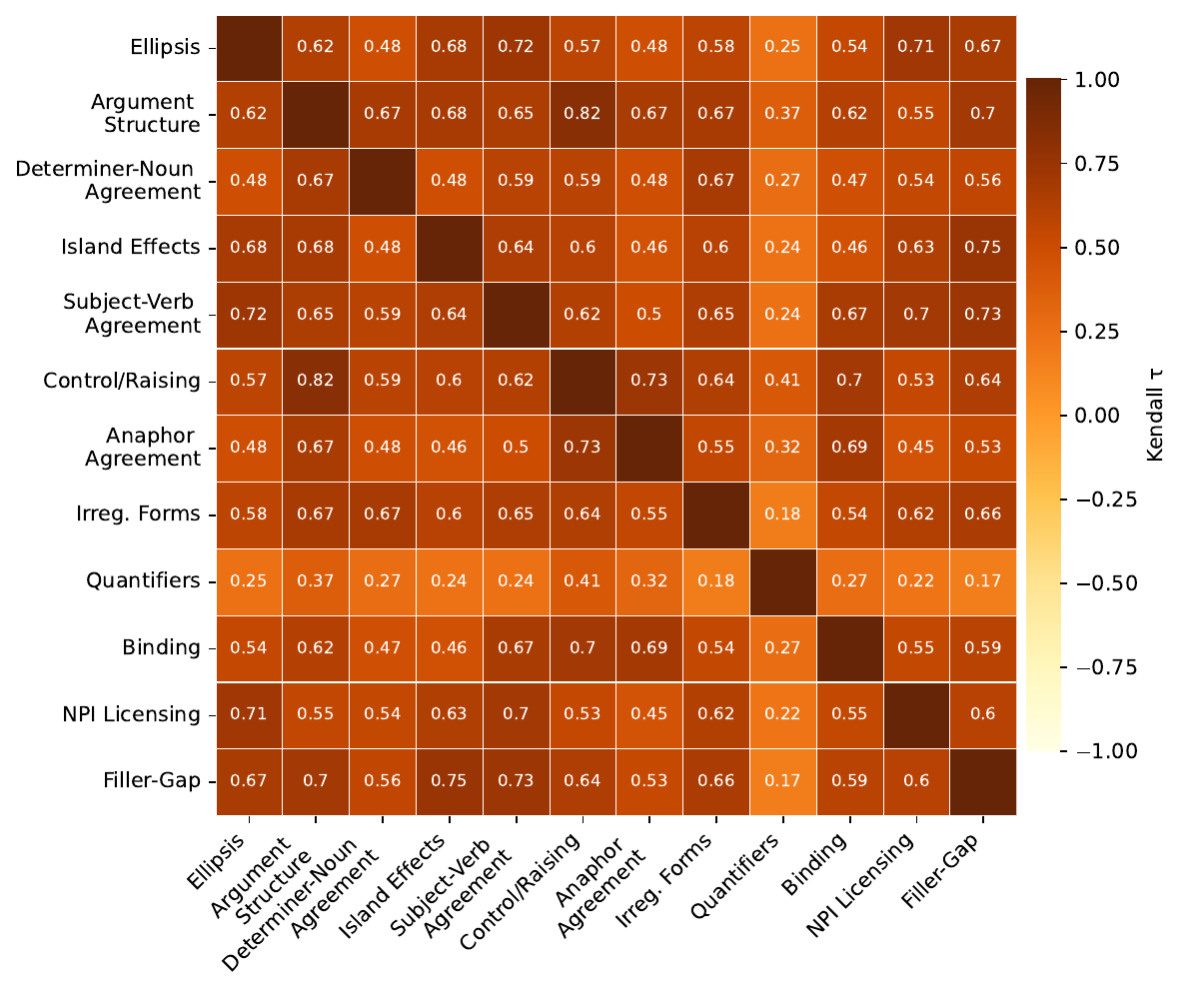}
  %\end{subfigure}
  \caption{\textbf Computed model rankings per phenomenon, then Kendall's $\tau$ coefficient, visualized as a heatmap.} \label{fig:RQ3blimp_size}
\end{figure}

\paragraph{Variance across phenomena.} We test whether, on average, models that are good/bad at a given phenomenon are also good/bad at other phenomena. We ranked TLMs by performance within each linguistic phenomenon and added slight jitter to handle ties. Then, we computed Kendall’s $\tau$ correlations between these rankings across linguistic phenomena categories.
%to see how similarly models perform on different linguistic phenomena.
Fig.~\ref{fig:RQ3blimp_size} presents a heatmap with the correlation score between each phenomenon. 
Kendall’s  $\tau$  correlations show strong and consistent relationships across most linguistic phenomena, with an average correlation of about 0.58 (SD $\approx$ 0.14), indicating very strong agreement in TLM performance across tasks. Quantifiers are a clear outlier, exhibiting weak to moderate correlations with other categories. In contrast, several pairs show especially high alignment, notably Argument Structure and Control/Raising ($\tau$  = 0.82) and Filler-Gap and Island Effects ($\tau$  = 0.75).
%\gb{Nora please comment on the results: overall strong correlations (give avg and stdev), quantifiers an exception. See how to qualitatively interpret tau according to Gemini in the commented-out text below; and note that tau is generally lower than spearman's rho, hence the optimism in the qualitative interpretation.}
%Kendall's Tau Value ($|\tau|$) | Interpretation |
%| :--- | :--- |
%| 0.00 – 0.10 | Negligible / Very Weak |
%| 0.10 – 0.19 | Weak |
%| 0.20 – 0.34 | Moderate |
%| 0.35 – 0.49 | Strong / Substantial |
%| 0.50 or above | Very Strong / Excellent | }

\subsection{Additional Results to Beyond Benchmark and English}\label{app:rq3_non-blimp}
% report on statistical model
% per category results

\begin{figure}[!ht]
    \centering
    %\subfloat{\includegraphics[scale=0.43]{figures/method_years.png}}\\
    \subfloat{\includegraphics[trim=0.2cm 0.2cm 0.2cm 1cm, clip, scale=0.55]{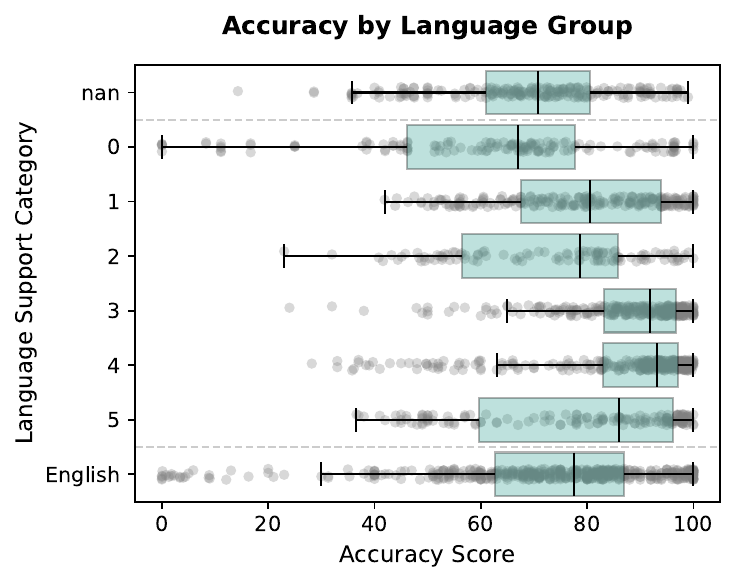}}\\
    \vspace{0.3cm}
    \subfloat{\includegraphics[trim=0.2cm 0.2cm 0.2cm 0.8cm, clip, scale=0.51]{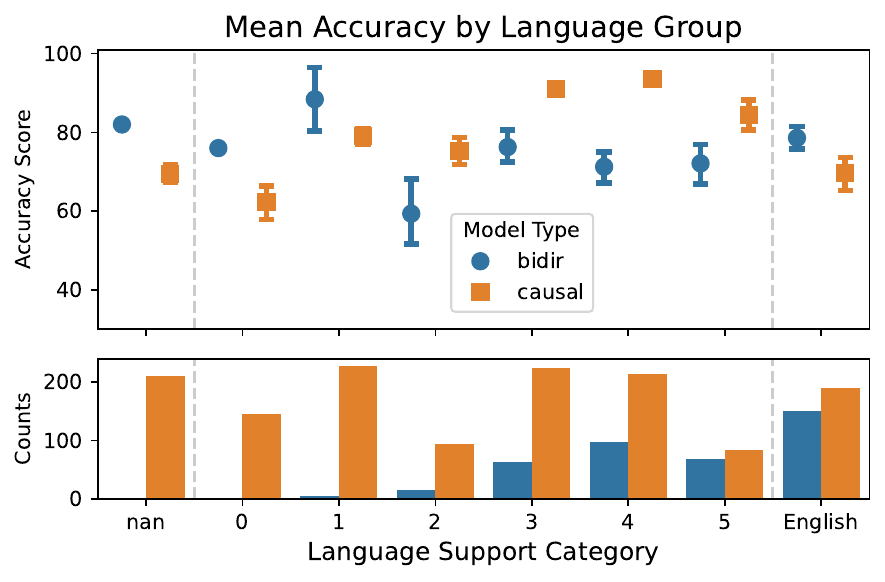}}
    \caption{\textbf{Top:} Language support categories, categories taken from \citet{Joshi:etal:2020}, with nan and English (category 5.) separately. \textbf{Bottom:} Mean accuracy scores by language group and model type.
    }\label{fig:language_support}
\end{figure}

Fig.~\ref{fig:language_support} displays the distribution of TLM
accuracy scores grouped by language support, based
on the five categories reported by \citet{Joshi:etal:2020} and model type (bottom).

\paragraph{Statistical modeling.} We estimate a mixed-effects regression model to predict accuracy scores as a function of model type, linguistic phenomenon, and language support category, with random intercepts for model name and study ID. To ensure comparability across languages, we exclude instances with missing language support annotations (i.e., ``nan'' category in Fig.~\ref{fig:language_support}), as well as all scores annotated as \textit{benchmark}. The resulting analysis is performed on 1,577 observations. Language support is treated as an ordered categorical variable \citep[following][]{Joshi:etal:2020}. Full model specifications and estimates are reported in Table~\ref{tab:regression}, and a forest plot of the estimated coefficients is shown in Fig.~\ref{fig:forestplot}.

\begin{table}[t]
\centering
\small
\begin{tabular}{p{4cm} c c}
\toprule
 & coeff.  & SE\\ %$\sigma$
\midrule
Intercept & 0.4*** & (0.03) \\
Model Type (causal)& 0.018 & (0.03) \\
Language support & 0.065*** & (0.003) \\
\midrule                 
Agreement \& feature-checking & 0.11***  & (0.02) \\
Argument structure & -0.09** & (0.03) \\
%Benchmarks & -0.032* & (0.015) \\
Binding \& coreference & 0.03 & (0.04) \\
Dependency parsing  & -0.04  & (0.03) \\ %\& constituency
Ellipsis \& filler–gap depend. & -0.002  & (0.04) \\
Hierarchical structure & -0.06*  & (0.03) \\
Negation \& NPIs & -0.09** & (0.03) \\
POS \& other lexical properties & 0.12*** & (0.03) \\
\midrule
%Group Var  0.011    0.022                           
%model Var   0.005    0.010   
Observations & 1577 &\\
%$R^2$        & 0.23 &\\
%Adj. $R^2$   & 0.22 &\\
Log-Likelihood &  539 &\\
BIC   &  -975 &\\
AIC   & -1050 &\\               
\bottomrule
\end{tabular}
\vspace{1mm}
\footnotesize Note: * $p < 0.05$, ** $p < 0.01$, *** $p < 0.001$.
%\end{tablenotes}
\caption{Mixed-effects regression model predicting accuracy scores; standard errors in parentheses.}
\label{tab:regression}
\end{table}

%\begin{figure*}[!ht]
    %\centering
    %\subfloat{\includegraphics[scale=0.43]{figures/method_years.png}}\\
    %\subfloat{\includegraphics[trim=0cm 0cm 0 0cm, clip, scale=0.35]{figures/material_method_phenomena.png}}
    %\caption{\small Counts of (interpretability technique) experimental design methods across experimental materials and syntactic phenomena.
    %}
    %\label{fig:phenomXmethodXmaterial}
%\end{figure*}

\subsection{Additional Results to Probing}\label{app:probing}
%We seek out to identify whether probing methods successfully capture syntactic representations. % or whether their findings are potentially influenced by publication bias. 
%To this end, we examine the distribution of reported outcomes across studies, distinguishing between positive, mixed and critical, and negative results. While the results show a high proportion of studies reporting successful recovery of syntactic information, the relative scarcity of negative findings suggests a possible skew in the literature toward positive results (potentially reflecting a publication bias).
%However, this pattern is partially mitigated in studies that combine probing with mechanistic analyses, which report more frequently mixed or negative outcomes. This observation provides evidence to support criticisms like those of \citet{Belinkov:2022}, who argue that standard probing may overestimate the presence of linguistic knowledge, as it primarily captures recoverability rather than functional use. Such observations emphasize the importance of complementing probing with more causally informative methods when assessing syntactic representations.

We analyze 147 studies in total that employ probing methods, which includes all papers coded as probing, plus some coded as mechanistic when they also employed probing methods. Fig.~\ref{fig:RQ3probing_lingPhenomena} (left) shows that probing-based studies successfully identify syntactic information at comparable rates across linguistic phenomena categories.

%\begin{figure}[!ht]
  %\centering
  %\subfloat{\includegraphics[trim=0.15cm 0.7cm 0.2cm 0.2cm, clip, scale=0.47]{figures/probing_individ_ling_phenomena-1.pdf}}
  %\caption{Successful recovery of syntactic information from TLMs' representations via probing across linguistic phenomena.}\label{fig:RQ3probing_lingPhenomena}
%\end{figure}

\subsection{Additional Results to Mechanistic Interpretability}\label{app:mechanistic}

We aim to characterize the mechanistic interpretability field by categorizing all 140 mechanistic papers by level of analysis and method. To abstract over this diverse set of approaches and assess what TLMs learn about syntax, we evaluate whether studies report evidence of syntactic knowledge or instead highlight its limitations.
Although few in counts, Fig.~\ref{fig:RQ3probing_lingPhenomena} (right), the mixed and critical findings suggest that models rely heavily on surface-level patterns rather than abstract syntactic rules, exhibiting limited and inconsistent sensitivity to structural changes. Also, TLMs' syntactic abilities vary across conditions, improving with more data and in high-resource languages but remaining weak in low-resource settings, and can even degrade with fine-tuning.

%\begin{figure*}[ht] 
    %\centering
        %\centering
        %\fbox{\includegraphics[width=0.8\textwidth]{figures/ratings_syntax_summary.png}}
       % \caption{Instance of a ChatGPT \citep{chatgpt5:2025} prompt showing our evaluation query with an example.}
    %\label{fig:subfig3}
%\end{figure*}

\end{document}

%% file: table_refrences.tex
% 337
\onecolumn
\begin{longtblr}[
  caption = {The 337 studies included in the database, sorted by year and alphabetically.},
  label = {tab:ref},
]{
  colspec = {|X|},
  %rowhead = 1,
  hlines,
  %row{even} = {gray9},
  %row{1} = {olive9},
} 
%\centering
%\small
%\begin{tabular}{|p{15cm}|}
%\hline
%\textbf{year:} 337 studies included and analyzed in our database.\\
%\hline
\textbf{2018:} 
\citet{tran:etal:2018}, 
\citet{Peters:etal:2018}\\
%\hline
\textbf{2019:}
\citet{abdou_higher-order_2019}, 
\citet{bacon_does_2019}, 
\citet{cao_multilingual_2019}, 
\citet{chrupala_correlating_2019}, 
\citet{clark_what_2019}, 
\citet{coenen_visualizing_2019}, 
\citet{goldberg_assessing_2019}, 
\citet{hewitt_structural_2019}, 
\citet{htut_attention_2019}, 
\citet{jawahar_what_2019},
\citet{krasnowska-kieras-wroblewska-2019-empirical},
\citet{lin_open_2019}, 
\citet{liu_linguistic_2019}, 
\citet{marecek_balustrades_2019}, 
\citet{mccoy_right_2019}, 
\citet{pires_how_2019}, 
\citet{ravishankar_multilingual_2019}, 
\citet{ronnqvist_is_2019}, 
\citet{rosa_inducing_2019}, 
\citet{salvatore_logical-based_2019}, 
\citet{sundararaman_syntax-infused_2019}, 
\citet{tenney_bert_2019}, 
\citet{Tenney:etal:2019b}, 
\citet{van_schijndel_quantity_2019}, 
\citet{vig_analyzing_2019}, 
\citet{wang_cross-lingual_2019}, 
\citet{wang_superglue_2019}, 
\citet{warstadt_investigating_2019}, 
\citet{wu_beto_2019},
\citet{yanaka-etal-2019-neural}\\
%\hline
\textbf{2020:} 
\citet{alt_probing_2020}, 
\citet{artetxe_cross-lingual_2020}, 
\citet{celikkanat_controlling_2020}, 
\citet{chi_finding_2020}, 
\citet{dalvi_analyzing_2020}, 
\citet{davis_discourse_2020}, 
\citet{durrani_analyzing_2020}, 
\citet{dufter-schutze-2020-identifying},
\citet{ettinger_what_2020}, 
\citet{torroba_hennigen_intrinsic_2020}, 
\citet{hu_closer_2020}, 
\citet{hu_systematic_2020}, 
\citet{jo_roles_2020}, 
\citet{k_cross-lingual_2020},
\citet{kahardipraja-etal-2020-exploring},
\citet{ma_kelly_which_2020},
\citet{Kim2020Are},
\citet{klafka_spying_2020}, 
\citet{kulmizev_neural_2020}, 
\citet{kunz_classifier_2020}, 
\citet{lepori_picking_2020}, 
\citet{limisiewicz_universal_2020}, 
\citet{tayyar_madabushi_cxgbert_2020},
\citet{manning_emergent_2020}, 
\citet{maudslay_tale_2020}, 
\citet{miaschi_contextual_2020}, 
\citet{miaschi_italian_2020},
\citet{miaschi-etal-2020-linguistic},
\citet{michael_asking_2020}, 
\citet{min_syntactic_2020}, 
\citet{mueller_cross-linguistic_2020}, 
\citet{pimentel_information-theoretic_2020}, 
\citet{pimentel_pareto_2020}, 
\citet{pruksachatkun_intermediate-task_2020}, 
\citet{ribeiro_beyond_2020}, 
\citet{salazar_masked_2020}, 
\citet{sorodoc_probing_2020}, 
\citet{thrush_investigating_2020}, 
\citet{vilares_parsing_2020}, 
\citet{Warstadt:etal:2020}, 
\citet{warstadt_can_2020}, 
\citet{warstadt_learning_2020}, 
\citet{wu_are_2020}, 
\citet{wu_perturbed_2020}, 
\citet{yanaka-etal-2020-neural},
\citet{yu_word_2020}, 
\citet{zhao_how_2020}, 
\citet{zhou_limit-bert_2020},
\citet{zhu_information_2020}\\
%\hline
\textbf{2021:} 
\citet{alleman_syntactic_2021}, 
\citet{bai_syntax-bert_2021}, 
\citet{bjerva_does_2021}, 
\citet{chang-etal-2021-convolutions},
\citet{chen2021probing},
\citet{chaves_look_2021}, 
\citet{elazar_amnesic_2021},
\citet{finlayson-etal-2021-causal}
\citet{fayyaz_not_2021}, 
\citet{glavas_is_2021}, 
\citet{gupta_bert_2021}, 
\citet{gupta_deep_2021}, 
\citet{hernandez_low-dimensional_2021}, 
\citet{hessel_how_2021}, 
\citet{hou_birds_2021}, 
\citet{huebner_babyberta_2021}, 
\citet{kasthuriarachchy_general_2021}, 
\citet{kim_testing_2021},
\citet{kuncoro-etal-2020-syntactic},
\citet{kunz_test_2021}, 
\citet{kuribayashi_lower_2021}, 
\citet{li_are_2021}, 
\citet{li_how_2021}, 
\citet{limisiewicz_introducing_2021}, 
\citet{liu_probing_2021}, 
\citet{Lovering2021PredictingIB},
\citet{luo_have_2021}, 
\citet{madhura_pande_heads_2021}, 
\citet{malkin_studying_2021}, 
\citet{maudslay_syntactic_2021}, 
\citet{miaschi-etal-2021-probing},
\citet{mikhailov_morph_2021}, 
\citet{mikhailov_rusenteval_2021}, 
\citet{mohebbi_exploring_2021}, 
\citet{nayak_using_2021}, 
\citet{newman_refining_2021}, 
\citet{nikoulina_rediscovery_2021}, 
\citet{oba_exploratory_2021}, 
\citet{papadimitriou_deep_2021}, 
\citet{park_deep_2021}, 
\citet{perez-mayos_assessing_2021}, 
\citet{perez-mayos_evolution_2021}, 
\citet{perez-mayos_how_2021}, 
\citet{pham_out_2021}, 
\citet{puccetti_how_2021}, 
\citet{ravfogel_counterfactual_2021}, 
\citet{ravishankar_attention_2021}, 
\citet{ryu_accounting_2021}, 
\citet{sachan_syntax_2021}, 
\citet{shapiro_multilabel_2021}, 
\citet{sinha_masked_2021}, 
\citet{sinha_unnatural_2021}, 
\citet{swamy_interpreting_2021}, 
\citet{taktasheva_shaking_2021}, 
\citet{timmapathini_probing_2021}, 
\citet{tucker_what_2021},
\citet{wei-etal-2021-frequency},
\citet{wang-etal-2021-controlled},
\citet{white_non-linear_2021},
\citet{white-cotterell-2021-examining},
\citet{wilcox_targeted_2021}, 
\citet{wu_infusing_2021}, 
\citet{xu_syntax-enhanced_2021}, 
\citet{yanaka_assessing_2021}, 
\citet{yun_transformer_2021}, 
\citet{zhang_dependency-based_2021}, 
\citet{zhang_when_2021}\\
%\hline
\textbf{2022:}
\citet{abdou_word_2022}, 
\citet{alajrami_how_2022}, 
\citet{alzetta_probing_2022}, 
\citet{arps_probing_2022},
\citet{aoyama:schneider:2022}
\citet{auyespek_hyperbolic_2022}, 
\citet{bolucu_analysing_2022}, 
\citet{chang_bergen_2022},
\citet{cherniavskii_acceptability_2022}, 
\citet{choenni_investigating_2022}, 
\citet{choshen_grammar-learning_2022}, 
\citet{clouatre_local_2022}, 
\citet{clouatre_local_2022-1}, 
\citet{conia_probing_2022}, 
\citet{de-dios-flores_computational_2022}, 
\citet{eisape_probing_2022},
\citet{garcia_targeted_2022}, 
\citet{guarasci_bert_2022}, 
\citet{kunz_where_2022}, 
\citet{la_malfa_emergent_2022}, 
\citet{lakretz_can_2022}, 
\citet{lasri_does_2022}, 
\citet{lasri_subject_2022}, 
\citet{lasri_word_2022}, 
\citet{lee_bert_2022}, 
\citet{lee_can_2022}, 
\citet{li_neural_2022}, 
\citet{li_probing_2022},
\citet{misra_minicons_2022}, 
\citet{mueller_causal_2022}, 
\citet{niu_does_2022}, 
\citet{niu_using_2022}, 
\citet{otmakhova_cross-linguistic_2022},
\citet{papadimitriou_when_2022}, 
\citet{pimentel_architectural_2022}, 
\citet{prange_linguistic_2022}, 
\citet{ravishankar_effects_2022}, 
\citet{ri_pretraining_2022}, 
\citet{sajjad_analyzing_2022}, 
\citet{sartran_transformer_2022}, 
\citet{schuster_berts_2022}, 
\citet{sevastjanova_lmfingerprints_2022}, 
\citet{sinclair_structural_2022}, 
\citet{sinha_curious_2022}, 
\citet{stanczak_same_2022}, 
\citet{tucker_when_2022}, 
\citet{wang_interpretability_2022}, 
\citet{weissweiler_better_2022},
\citet{wilcox_2022_chapter},
\citet{xu_cross-linguistic_2022}, 
\citet{yi_probing_2022}, 
\citet{yin_interpreting_2022}, 
\citet{yoshida-oseki-2022-composition},
\citet{zhang_probing_2022}, 
\citet{zheng_probing_2022}\\
%\hline
\textbf{2023:} 
\citet{amini_naturalistic_2023}, 
\citet{asher_limits_2023}, 
\citet{bai_constituency_2023}, 
\citet{blevins_prompting_2023}, 
\citet{chen_sudden_2023}, 
\citet{georges_gabriel_charpentier_not_2023},
\citet{clark_cross-linguistic_2023},
\citet{de-dios-flores-etal-2023-dependency},
\citet{de_varda_data-driven_2023}, 
\citet{dentella_systematic_2023}, 
\citet{el_mesbahi_utility_2023}, 
\citet{elgaar_ling-cl_2023}, 
\citet{gessler_syntactic_2023}, 
\citet{guarasci_assessing_2023}, 
\citet{gurnee_finding_2023}, 
\citet{hao_verb_2023}, 
\citet{hlavnova_empowering_2023}, 
\citet{hewitt-etal-2023-backpack},
\citet{hou_detecting_2023}, 
\citet{hu_prompting_2023}, 
\citet{huang_rigorously_2023}, 
\citet{jumelet_feature_2023}, 
\citet{kervadec_unnatural_2023}, 
\citet{kim_reconstruction_2023}, 
\citet{lee_decoding_2023}, 
\citet{lin_chatgpt_2023}, 
\citet{mahowald_discerning_2023}, 
\citet{miaschi_testing_2023}, 
\citet{michaelov_structural_2023}, 
\citet{muller-eberstein_subspace_2023}, 
\citet{munoz-ortiz_assessment_2023}, 
\citet{murty_grokking_2023}, 
\citet{mysiak_is_2023}, 
\citet{nikolaev_universe_2023}, 
\citet{oota_joint_2023}, 
\citet{papadimitriou_injecting_2023}, 
\citet{petersen_lexical_2023}, 
\citet{roy_benchclamp_2023}, 
\citet{ruzzetti_exploring_2023}, 
\citet{sharma-etal-2023-learning},
\citet{sieker_when_2023}, 
\citet{sinha_language_2023}, 
\citet{someya_jblimp_2023}, 
\citet{srivastava_beyond_2023}, 
\citet{stanczak_latent-variable_2023}, 
\citet{tjuatja_syntax_2023}, 
\citet{weissweiler_explaining_2023}, 
\citet{wettig_should_2023}, 
\citet{wijnholds_assessing_2023}, 
\citet{xu_are_2023}, 
\citet{yamakoshi_causal_2023}, 
\citet{zhang_closer_2023}, 
\citet{zhang_how_2023}, 
\citet{zhao_transformers_2023}, 
\citet{zheng_what_2023}, 
\citet{zhou_how_2023}\\
%\hline
\textbf{2024:}
\citet{acs_morphosyntactic_2024}, 
\citet{allen-zhu_interpretability_2024}, 
\citet{ambridge_large_2024}, 
\citet{arehalli_neural_2024}, 
\citet{arora_causalgym_2024}, 
\citet{arps_multilingual_2024},
\citet{n_atox_evaluating_nodate},
\citet{chang_when_2024}, 
\citet{chen_when_2024}, 
\citet{diehl_martinez_mitigating_2024}, 
\citet{simon2024a}, 
\citet{feng-etal-2024-child},
\citet{ferrando_information_2024}, 
\citet{ferrando_similarity_2024}, 
\citet{ginn_tree_2024}, 
\citet{hale_llms_2024}, 
\citet{he_decoding_2024}, 
\citet{hu_auxiliary_2024}, 
\citet{hu_language_2024}, 
\citet{huang_large-scale_2024}, 
\citet{kallini_mission_2024}, 
\citet{kim_does_2024}, 
\citet{kuribayashi_emergent_2024}, 
\citet{lampinen_can_2024}, 
\citet{leong2024testinglearninghypothesesusing},
\citet{li_incremental_2024}, 
\citet{marks_geometry_2024}, 
\citet{marks_sparse_2024},
\citet{mccoy_embers_2023}, 
\citet{mcgee_evidence_2024}, 
\citet{miaschi_evaluating_2024}, 
\citet{misra_language_2024}, 
\citet{patil_filtered_2024}, 
\citet{someya_targeted_2024}, 
\citet{turner_steering_2024}, 
\citet{waldis_holmes_2024}, 
\citet{weber_interpretability_2024}, 
\citet{wijesiriwardene_relationship_2024}, 
\citet{wilcox_using_2024}, 
\citet{zhang_unveiling_2024}\\
%\hline
\textbf{2025:} 
\citet{acs_morphosyntactic_2024}
\citet{agarwal:etal:2025}, 
\citet{alkhamissi-etal-2025-language},
\citet{alzetta_parallel_2025}, 
\citet{ahuja_learning_2025}, 
\citet{aljaafari_interpreting_2025}, 
\citet{aoyama_language_2025}, 
\citet{brinkmann_large_2025}, 
\citet{bunzeck_subword_2025}, 
\citet{cheng_linguistic_2025}, 
\citet{constantinescu_investigating_2025}, 
\citet{dentella_language_2025}, 
\citet{duan_how_2025}, 
\citet{duan_unnatural_2025}, 
\citet{he_large_2025}, 
\citet{hu_between_2025}, 
\citet{ide_how_2025}, 
\citet{ju_domain_2025},
\citet{jumelet:etal:2025},
\citet{kennedy_evidence_2025},
\citet{kennedy_evidence_2025-1}, 
\citet{kissane_probing_2025}, 
\citet{koyama_targeted_2025}, 
\citet{kryvosheieva_controlled_2025}, 
\citet{nandi_sneaking_2025}, 
\citet{qiu_grammaticality_2025}, 
\citet{someya_derivational_2025}, 
\citet{swarup-etal-2025-syntax},
\citet{tian_large_2025}, 
\citet{vandermeerschen_supervised_2025}, 
\citet{wang_extracting_2025}, 
\citet{WILCOX2025104650},
\citet{xefteri-etal-2025-syntactic},
\citet{xu2025languagemodelslearntypologically},
\citet{yang_anything_2025}, 
\citet{zhao_systematic_2025}, 
\citet{zhou_linguistic_2025}, 
\citet{zimmerman_tokens_2025}\\
%\hline
%\end{tabular}
%\caption{Studies included and analyzed in our database, sorted by publication year and alphabetical order.}\label{tab:ref}
\end{longtblr}
%\end{table*}

\twocolumn

%\begin{table*}[t]
%\centering
%\small
%\begin{tabular}{|p{15cm}|}
%\hline
%\textbf{year:} Studies included and analyzed in our database.\\
%\hline

%% file: SyntaxInLLMs.bib
@inproceedings{chang-etal-2021-convolutions,
    title = "Convolutions and Self-Attention: {R}e-interpreting Relative Positions in Pre-trained Language Models",
    author = "Chang, Tyler A.  and
      Xu, Yifan  and
      Xu, Weijian  and
      Tu, Zhuowen",
    editor = "Zong, Chengqing  and
      Xia, Fei  and
      Li, Wenjie  and
      Navigli, Roberto",
    booktitle = "Proceedings of the 59th Annual Meeting of the Association for Computational Linguistics and the 11th International Joint Conference on Natural Language Processing (Volume 1: Long Papers)",
    month = aug,
    year = "2021",
    address = "Online",
    publisher = "Association for Computational Linguistics",
    url = "https://aclanthology.org/2021.acl-long.333/",
    doi = "10.18653/v1/2021.acl-long.333",
    pages = "4322--4333"
}

@article{sevastjanova_lmfingerprints_2022,
	title = {{LMFingerprints}: {Visual} {Explanations} of {Language} {Model} {Embedding} {Spaces} through {Layerwise} {Contextualization} {Scores}},
	volume = {41},
	issn = {0167-7055, 1467-8659},
	shorttitle = {\textit{{LMFingerprints}}},
	url = {https://onlinelibrary.wiley.com/doi/10.1111/cgf.14541},
	doi = {10.1111/cgf.14541},
	language = {en},
	number = {3},
	urldate = {2025-07-22},
	journal = {Computer Graphics Forum},
	author = {Sevastjanova, R. and Kalouli, A. and Beck, C. and Hauptmann, H. and El‐Assady, M.},
	month = jun,
	year = {2022},
	pages = {295--307},
}

@inproceedings{sajjad_analyzing_2022,
	address = {Seattle, United States},
	title = {Analyzing {Encoded} {Concepts} in {Transformer} {Language} {Models}},
	url = {https://aclanthology.org/2022.naacl-main.225/},
	doi = {10.18653/v1/2022.naacl-main.225},
	urldate = {2025-07-22},
	booktitle = {Proceedings of the 2022 {Conference} of the {North} {American} {Chapter} of the {Association} for {Computational} {Linguistics}: {Human} {Language} {Technologies}},
	publisher = {Association for Computational Linguistics},
	author = {Sajjad, Hassan and Durrani, Nadir and Dalvi, Fahim and Alam, Firoj and Khan, Abdul and Xu, Jia},
	editor = {Carpuat, Marine and de Marneffe, Marie-Catherine and Meza Ruiz, Ivan Vladimir},
	month = jul,
	year = {2022},
	pages = {3082--3101},
}

@inproceedings{krasnowska-kieras-wroblewska-2019-empirical,
    title = "Empirical Linguistic Study of Sentence Embeddings",
    author = "Krasnowska-Kiera{\'s}, Katarzyna  and
      Wr{\'o}blewska, Alina",
    editor = "Korhonen, Anna  and
      Traum, David  and
      M{\`a}rquez, Llu{\'i}s",
    booktitle = "Proceedings of the 57th Annual Meeting of the Association for Computational Linguistics",
    month = jul,
    year = "2019",
    address = "Florence, Italy",
    publisher = "Association for Computational Linguistics",
    url = "https://aclanthology.org/P19-1573/",
    doi = "10.18653/v1/P19-1573",
    pages = "5729--5739"
}

@inproceedings{schuster_berts_2022,
	address = {Dublin, Ireland},
	title = {From {BERT}`s {Point} of {View}: {Revealing} the {Prevailing} {Contextual} {Differences}},
	shorttitle = {From {BERT}`s {Point} of {View}},
	url = {https://aclanthology.org/2022.findings-acl.89/},
	doi = {10.18653/v1/2022.findings-acl.89},
	urldate = {2025-07-22},
	booktitle = {Findings of the {Association} for {Computational} {Linguistics}: {ACL} 2022},
	publisher = {Association for Computational Linguistics},
	author = {Schuster, Carolin M. and Hegelich, Simon},
	editor = {Muresan, Smaranda and Nakov, Preslav and Villavicencio, Aline},
	month = may,
	year = {2022},
	pages = {1120--1138},
}

@inproceedings{finlayson-etal-2021-causal,
    title = "Causal Analysis of Syntactic Agreement Mechanisms in Neural Language Models",
    author = "Finlayson, Matthew  and
      Mueller, Aaron  and
      Gehrmann, Sebastian  and
      Shieber, Stuart  and
      Linzen, Tal  and
      Belinkov, Yonatan",
    editor = "Zong, Chengqing  and
      Xia, Fei  and
      Li, Wenjie  and
      Navigli, Roberto",
    booktitle = "Proceedings of the 59th Annual Meeting of the Association for Computational Linguistics and the 11th International Joint Conference on Natural Language Processing (Volume 1: Long Papers)",
    month = aug,
    year = "2021",
    address = "Online",
    publisher = "Association for Computational Linguistics",
    url = "https://aclanthology.org/2021.acl-long.144/",
    doi = "10.18653/v1/2021.acl-long.144",
    pages = "1828--1843"
}

@inproceedings{ravishankar_multilingual_2019,
	address = {Turku, Finland},
	title = {Multilingual {Probing} of {Deep} {Pre}-{Trained} {Contextual} {Encoders}},
	url = {https://aclanthology.org/W19-6205/},
	urldate = {2025-07-22},
	booktitle = {Proceedings of the {First} {NLPL} {Workshop} on {Deep} {Learning} for {Natural} {Language} {Processing}},
	publisher = {Linköping University Electronic Press},
	author = {Ravishankar, Vinit and Gökırmak, Memduh and Øvrelid, Lilja and Velldal, Erik},
	editor = {Nivre, Joakim and Derczynski, Leon and Ginter, Filip and Lindi, Bjørn and Oepen, Stephan and Søgaard, Anders and Tidemann, Jörg},
	month = sep,
	year = {2019},
	pages = {37--47},
}

@inproceedings{puccetti_how_2021,
	address = {Online},
	title = {How {Do} {BERT} {Embeddings} {Organize} {Linguistic} {Knowledge}?},
	url = {https://aclanthology.org/2021.deelio-1.6/},
	doi = {10.18653/v1/2021.deelio-1.6},
	urldate = {2025-07-22},
	booktitle = {Proceedings of {Deep} {Learning} {Inside} {Out} ({DeeLIO}): {The} 2nd {Workshop} on {Knowledge} {Extraction} and {Integration} for {Deep} {Learning} {Architectures}},
	publisher = {Association for Computational Linguistics},
	author = {Puccetti, Giovanni and Miaschi, Alessio and Dell'Orletta, Felice},
	editor = {Agirre, Eneko and Apidianaki, Marianna and Vulić, Ivan},
	month = jun,
	year = {2021},
	pages = {48--57},
}

@inproceedings{pimentel_pareto_2020,
	address = {Online},
	title = {Pareto {Probing}: {Trading} {Off} {Accuracy} for {Complexity}},
	shorttitle = {Pareto {Probing}},
	url = {https://aclanthology.org/2020.emnlp-main.254/},
	doi = {10.18653/v1/2020.emnlp-main.254},
	urldate = {2025-07-22},
	booktitle = {Proceedings of the 2020 {Conference} on {Empirical} {Methods} in {Natural} {Language} {Processing} ({EMNLP})},
	publisher = {Association for Computational Linguistics},
	author = {Pimentel, Tiago and Saphra, Naomi and Williams, Adina and Cotterell, Ryan},
	editor = {Webber, Bonnie and Cohn, Trevor and He, Yulan and Liu, Yang},
	month = nov,
	year = {2020},
	pages = {3138--3153},
}

@inproceedings{pimentel_information-theoretic_2020,
	address = {Online},
	title = {Information-{Theoretic} {Probing} for {Linguistic} {Structure}},
	url = {https://aclanthology.org/2020.acl-main.420/},
	doi = {10.18653/v1/2020.acl-main.420},
	urldate = {2025-07-22},
	booktitle = {Proceedings of the 58th {Annual} {Meeting} of the {Association} for {Computational} {Linguistics}},
	publisher = {Association for Computational Linguistics},
	author = {Pimentel, Tiago and Valvoda, Josef and Maudslay, Rowan Hall and Zmigrod, Ran and Williams, Adina and Cotterell, Ryan},
	editor = {Jurafsky, Dan and Chai, Joyce and Schluter, Natalie and Tetreault, Joel},
	month = jul,
	year = {2020},
	pages = {4609--4622},
}

@inproceedings{pimentel_architectural_2022,
	address = {Abu Dhabi, United Arab Emirates},
	title = {The {Architectural} {Bottleneck} {Principle}},
	url = {https://aclanthology.org/2022.emnlp-main.788/},
	doi = {10.18653/v1/2022.emnlp-main.788},
	urldate = {2025-07-22},
	booktitle = {Proceedings of the 2022 {Conference} on {Empirical} {Methods} in {Natural} {Language} {Processing}},
	publisher = {Association for Computational Linguistics},
	author = {Pimentel, Tiago and Valvoda, Josef and Stoehr, Niklas and Cotterell, Ryan},
	editor = {Goldberg, Yoav and Kozareva, Zornitsa and Zhang, Yue},
	month = dec,
	year = {2022},
	pages = {11459--11472},
}

@article{htut_attention_2019,
	title = {Do {Attention} {Heads} in {BERT} {Track} {Syntactic} {Dependencies}?},
	url = {http://arxiv.org/abs/1911.12246},
	doi = {10.48550/arXiv.1911.12246},
	urldate = {2025-07-22},
	journal = {NY Academy of Sciences NLP, Dialog, and Speech Workshop},
	author = {Htut, Phu Mon and Phang, Jason and Bordia, Shikha and Bowman, Samuel R.},
	year = {2019},
	note = {arXiv:1911.12246},
}

@inproceedings{papadimitriou_when_2022,
	address = {online},
	title = {When classifying arguments, {BERT} doesn't care about word order...except when it matters},
	url = {https://aclanthology.org/2022.scil-1.17/},
	urldate = {2025-07-22},
	booktitle = {Proceedings of the {Society} for {Computation} in {Linguistics} 2022},
	publisher = {Association for Computational Linguistics},
	author = {Papadimitriou, Isabel and Futrell, Richard and Mahowald, Kyle},
	editor = {Ettinger, Allyson and Hunter, Tim and Prickett, Brandon},
	month = feb,
	year = {2022},
	pages = {203--205},
}

@inproceedings{papadimitriou_deep_2021,
	address = {Online},
	title = {Deep {Subjecthood}: {Higher}-{Order} {Grammatical} {Features} in {Multilingual} {BERT}},
	shorttitle = {Deep {Subjecthood}},
	url = {https://aclanthology.org/2021.eacl-main.215/},
	doi = {10.18653/v1/2021.eacl-main.215},
	urldate = {2025-07-22},
	booktitle = {Proceedings of the 16th {Conference} of the {European} {Chapter} of the {Association} for {Computational} {Linguistics}: {Main} {Volume}},
	publisher = {Association for Computational Linguistics},
	author = {Papadimitriou, Isabel and Chi, Ethan A. and Futrell, Richard and Mahowald, Kyle},
	editor = {Merlo, Paola and Tiedemann, Jorg and Tsarfaty, Reut},
	month = apr,
	year = {2021},
	pages = {2522--2532},
}

@inproceedings{madhura_pande_heads_2021,
	series = {15},
	title = {The heads hypothesis: {A} unifying statistical approach towards understanding multi-headed attention in {BERT}},
	volume = {35},
	doi = {https://doi.org/10.1609/aaai.v35i15.17605},
	booktitle = {Proceedings of the {AAAI} {Conference} on {Artificial} {Intelligence}},
	author = {{Madhura Pande} and {Aakriti Budhraja} and {Preksha Nema} and {Pratyush Kumar}},
	month = may,
	year = {2021},
	pages = {13613--13621},
}

@inproceedings{otmakhova_cross-linguistic_2022,
	address = {Seattle, Washington},
	title = {Cross-linguistic {Comparison} of {Linguistic} {Feature} {Encoding} in {BERT} {Models} for {Typologically} {Different} {Languages}},
	url = {https://aclanthology.org/2022.sigtyp-1.4/},
	doi = {10.18653/v1/2022.sigtyp-1.4},
	urldate = {2025-07-22},
	booktitle = {Proceedings of the 4th {Workshop} on {Research} in {Computational} {Linguistic} {Typology} and {Multilingual} {NLP}},
	publisher = {Association for Computational Linguistics},
	author = {Otmakhova, Yulia and Verspoor, Karin and Lau, Jey Han},
	editor = {Vylomova, Ekaterina and Ponti, Edoardo and Cotterell, Ryan},
	month = jul,
	year = {2022},
	pages = {27--35},
}

@inproceedings{oba_exploratory_2021,
	address = {Punta Cana, Dominican Republic},
	title = {Exploratory {Model} {Analysis} {Using} {Data}-{Driven} {Neuron} {Representations}},
	url = {https://aclanthology.org/2021.blackboxnlp-1.41/},
	doi = {10.18653/v1/2021.blackboxnlp-1.41},
	urldate = {2025-07-22},
	booktitle = {Proceedings of the {Fourth} {BlackboxNLP} {Workshop} on {Analyzing} and {Interpreting} {Neural} {Networks} for {NLP}},
	publisher = {Association for Computational Linguistics},
	author = {Oba, Daisuke and Yoshinaga, Naoki and Toyoda, Masashi},
	editor = {Bastings, Jasmijn and Belinkov, Yonatan and Dupoux, Emmanuel and Giulianelli, Mario and Hupkes, Dieuwke and Pinter, Yuval and Sajjad, Hassan},
	month = nov,
	year = {2021},
	pages = {518--528},
}

@inproceedings{oota_joint_2023,
	title = {Joint processing of linguistic properties in brains and language models},
	url = {https://www.microsoft.com/en-us/research/publication/joint-processing-of-linguistic-properties-in-brains-and-language-models/},
	language = {en-US},
	urldate = {2025-07-22},
	author = {Oota, Subba Reddy and Gupta, Manish and Toneva, Mariya},
	month = nov,
	year = {2023},
}

@inproceedings{niu_using_2022,
	address = {Abu Dhabi, United Arab Emirates (Hybrid)},
	title = {Using {Roark}-{Hollingshead} {Distance} to {Probe} {BERT}'s {Syntactic} {Competence}},
	url = {https://aclanthology.org/2022.blackboxnlp-1.27/},
	doi = {10.18653/v1/2022.blackboxnlp-1.27},
	urldate = {2025-07-22},
	booktitle = {Proceedings of the {Fifth} {BlackboxNLP} {Workshop} on {Analyzing} and {Interpreting} {Neural} {Networks} for {NLP}},
	publisher = {Association for Computational Linguistics},
	author = {Niu, Jingcheng and Lu, Wenjie and Corlett, Eric and Penn, Gerald},
	editor = {Bastings, Jasmijn and Belinkov, Yonatan and Elazar, Yanai and Hupkes, Dieuwke and Saphra, Naomi and Wiegreffe, Sarah},
	month = dec,
	year = {2022},
	pages = {325--334},
}

@inproceedings{niu_does_2022,
	address = {Gyeongju, Republic of Korea},
	title = {Does {BERT} {Rediscover} a {Classical} {NLP} {Pipeline}?},
	url = {https://aclanthology.org/2022.coling-1.278/},
	urldate = {2025-07-22},
	booktitle = {Proceedings of the 29th {International} {Conference} on {Computational} {Linguistics}},
	publisher = {International Committee on Computational Linguistics},
	author = {Niu, Jingcheng and Lu, Wenjie and Penn, Gerald},
	editor = {Calzolari, Nicoletta and Huang, Chu-Ren and Kim, Hansaem and Pustejovsky, James and Wanner, Leo and Choi, Key-Sun and Ryu, Pum-Mo and Chen, Hsin-Hsi and Donatelli, Lucia and Ji, Heng and Kurohashi, Sadao and Paggio, Patrizia and Xue, Nianwen and Kim, Seokhwan and Hahm, Younggyun and He, Zhong and Lee, Tony Kyungil and Santus, Enrico and Bond, Francis and Na, Seung-Hoon},
	month = oct,
	year = {2022},
	pages = {3143--3153},
}

@article{nikoulina_rediscovery_2021,
	title = {The {Rediscovery} {Hypothesis}: {Language} {Models} {Need} to {Meet} {Linguistics}},
	volume = {72},
	issn = {1076-9757},
	shorttitle = {The {Rediscovery} {Hypothesis}},
	url = {https://jair.org/index.php/jair/article/view/12788},
	doi = {10.1613/jair.1.12788},
	urldate = {2025-07-22},
	journal = {Journal of Artificial Intelligence Research},
	author = {Nikoulina, Vassilina and Tezekbayev, Maxat and Kozhakhmet, Nuradil and Babazhanova, Madina and Gallé, Matthias and Assylbekov, Zhenisbek},
	month = dec,
	year = {2021},
	pages = {1343--1384},
}

@inproceedings{newman_refining_2021,
	address = {Online},
	title = {Refining {Targeted} {Syntactic} {Evaluation} of {Language} {Models}},
	url = {https://aclanthology.org/2021.naacl-main.290/},
	doi = {10.18653/v1/2021.naacl-main.290},
	urldate = {2025-07-22},
	booktitle = {Proceedings of the 2021 {Conference} of the {North} {American} {Chapter} of the {Association} for {Computational} {Linguistics}: {Human} {Language} {Technologies}},
	publisher = {Association for Computational Linguistics},
	author = {Newman, Benjamin and Ang, Kai-Siang and Gong, Julia and Hewitt, John},
	editor = {Toutanova, Kristina and Rumshisky, Anna and Zettlemoyer, Luke and Hakkani-Tur, Dilek and Beltagy, Iz and Bethard, Steven and Cotterell, Ryan and Chakraborty, Tanmoy and Zhou, Yichao},
	month = jun,
	year = {2021},
	pages = {3710--3723},
}

@inproceedings{mohebbi_exploring_2021,
	address = {Online and Punta Cana, Dominican Republic},
	title = {Exploring the {Role} of {BERT} {Token} {Representations} to {Explain} {Sentence} {Probing} {Results}},
	url = {https://aclanthology.org/2021.emnlp-main.61/},
	doi = {10.18653/v1/2021.emnlp-main.61},
	urldate = {2025-07-21},
	booktitle = {Proceedings of the 2021 {Conference} on {Empirical} {Methods} in {Natural} {Language} {Processing}},
	publisher = {Association for Computational Linguistics},
	author = {Mohebbi, Hosein and Modarressi, Ali and Pilehvar, Mohammad Taher},
	editor = {Moens, Marie-Francine and Huang, Xuanjing and Specia, Lucia and Yih, Scott Wen-tau},
	month = nov,
	year = {2021},
	pages = {792--806},
}

@inproceedings{mueller_causal_2022,
	address = {Abu Dhabi, United Arab Emirates (Hybrid)},
	title = {Causal {Analysis} of {Syntactic} {Agreement} {Neurons} in {Multilingual} {Language} {Models}},
	url = {https://aclanthology.org/2022.conll-1.8/},
	doi = {10.18653/v1/2022.conll-1.8},
	urldate = {2025-07-21},
	booktitle = {Proceedings of the 26th {Conference} on {Computational} {Natural} {Language} {Learning} ({CoNLL})},
	publisher = {Association for Computational Linguistics},
	author = {Mueller, Aaron and Xia, Yu and Linzen, Tal},
	editor = {Fokkens, Antske and Srikumar, Vivek},
	month = dec,
	year = {2022},
	pages = {95--109},
}

@inproceedings{muller-eberstein_subspace_2023,
	address = {Singapore},
	title = {Subspace {Chronicles}: {How} {Linguistic} {Information} {Emerges}, {Shifts} and {Interacts} during {Language} {Model} {Training}},
	shorttitle = {Subspace {Chronicles}},
	url = {https://aclanthology.org/2023.findings-emnlp.879/},
	doi = {10.18653/v1/2023.findings-emnlp.879},
	urldate = {2025-07-21},
	booktitle = {Findings of the {Association} for {Computational} {Linguistics}: {EMNLP} 2023},
	publisher = {Association for Computational Linguistics},
	author = {Müller-Eberstein, Max and van der Goot, Rob and Plank, Barbara and Titov, Ivan},
	editor = {Bouamor, Houda and Pino, Juan and Bali, Kalika},
	month = dec,
	year = {2023},
	pages = {13190--13208},
}

@inproceedings{mysiak_is_2023,
	address = {Dubrovnik, Croatia},
	title = {Is {German} secretly a {Slavic} language? {What} {BERT} probing can tell us about language groups},
	shorttitle = {Is {German} secretly a {Slavic} language?},
	url = {https://aclanthology.org/2023.bsnlp-1.11/},
	doi = {10.18653/v1/2023.bsnlp-1.11},
	urldate = {2025-07-21},
	booktitle = {Proceedings of the 9th {Workshop} on {Slavic} {Natural} {Language} {Processing} 2023 ({SlavicNLP} 2023)},
	publisher = {Association for Computational Linguistics},
	author = {Mysiak, Aleksandra and Cyranka, Jacek},
	editor = {Piskorski, Jakub and Marcińczuk, Michał and Nakov, Preslav and Ogrodniczuk, Maciej and Pollak, Senja and Přibáň, Pavel and Rybak, Piotr and Steinberger, Josef and Yangarber, Roman},
	month = may,
	year = {2023},
	pages = {86--93},
}

@inproceedings{mikhailov_rusenteval_2021,
	address = {Kiyv, Ukraine},
	title = {{RuSentEval}: {Linguistic} {Source}, {Encoder} {Force}!},
	shorttitle = {{RuSentEval}},
	url = {https://aclanthology.org/2021.bsnlp-1.6/},
	urldate = {2025-07-21},
	booktitle = {Proceedings of the 8th {Workshop} on {Balto}-{Slavic} {Natural} {Language} {Processing}},
	publisher = {Association for Computational Linguistics},
	author = {Mikhailov, Vladislav and Taktasheva, Ekaterina and Sigdel, Elina and Artemova, Ekaterina},
	editor = {Babych, Bogdan and Kanishcheva, Olga and Nakov, Preslav and Piskorski, Jakub and Pivovarova, Lidia and Starko, Vasyl and Steinberger, Josef and Yangarber, Roman and Marcińczuk, Michał and Pollak, Senja and Přibáň, Pavel and Robnik-Šikonja, Marko},
	month = apr,
	year = {2021},
	pages = {43--65},
}

@inproceedings{mikhailov_morph_2021,
	address = {Online},
	title = {Morph {Call}: {Probing} {Morphosyntactic} {Content} of {Multilingual} {Transformers}},
	shorttitle = {Morph {Call}},
	url = {https://aclanthology.org/2021.sigtyp-1.10/},
	doi = {10.18653/v1/2021.sigtyp-1.10},
	urldate = {2025-07-21},
	booktitle = {Proceedings of the {Third} {Workshop} on {Computational} {Typology} and {Multilingual} {NLP}},
	publisher = {Association for Computational Linguistics},
	author = {Mikhailov, Vladislav and Serikov, Oleg and Artemova, Ekaterina},
	editor = {Vylomova, Ekaterina and Salesky, Elizabeth and Mielke, Sabrina and Lapesa, Gabriella and Kumar, Ritesh and Hammarström, Harald and Vulić, Ivan and Korhonen, Anna and Reichart, Roi and Ponti, Edoardo Maria and Cotterell, Ryan},
	month = jun,
	year = {2021},
	pages = {97--121},
}

@inproceedings{maudslay_tale_2020,
	address = {Online},
	title = {A {Tale} of a {Probe} and a {Parser}},
	url = {https://aclanthology.org/2020.acl-main.659/},
	doi = {10.18653/v1/2020.acl-main.659},
	urldate = {2025-07-21},
	booktitle = {Proceedings of the 58th {Annual} {Meeting} of the {Association} for {Computational} {Linguistics}},
	publisher = {Association for Computational Linguistics},
	author = {Maudslay, Rowan Hall and Valvoda, Josef and Pimentel, Tiago and Williams, Adina and Cotterell, Ryan},
	editor = {Jurafsky, Dan and Chai, Joyce and Schluter, Natalie and Tetreault, Joel},
	month = jul,
	year = {2020},
	pages = {7389--7395},
}

@inproceedings{sharma-etal-2023-learning,
    title = "Learning Non-linguistic Skills without Sacrificing Linguistic Proficiency",
    author = "Sharma, Mandar  and
      Muralidhar, Nikhil  and
      Ramakrishnan, Naren",
    editor = "Rogers, Anna  and
      Boyd-Graber, Jordan  and
      Okazaki, Naoaki",
    booktitle = "Proceedings of the 61st Annual Meeting of the Association for Computational Linguistics (Volume 1: Long Papers)",
    month = jul,
    year = "2023",
    address = "Toronto, Canada",
    publisher = "Association for Computational Linguistics",
    url = "https://aclanthology.org/2023.acl-long.340/",
    doi = "10.18653/v1/2023.acl-long.340",
    pages = "6178--6191",
}

@inproceedings{michael_asking_2020,
	address = {Online},
	title = {Asking without {Telling}: {Exploring} {Latent} {Ontologies} in {Contextual} {Representations}},
	shorttitle = {Asking without {Telling}},
	url = {https://aclanthology.org/2020.emnlp-main.552/},
	doi = {10.18653/v1/2020.emnlp-main.552},
	urldate = {2025-07-21},
	booktitle = {Proceedings of the 2020 {Conference} on {Empirical} {Methods} in {Natural} {Language} {Processing} ({EMNLP})},
	publisher = {Association for Computational Linguistics},
	author = {Michael, Julian and Botha, Jan A. and Tenney, Ian},
	editor = {Webber, Bonnie and Cohn, Trevor and He, Yulan and Liu, Yang},
	month = nov,
	year = {2020},
	pages = {6792--6812},
}

@inproceedings{miaschi-etal-2021-probing,
    title = "Probing Tasks Under Pressure",
    author = "Miaschi, Alessio  and
      Alzetta, Chiara  and
      Brunato, Dominique  and
      Dell{'}Orletta, Felice  and
      Venturi, Giulia",
    editor = "Fersini, Elisabetta  and
      Passarotti, Marco  and
      Patti, Viviana",
    booktitle = "Proceedings of the Eighth Italian Conference on Computational Linguistics (CLiC-it 2021)",
    month = jun,
    year = "2021",
    address = "Milan, Italy",
    publisher = "CEUR Workshop Proceedings",
    url = "https://aclanthology.org/2021.clicit-1.36/",
    pages = "230--236",
    ISBN = "979-12-80136-94-7"
}

@incollection{miaschi_italian_2020,
	address = {Torino},
	series = {Collana dell'{Associazione} {Italiana} di {Linguistica} {Computazionale}},
	title = {Italian {Transformers} {Under} the {Linguistic} {Lens}},
	copyright = {https://www.openedition.org/12554},
	isbn = {979-12-80136-33-6},
	url = {https://books.openedition.org/aaccademia/8745},
	language = {en},
	urldate = {2025-07-21},
	booktitle = {Proceedings of the {Seventh} {Italian} {Conference} on {Computational} {Linguistics} {CLiC}-it 2020 : {Bologna}, {Italy}, {March} 1-3, 2021},
	publisher = {Accademia University Press},
	author = {Miaschi, Alessio and Sarti, Gabriele and Brunato, Dominique and Dell’Orletta, Felice and Venturi, Giulia},
	editor = {Dell'Orletta, Felice and Monti, Johanna and Tamburini, Fabio},
	year = {2020},
	pages = {310--316},
}

@inproceedings{miaschi_contextual_2020,
	address = {Online},
	title = {Contextual and {Non}-{Contextual} {Word} {Embeddings}: an in-depth {Linguistic} {Investigation}},
	shorttitle = {Contextual and {Non}-{Contextual} {Word} {Embeddings}},
	url = {https://aclanthology.org/2020.repl4nlp-1.15/},
	doi = {10.18653/v1/2020.repl4nlp-1.15},
	urldate = {2025-07-21},
	booktitle = {Proceedings of the 5th {Workshop} on {Representation} {Learning} for {NLP}},
	publisher = {Association for Computational Linguistics},
	author = {Miaschi, Alessio and Dell'Orletta, Felice},
	editor = {Gella, Spandana and Welbl, Johannes and Rei, Marek and Petroni, Fabio and Lewis, Patrick and Strubell, Emma and Seo, Minjoon and Hajishirzi, Hannaneh},
	month = jul,
	year = {2020},
	pages = {110--119},
}

@article{miaschi_testing_2023,
	title = {Testing the {Effectiveness} of the {Diagnostic} {Probing} {Paradigm} on {Italian} {Treebanks}},
	volume = {3},
	url = {http://ouci.dntb.gov.ua/en/works/9jAW0VNl/},
	doi = {10.3390/info14030144},
	language = {en},
	urldate = {2025-07-21},
	author = {Miaschi, Alessio and {Alzetta,  Chiara} and {Dominique Brunato} and {Felice Dell’Orletta} and {Giulia Venturi}},
	year = {2023},
	pages = {144},
}

@inproceedings{marecek_balustrades_2019,
	address = {Florence, Italy},
	title = {From {Balustrades} to {Pierre} {Vinken}: {Looking} for {Syntax} in {Transformer} {Self}-{Attentions}},
	shorttitle = {From {Balustrades} to {Pierre} {Vinken}},
	url = {https://aclanthology.org/W19-4827/},
	doi = {10.18653/v1/W19-4827},
	urldate = {2025-07-21},
	booktitle = {Proceedings of the 2019 {ACL} {Workshop} {BlackboxNLP}: {Analyzing} and {Interpreting} {Neural} {Networks} for {NLP}},
	publisher = {Association for Computational Linguistics},
	author = {Mareček, David and Rosa, Rudolf},
	editor = {Linzen, Tal and Chrupała, Grzegorz and Belinkov, Yonatan and Hupkes, Dieuwke},
	month = aug,
	year = {2019},
	pages = {263--275},
}

@inproceedings{luo_have_2021,
	address = {Online},
	title = {Have {Attention} {Heads} in {BERT} {Learned} {Constituency} {Grammar}?},
	url = {https://aclanthology.org/2021.eacl-srw.2/},
	doi = {10.18653/v1/2021.eacl-srw.2},
	urldate = {2025-07-21},
	booktitle = {Proceedings of the 16th {Conference} of the {European} {Chapter} of the {Association} for {Computational} {Linguistics}: {Student} {Research} {Workshop}},
	publisher = {Association for Computational Linguistics},
	author = {Luo, Ziyang},
	editor = {Sorodoc, Ionut-Teodor and Sushil, Madhumita and Takmaz, Ece and Agirre, Eneko},
	month = apr,
	year = {2021},
	pages = {8--15},
}

@inproceedings{liu_probing_2021,
	address = {Punta Cana, Dominican Republic},
	title = {Probing {Across} {Time}: {What} {Does} {RoBERTa} {Know} and {When}?},
	shorttitle = {Probing {Across} {Time}},
	url = {https://aclanthology.org/2021.findings-emnlp.71/},
	doi = {10.18653/v1/2021.findings-emnlp.71},
	urldate = {2025-07-21},
	booktitle = {Findings of the {Association} for {Computational} {Linguistics}: {EMNLP} 2021},
	publisher = {Association for Computational Linguistics},
	author = {Liu, Zeyu and Wang, Yizhong and Kasai, Jungo and Hajishirzi, Hannaneh and Smith, Noah A.},
	editor = {Moens, Marie-Francine and Huang, Xuanjing and Specia, Lucia and Yih, Scott Wen-tau},
	month = nov,
	year = {2021},
	pages = {820--842},
}

@inproceedings{Lovering2021PredictingIB,
  title={Predicting Inductive Biases of Pre-Trained Models},
  author={Charles Lovering and Rohan Jha and Tal Linzen and Ellie Pavlick},
  booktitle={International Conference on Learning Representations},
  year={2021},
  url={https://api.semanticscholar.org/CorpusID:235614282}
}

@inproceedings{lin_open_2019,
	address = {Florence, Italy},
	title = {Open {Sesame}: {Getting} inside {BERT}'s {Linguistic} {Knowledge}},
	shorttitle = {Open {Sesame}},
	url = {https://aclanthology.org/W19-4825/},
	doi = {10.18653/v1/W19-4825},
	urldate = {2025-07-21},
	booktitle = {Proceedings of the 2019 {ACL} {Workshop} {BlackboxNLP}: {Analyzing} and {Interpreting} {Neural} {Networks} for {NLP}},
	publisher = {Association for Computational Linguistics},
	author = {Lin, Yongjie and Tan, Yi Chern and Frank, Robert},
	editor = {Linzen, Tal and Chrupała, Grzegorz and Belinkov, Yonatan and Hupkes, Dieuwke},
	month = aug,
	year = {2019},
	pages = {241--253},
}

@inproceedings{liu_linguistic_2019,
    title = "Linguistic Knowledge and Transferability of Contextual Representations",
    author = "Liu, Nelson F.  and
      Gardner, Matt  and
      Belinkov, Yonatan  and
      Peters, Matthew E.  and
      Smith, Noah A.",
    editor = "Burstein, Jill  and
      Doran, Christy  and
      Solorio, Thamar",
    booktitle = "Proceedings of the 2019 Conference of the North {A}merican Chapter of the Association for Computational Linguistics: Human Language Technologies, Volume 1 (Long and Short Papers)",
    month = jun,
    year = "2019",
    address = "Minneapolis, Minnesota",
    publisher = "Association for Computational Linguistics",
    url = "https://aclanthology.org/N19-1112/",
    doi = "10.18653/v1/N19-1112",
    pages = "1073--1094"
}

@inproceedings{li_how_2021,
	address = {Online},
	title = {How is {BERT} surprised? {Layerwise} detection of linguistic anomalies},
	shorttitle = {How is {BERT} surprised?},
	url = {https://aclanthology.org/2021.acl-long.325/},
	doi = {10.18653/v1/2021.acl-long.325},
	urldate = {2025-07-21},
	booktitle = {Proceedings of the 59th {Annual} {Meeting} of the {Association} for {Computational} {Linguistics} and the 11th {International} {Joint} {Conference} on {Natural} {Language} {Processing} ({Volume} 1: {Long} {Papers})},
	publisher = {Association for Computational Linguistics},
	author = {Li, Bai and Zhu, Zining and Thomas, Guillaume and Xu, Yang and Rudzicz, Frank},
	editor = {Zong, Chengqing and Xia, Fei and Li, Wenjie and Navigli, Roberto},
	month = aug,
	year = {2021},
	pages = {4215--4228},
}

@inproceedings{li_probing_2022,
	address = {Seattle, United States},
	title = {Probing via {Prompting}},
	url = {https://aclanthology.org/2022.naacl-main.84/},
	doi = {10.18653/v1/2022.naacl-main.84},
	urldate = {2025-07-21},
	booktitle = {Proceedings of the 2022 {Conference} of the {North} {American} {Chapter} of the {Association} for {Computational} {Linguistics}: {Human} {Language} {Technologies}},
	publisher = {Association for Computational Linguistics},
	author = {Li, Jiaoda and Cotterell, Ryan and Sachan, Mrinmaya},
	editor = {Carpuat, Marine and de Marneffe, Marie-Catherine and Meza Ruiz, Ivan Vladimir},
	month = jul,
	year = {2022},
	pages = {1144--1157},
}

@inproceedings{limisiewicz_universal_2020,
	address = {Online},
	title = {Universal {Dependencies} {According} to {BERT}: {Both} {More} {Specific} and {More} {General}},
	shorttitle = {Universal {Dependencies} {According} to {BERT}},
	url = {https://aclanthology.org/2020.findings-emnlp.245/},
	doi = {10.18653/v1/2020.findings-emnlp.245},
	urldate = {2025-07-21},
	booktitle = {Findings of the {Association} for {Computational} {Linguistics}: {EMNLP} 2020},
	publisher = {Association for Computational Linguistics},
	author = {Limisiewicz, Tomasz and Mareček, David and Rosa, Rudolf},
	editor = {Cohn, Trevor and He, Yulan and Liu, Yang},
	month = nov,
	year = {2020},
	pages = {2710--2722},
}

@article{lee_bert_2022,
	title = {({AL}){BERT} {Down} the {Garden} {Path}: {Psycholinguistic} {Experiments} for {Pre}-trained {Language} {Models}},
	volume = {22},
	doi = {10.15738/kjell.22..202210.1033},
	journal = {Korean Journal of English Language and Linguistics},
	author = {Lee, Jonghyun and Shin, Jeong Ah and Park, Myung Kwan},
	year = {2022},
	pages = {1033--1050},
}

@article{lee_decoding_2023,
	title = {Decoding {BERT}’s {Internal} {Processing} of {Garden}-{Path} {Structures} through {Attention} {Maps}},
	volume = {23},
	doi = {10.15738/kjell.23..202306.461},
	journal = {Korean Journal of English Language and Linguistics},
	author = {Lee, Jonghyun and Shin, Jeong-Ah},
	month = jan,
	year = {2023},
	pages = {461--481},
}

@inproceedings{kunz_where_2022,
	address = {Gyeongju, Republic of Korea},
	title = {Where {Does} {Linguistic} {Information} {Emerge} in {Neural} {Language} {Models}? {Measuring} {Gains} and {Contributions} across {Layers}},
	shorttitle = {Where {Does} {Linguistic} {Information} {Emerge} in {Neural} {Language} {Models}?},
	url = {https://aclanthology.org/2022.coling-1.413/},
	urldate = {2025-07-21},
	booktitle = {Proceedings of the 29th {International} {Conference} on {Computational} {Linguistics}},
	publisher = {International Committee on Computational Linguistics},
	author = {Kunz, Jenny and Kuhlmann, Marco},
	editor = {Calzolari, Nicoletta and Huang, Chu-Ren and Kim, Hansaem and Pustejovsky, James and Wanner, Leo and Choi, Key-Sun and Ryu, Pum-Mo and Chen, Hsin-Hsi and Donatelli, Lucia and Ji, Heng and Kurohashi, Sadao and Paggio, Patrizia and Xue, Nianwen and Kim, Seokhwan and Hahm, Younggyun and He, Zhong and Lee, Tony Kyungil and Santus, Enrico and Bond, Francis and Na, Seung-Hoon},
	month = oct,
	year = {2022},
	pages = {4664--4676},
}

@inproceedings{kunz_classifier_2020,
	address = {Barcelona, Spain (Online)},
	title = {Classifier {Probes} {May} {Just} {Learn} from {Linear} {Context} {Features}},
	url = {https://aclanthology.org/2020.coling-main.450/},
	doi = {10.18653/v1/2020.coling-main.450},
	urldate = {2025-07-21},
	booktitle = {Proceedings of the 28th {International} {Conference} on {Computational} {Linguistics}},
	publisher = {International Committee on Computational Linguistics},
	author = {Kunz, Jenny and Kuhlmann, Marco},
	editor = {Scott, Donia and Bel, Nuria and Zong, Chengqing},
	month = dec,
	year = {2020},
	pages = {5136--5146},
}

@inproceedings{kulmizev_neural_2020,
	address = {Online},
	title = {Do {Neural} {Language} {Models} {Show} {Preferences} for {Syntactic} {Formalisms}?},
	url = {https://aclanthology.org/2020.acl-main.375/},
	doi = {10.18653/v1/2020.acl-main.375},
	urldate = {2025-07-21},
	booktitle = {Proceedings of the 58th {Annual} {Meeting} of the {Association} for {Computational} {Linguistics}},
	publisher = {Association for Computational Linguistics},
	author = {Kulmizev, Artur and Ravishankar, Vinit and Abdou, Mostafa and Nivre, Joakim},
	editor = {Jurafsky, Dan and Chai, Joyce and Schluter, Natalie and Tetreault, Joel},
	month = jul,
	year = {2020},
	pages = {4077--4091},
}

@inproceedings{klafka_spying_2020,
	address = {Online},
	title = {Spying on {Your} {Neighbors}: {Fine}-grained {Probing} of {Contextual} {Embeddings} for {Information} about {Surrounding} {Words}},
	shorttitle = {Spying on {Your} {Neighbors}},
	url = {https://aclanthology.org/2020.acl-main.434/},
	doi = {10.18653/v1/2020.acl-main.434},
	urldate = {2025-07-21},
	booktitle = {Proceedings of the 58th {Annual} {Meeting} of the {Association} for {Computational} {Linguistics}},
	publisher = {Association for Computational Linguistics},
	author = {Klafka, Josef and Ettinger, Allyson},
	editor = {Jurafsky, Dan and Chai, Joyce and Schluter, Natalie and Tetreault, Joel},
	month = jul,
	year = {2020},
	pages = {4801--4811},
}

@article{kasthuriarachchy_general_2021,
	title = {From {General} {Language} {Understanding} to {Noisy} {Text} {Comprehension}},
	volume = {11},
	url = {http://ouci.dntb.gov.ua/en/works/lR8GEzG7/},
	doi = {10.3390/app11177814},
	language = {en},
	number = {27},
	urldate = {2025-07-21},
	journal = {Applied Sciences},
	author = {Kasthuriarachchy, Buddhika and Chetty, Madhu and Shatte, Adrian and Walls, Darren},
	month = aug,
	year = {2021},
}

@inproceedings{jo_roles_2020,
	address = {Online},
	title = {Roles and {Utilization} of {Attention} {Heads} in {Transformer}-based {Neural} {Language} {Models}},
	url = {https://aclanthology.org/2020.acl-main.311/},
	doi = {10.18653/v1/2020.acl-main.311},
	urldate = {2025-07-21},
	booktitle = {Proceedings of the 58th {Annual} {Meeting} of the {Association} for {Computational} {Linguistics}},
	publisher = {Association for Computational Linguistics},
	author = {Jo, Jae-young and Myaeng, Sung-Hyon},
	editor = {Jurafsky, Dan and Chai, Joyce and Schluter, Natalie and Tetreault, Joel},
	month = jul,
	year = {2020},
	pages = {3404--3417},
}

@inproceedings{hou_birds_2021,
	address = {Online},
	title = {Bird's {Eye}: {Probing} for {Linguistic} {Graph} {Structures} with a {Simple} {Information}-{Theoretic} {Approach}},
	shorttitle = {Bird's {Eye}},
	url = {https://aclanthology.org/2021.acl-long.145/},
	doi = {10.18653/v1/2021.acl-long.145},
	urldate = {2025-07-21},
	booktitle = {Proceedings of the 59th {Annual} {Meeting} of the {Association} for {Computational} {Linguistics} and the 11th {International} {Joint} {Conference} on {Natural} {Language} {Processing} ({Volume} 1: {Long} {Papers})},
	publisher = {Association for Computational Linguistics},
	author = {Hou, Yifan and Sachan, Mrinmaya},
	editor = {Zong, Chengqing and Xia, Fei and Li, Wenjie and Navigli, Roberto},
	month = aug,
	year = {2021},
	pages = {1844--1859},
}

@inproceedings{hewitt_structural_2019,
	address = {Minneapolis, Minnesota},
	title = {A {Structural} {Probe} for {Finding} {Syntax} in {Word} {Representations}},
	url = {https://aclanthology.org/N19-1419/},
	doi = {10.18653/v1/N19-1419},
	urldate = {2025-07-21},
	booktitle = {Proceedings of the 2019 {Conference} of the {North} {American} {Chapter} of the {Association} for {Computational} {Linguistics}: {Human} {Language} {Technologies}, {Volume} 1 ({Long} and {Short} {Papers})},
	publisher = {Association for Computational Linguistics},
	author = {Hewitt, John and Manning, Christopher D.},
	editor = {Burstein, Jill and Doran, Christy and Solorio, Thamar},
	month = jun,
	year = {2019},
	pages = {4129--4138},
}

@inproceedings{hessel_how_2021,
	address = {Online},
	title = {How effective is {BERT} without word ordering? {Implications} for language understanding and data privacy},
	shorttitle = {How effective is {BERT} without word ordering?},
	url = {https://aclanthology.org/2021.acl-short.27/},
	doi = {10.18653/v1/2021.acl-short.27},
	urldate = {2025-07-21},
	booktitle = {Proceedings of the 59th {Annual} {Meeting} of the {Association} for {Computational} {Linguistics} and the 11th {International} {Joint} {Conference} on {Natural} {Language} {Processing} ({Volume} 2: {Short} {Papers})},
	publisher = {Association for Computational Linguistics},
	author = {Hessel, Jack and Schofield, Alexandra},
	editor = {Zong, Chengqing and Xia, Fei and Li, Wenjie and Navigli, Roberto},
	month = aug,
	year = {2021},
	pages = {204--211},
}

@inproceedings{hernandez_low-dimensional_2021,
	address = {Online},
	title = {The {Low}-{Dimensional} {Linear} {Geometry} of {Contextualized} {Word} {Representations}},
	url = {https://aclanthology.org/2021.conll-1.7/},
	doi = {10.18653/v1/2021.conll-1.7},
	urldate = {2025-07-21},
	booktitle = {Proceedings of the 25th {Conference} on {Computational} {Natural} {Language} {Learning}},
	publisher = {Association for Computational Linguistics},
	author = {Hernandez, Evan and Andreas, Jacob},
	editor = {Bisazza, Arianna and Abend, Omri},
	month = nov,
	year = {2021},
	pages = {82--93},
}

@misc{gupta_deep_2021,
	title = {Deep {Clustering} of {Text} {Representations} for {Supervision}-free {Probing} of {Syntax}},
	url = {http://arxiv.org/abs/2010.12784},
	doi = {10.48550/arXiv.2010.12784},
	urldate = {2025-07-21},
	publisher = {arXiv},
	author = {Gupta, Vikram and Shi, Haoyue and Gimpel, Kevin and Sachan, Mrinmaya},
	month = dec,
	year = {2021},
	note = {arXiv:2010.12784},
}

@article{guarasci_assessing_2023,
	title = {Assessing {BERT}’s ability to learn {Italian} syntax: a study on null-subject and agreement phenomena},
	volume = {14},
	issn = {1868-5145},
	shorttitle = {Assessing {BERT}’s ability to learn {Italian} syntax},
	url = {https://doi.org/10.1007/s12652-021-03297-4},
	doi = {10.1007/s12652-021-03297-4},
	language = {en},
	number = {1},
	urldate = {2025-07-21},
	journal = {Journal of Ambient Intelligence and Humanized Computing},
	author = {Guarasci, Raffaele and Silvestri, Stefano and De Pietro, Giuseppe and Fujita, Hamido and Esposito, Massimo},
	month = jan,
	year = {2023},
	pages = {289--303},
}

@article{guarasci_bert_2022,
	title = {{BERT} syntactic transfer: {A} computational experiment on {Italian}, {French} and {English} languages},
	volume = {71},
	issn = {0885-2308},
	shorttitle = {{BERT} syntactic transfer},
	url = {https://www.sciencedirect.com/science/article/pii/S0885230821000681},
	doi = {10.1016/j.csl.2021.101261},
	urldate = {2025-07-21},
	journal = {Computer Speech \& Language},
	author = {Guarasci, Raffaele and Silvestri, Stefano and De Pietro, Giuseppe and Fujita, Hamido and Esposito, Massimo},
	month = jan,
	year = {2022},
	pages = {101261},
}

@inproceedings{glavas_is_2021,
	address = {Online},
	title = {Is {Supervised} {Syntactic} {Parsing} {Beneficial} for {Language} {Understanding} {Tasks}? {An} {Empirical} {Investigation}},
	shorttitle = {Is {Supervised} {Syntactic} {Parsing} {Beneficial} for {Language} {Understanding} {Tasks}?},
	url = {https://aclanthology.org/2021.eacl-main.270/},
	doi = {10.18653/v1/2021.eacl-main.270},
	urldate = {2025-07-21},
	booktitle = {Proceedings of the 16th {Conference} of the {European} {Chapter} of the {Association} for {Computational} {Linguistics}: {Main} {Volume}},
	publisher = {Association for Computational Linguistics},
	author = {Glavaš, Goran and Vulić, Ivan},
	editor = {Merlo, Paola and Tiedemann, Jorg and Tsarfaty, Reut},
	month = apr,
	year = {2021},
	pages = {3090--3104},
}

@inproceedings{davis_discourse_2020,
	address = {Online},
	title = {Discourse structure interacts with reference but not syntax in neural language models},
	url = {https://aclanthology.org/2020.conll-1.32/},
	doi = {10.18653/v1/2020.conll-1.32},
	urldate = {2025-07-21},
	booktitle = {Proceedings of the 24th {Conference} on {Computational} {Natural} {Language} {Learning}},
	publisher = {Association for Computational Linguistics},
	author = {Davis, Forrest and van Schijndel, Marten},
	editor = {Fernández, Raquel and Linzen, Tal},
	month = nov,
	year = {2020},
	pages = {396--407},
}

@inproceedings{durrani_analyzing_2020,
	address = {Online},
	title = {Analyzing {Individual} {Neurons} in {Pre}-trained {Language} {Models}},
	url = {https://aclanthology.org/2020.emnlp-main.395/},
	doi = {10.18653/v1/2020.emnlp-main.395},
	urldate = {2025-07-21},
	booktitle = {Proceedings of the 2020 {Conference} on {Empirical} {Methods} in {Natural} {Language} {Processing} ({EMNLP})},
	publisher = {Association for Computational Linguistics},
	author = {Durrani, Nadir and Sajjad, Hassan and Dalvi, Fahim and Belinkov, Yonatan},
	editor = {Webber, Bonnie and Cohn, Trevor and He, Yulan and Liu, Yang},
	month = nov,
	year = {2020},
	pages = {4865--4880},
}

@inproceedings{conia_probing_2022,
	address = {Dublin, Ireland},
	title = {Probing for {Predicate} {Argument} {Structures} in {Pretrained} {Language} {Models}},
	url = {https://aclanthology.org/2022.acl-long.316/},
	doi = {10.18653/v1/2022.acl-long.316},
	urldate = {2025-07-21},
	booktitle = {Proceedings of the 60th {Annual} {Meeting} of the {Association} for {Computational} {Linguistics} ({Volume} 1: {Long} {Papers})},
	publisher = {Association for Computational Linguistics},
	author = {Conia, Simone and Navigli, Roberto},
	editor = {Muresan, Smaranda and Nakov, Preslav and Villavicencio, Aline},
	month = may,
	year = {2022},
	pages = {4622--4632},
}

@inproceedings{clark_what_2019,
	address = {Florence, Italy},
	title = {What {Does} {BERT} {Look} at? {An} {Analysis} of {BERT}'s {Attention}},
	shorttitle = {What {Does} {BERT} {Look} at?},
	url = {https://aclanthology.org/W19-4828/},
	doi = {10.18653/v1/W19-4828},
	urldate = {2025-07-21},
	booktitle = {Proceedings of the 2019 {ACL} {Workshop} {BlackboxNLP}: {Analyzing} and {Interpreting} {Neural} {Networks} for {NLP}},
	publisher = {Association for Computational Linguistics},
	author = {Clark, Kevin and Khandelwal, Urvashi and Levy, Omer and Manning, Christopher D.},
	editor = {Linzen, Tal and Chrupała, Grzegorz and Belinkov, Yonatan and Hupkes, Dieuwke},
	month = aug,
	year = {2019},
	pages = {276--286},
}

@inproceedings{chrupala_correlating_2019,
	address = {Florence, Italy},
	title = {Correlating {Neural} and {Symbolic} {Representations} of {Language}},
	url = {https://aclanthology.org/P19-1283/},
	doi = {10.18653/v1/P19-1283},
	urldate = {2025-07-21},
	booktitle = {Proceedings of the 57th {Annual} {Meeting} of the {Association} for {Computational} {Linguistics}},
	publisher = {Association for Computational Linguistics},
	author = {Chrupała, Grzegorz and Alishahi, Afra},
	editor = {Korhonen, Anna and Traum, David and Màrquez, Lluís},
	month = jul,
	year = {2019},
	pages = {2952--2962},
}

@article{choenni_investigating_2022,
	title = {Investigating {Language} {Relationships} in {Multilingual} {Sentence} {Encoders} {Through} the {Lens} of {Linguistic} {Typology}},
	volume = {48},
	url = {https://aclanthology.org/2022.cl-3.5/},
	doi = {10.1162/coli_a_00444},
	number = {3},
	urldate = {2025-07-21},
	journal = {Computational Linguistics},
	author = {Choenni, Rochelle and Shutova, Ekaterina},
	month = sep,
	year = {2022},
	pages = {635--672},
}

@inproceedings{chi_finding_2020,
	address = {Online},
	title = {Finding {Universal} {Grammatical} {Relations} in {Multilingual} {BERT}},
	url = {https://aclanthology.org/2020.acl-main.493/},
	doi = {10.18653/v1/2020.acl-main.493},
	urldate = {2025-07-21},
	booktitle = {Proceedings of the 58th {Annual} {Meeting} of the {Association} for {Computational} {Linguistics}},
	publisher = {Association for Computational Linguistics},
	author = {Chi, Ethan A. and Hewitt, John and Manning, Christopher D.},
	editor = {Jurafsky, Dan and Chai, Joyce and Schluter, Natalie and Tetreault, Joel},
	month = jul,
	year = {2020},
	pages = {5564--5577},
}

@inproceedings{celikkanat_controlling_2020,
	address = {Online},
	title = {Controlling the {Imprint} of {Passivization} and {Negation} in {Contextualized} {Representations}},
	url = {https://aclanthology.org/2020.blackboxnlp-1.13/},
	doi = {10.18653/v1/2020.blackboxnlp-1.13},
	urldate = {2025-07-21},
	booktitle = {Proceedings of the {Third} {BlackboxNLP} {Workshop} on {Analyzing} and {Interpreting} {Neural} {Networks} for {NLP}},
	publisher = {Association for Computational Linguistics},
	author = {Celikkanat, Hande and Virpioja, Sami and Tiedemann, Jörg and Apidianaki, Marianna},
	editor = {Alishahi, Afra and Belinkov, Yonatan and Chrupała, Grzegorz and Hupkes, Dieuwke and Pinter, Yuval and Sajjad, Hassan},
	month = nov,
	year = {2020},
	pages = {136--148},
}

@inproceedings{bolucu_analysing_2022,
	address = {New Delhi, India},
	title = {Analysing {Syntactic} and {Semantic} {Features} in {Pre}-trained {Language} {Models} in a {Fully} {Unsupervised} {Setting}},
	url = {https://aclanthology.org/2022.icon-main.3/},
	urldate = {2025-07-21},
	booktitle = {Proceedings of the 19th {International} {Conference} on {Natural} {Language} {Processing} ({ICON})},
	publisher = {Association for Computational Linguistics},
	author = {Bölücü, Necva and Can, Burcu},
	editor = {Akhtar, Md. Shad and Chakraborty, Tanmoy},
	month = dec,
	year = {2022},
	pages = {19--31},
}

@inproceedings{
chen2021probing,
title={Probing {\{}BERT{\}} in Hyperbolic Spaces},
author={Boli Chen and Yao Fu and Guangwei Xu and Pengjun Xie and Chuanqi Tan and Mosha Chen and Liping Jing},
booktitle={International Conference on Learning Representations},
year={2021},
url={https://openreview.net/forum?id=17VnwXYZyhH}
}

@inproceedings{auyespek_hyperbolic_2022,
	title = {Hyperbolic {Embedding} for {Finding} {Syntax} in {BERT}},
	volume = {3078},
	booktitle = {International {Conference} of the {Italian} {Association} for {Artificial} {Intelligence}, {AIxIA} 2021 {DP} - {Virtual}, {Online}},
	author = {Auyespek, Temirlan and Mach, Thomas and Zhenisbek, Assylbekov},
	year = {2022},
}

@inproceedings{arps_probing_2022,
	address = {Abu Dhabi, United Arab Emirates},
	title = {Probing for {Constituency} {Structure} in {Neural} {Language} {Models}},
	url = {https://aclanthology.org/2022.findings-emnlp.502/},
	doi = {10.18653/v1/2022.findings-emnlp.502},
	urldate = {2025-07-21},
	booktitle = {Findings of the {Association} for {Computational} {Linguistics}: {EMNLP} 2022},
	publisher = {Association for Computational Linguistics},
	author = {Arps, David and Samih, Younes and Kallmeyer, Laura and Sajjad, Hassan},
	editor = {Goldberg, Yoav and Kozareva, Zornitsa and Zhang, Yue},
	month = dec,
	year = {2022},
	pages = {6738--6757},
}

@article{amini_naturalistic_2023,
	title = {Naturalistic {Causal} {Probing} for {Morpho}-{Syntax}},
	volume = {11},
	url = {https://aclanthology.org/2023.tacl-1.23/},
	doi = {10.1162/tacl_a_00554},
	urldate = {2025-07-21},
	journal = {Transactions of the Association for Computational Linguistics},
	author = {Amini, Afra and Pimentel, Tiago and Meister, Clara and Cotterell, Ryan},
	year = {2023},
	pages = {384--403},
}

@inproceedings{alleman_syntactic_2021,
	address = {Online},
	title = {Syntactic {Perturbations} {Reveal} {Representational} {Correlates} of {Hierarchical} {Phrase} {Structure} in {Pretrained} {Language} {Models}},
	url = {https://aclanthology.org/2021.repl4nlp-1.27/},
	doi = {10.18653/v1/2021.repl4nlp-1.27},
	urldate = {2025-07-21},
	booktitle = {Proceedings of the 6th {Workshop} on {Representation} {Learning} for {NLP} ({RepL4NLP}-2021)},
	publisher = {Association for Computational Linguistics},
	author = {Alleman, Matteo and Mamou, Jonathan and A Del Rio, Miguel and Tang, Hanlin and Kim, Yoon and Chung, SueYeon},
	editor = {Rogers, Anna and Calixto, Iacer and Vulić, Ivan and Saphra, Naomi and Kassner, Nora and Camburu, Oana-Maria and Bansal, Trapit and Shwartz, Vered},
	month = aug,
	year = {2021},
	pages = {263--276},
}

@inproceedings{alajrami_how_2022,
	address = {Dublin, Ireland},
	title = {How does the pre-training objective affect what large language models learn about linguistic properties?},
	url = {https://aclanthology.org/2022.acl-short.16/},
	doi = {10.18653/v1/2022.acl-short.16},
	urldate = {2025-07-21},
	booktitle = {Proceedings of the 60th {Annual} {Meeting} of the {Association} for {Computational} {Linguistics} ({Volume} 2: {Short} {Papers})},
	publisher = {Association for Computational Linguistics},
	author = {Alajrami, Ahmed and Aletras, Nikolaos},
	editor = {Muresan, Smaranda and Nakov, Preslav and Villavicencio, Aline},
	month = may,
	year = {2022},
	pages = {131--147},
}

@article{acs_morphosyntactic_2024,
	title = {Morphosyntactic probing of multilingual {BERT} models},
	volume = {30},
	issn = {1351-3249, 1469-8110},
	url = {https://www.cambridge.org/core/journals/natural-language-engineering/article/morphosyntactic-probing-of-multilingual-bert-models/8C0D539D3F11FB188AB73228BA7F5805},
	doi = {10.1017/S1351324923000190},
	language = {en},
	number = {4},
	urldate = {2025-07-21},
	journal = {Natural Language Engineering},
	author = {Acs, Judit and Hamerlik, Endre and Schwartz, Roy and Smith, Noah A. and Kornai, Andras},
	month = jul,
	year = {2024},
	pages = {753--792},
}

@inproceedings{abdou_higher-order_2019,
	address = {Hong Kong, China},
	title = {Higher-order {Comparisons} of {Sentence} {Encoder} {Representations}},
	url = {https://aclanthology.org/D19-1593/},
	doi = {10.18653/v1/D19-1593},
	urldate = {2025-07-16},
	booktitle = {Proceedings of the 2019 {Conference} on {Empirical} {Methods} in {Natural} {Language} {Processing} and the 9th {International} {Joint} {Conference} on {Natural} {Language} {Processing} ({EMNLP}-{IJCNLP})},
	publisher = {Association for Computational Linguistics},
	author = {Abdou, Mostafa and Kulmizev, Artur and Hill, Felix and Low, Daniel M. and Søgaard, Anders},
	editor = {Inui, Kentaro and Jiang, Jing and Ng, Vincent and Wan, Xiaojun},
	month = nov,
	year = {2019},
	pages = {5838--5845},
}

@inproceedings{tjuatja_syntax_2023,
	address = {Toronto, Canada},
	title = {Syntax and {Semantics} {Meet} in the “{Middle}”: {Probing} the {Syntax}-{Semantics} {Interface} of {LMs} {Through} {Agentivity}},
	shorttitle = {Syntax and {Semantics} {Meet} in the “{Middle}”},
	url = {https://aclanthology.org/2023.starsem-1.14/},
	doi = {10.18653/v1/2023.starsem-1.14},
	urldate = {2025-07-16},
	booktitle = {Proceedings of the 12th {Joint} {Conference} on {Lexical} and {Computational} {Semantics} (*{SEM} 2023)},
	publisher = {Association for Computational Linguistics},
	author = {Tjuatja, Lindia and Liu, Emmy and Levin, Lori and Neubig, Graham},
	editor = {Palmer, Alexis and Camacho-collados, Jose},
	month = jul,
	year = {2023},
	pages = {149--164},
}

@inproceedings{coenen_visualizing_2019,
	title = {Visualizing and {Measuring} the {Geometry} of {BERT}},
	url = {https://www.semanticscholar.org/paper/Visualizing-and-Measuring-the-Geometry-of-BERT-Coenen-Reif/afd110eace912c2b273e64851c6b4df2658622eb},
	urldate = {2025-08-18},
	author = {Coenen, Andy and Reif, Emily and Yuan, Ann and Kim, Been and Pearce, Adam and Viégas, F. and Wattenberg, M.},
	month = jun,
	year = {2019},
}

@inproceedings{swamy_interpreting_2021,
	title = {Interpreting {Language} {Models} {Through} {Knowledge} {Graph} {Extraction}},
	url = {https://openreview.net/forum?id=PW4AGjla3sx},
	language = {en},
	urldate = {2025-08-18},
	author = {Swamy, Vinitra and Romanou, Angelika and Jaggi, Martin},
	month = oct,
	year = {2021},
}

@inproceedings{perez-mayos_assessing_2021,
	address = {Online},
	title = {Assessing the {Syntactic} {Capabilities} of {Transformer}-based {Multilingual} {Language} {Models}},
	url = {https://aclanthology.org/2021.findings-acl.333/},
	doi = {10.18653/v1/2021.findings-acl.333},
	urldate = {2025-08-18},
	booktitle = {Findings of the {Association} for {Computational} {Linguistics}: {ACL}-{IJCNLP} 2021},
	publisher = {Association for Computational Linguistics},
	author = {Pérez-Mayos, Laura and Táboas García, Alba and Mille, Simon and Wanner, Leo},
	editor = {Zong, Chengqing and Xia, Fei and Li, Wenjie and Navigli, Roberto},
	month = aug,
	year = {2021},
	pages = {3799--3812},
}

@inproceedings{sachan_syntax_2021,
	address = {Online},
	title = {Do {Syntax} {Trees} {Help} {Pre}-trained {Transformers} {Extract} {Information}?},
	url = {https://aclanthology.org/2021.eacl-main.228/},
	doi = {10.18653/v1/2021.eacl-main.228},
	urldate = {2025-08-18},
	booktitle = {Proceedings of the 16th {Conference} of the {European} {Chapter} of the {Association} for {Computational} {Linguistics}: {Main} {Volume}},
	publisher = {Association for Computational Linguistics},
	author = {Sachan, Devendra and Zhang, Yuhao and Qi, Peng and Hamilton, William L.},
	editor = {Merlo, Paola and Tiedemann, Jorg and Tsarfaty, Reut},
	month = apr,
	year = {2021},
	pages = {2647--2661},
}

@inproceedings{dufter-schutze-2020-identifying,
    title = "Identifying Elements Essential for {BERT}{'}s Multilinguality",
    author = {Dufter, Philipp  and
      Sch{\"u}tze, Hinrich},
    editor = "Webber, Bonnie  and
      Cohn, Trevor  and
      He, Yulan  and
      Liu, Yang",
    booktitle = "Proceedings of the 2020 Conference on Empirical Methods in Natural Language Processing (EMNLP)",
    month = nov,
    year = "2020",
    address = "Online",
    publisher = "Association for Computational Linguistics",
    url = "https://aclanthology.org/2020.emnlp-main.358/",
    doi = "10.18653/v1/2020.emnlp-main.358",
    pages = "4423--4437"
}

@inproceedings{yanaka-etal-2020-neural,
    title = "Do Neural Models Learn Systematicity of Monotonicity Inference in Natural Language?",
    author = "Yanaka, Hitomi  and
      Mineshima, Koji  and
      Bekki, Daisuke  and
      Inui, Kentaro",
    editor = "Jurafsky, Dan  and
      Chai, Joyce  and
      Schluter, Natalie  and
      Tetreault, Joel",
    booktitle = "Proceedings of the 58th Annual Meeting of the Association for Computational Linguistics",
    month = jul,
    year = "2020",
    address = "Online",
    publisher = "Association for Computational Linguistics",
    url = "https://aclanthology.org/2020.acl-main.543/",
    doi = "10.18653/v1/2020.acl-main.543",
    pages = "6105--6117"
}

@inproceedings{yanaka-etal-2019-neural,
    title = "Can Neural Networks Understand Monotonicity Reasoning?",
    author = "Yanaka, Hitomi  and
      Mineshima, Koji  and
      Bekki, Daisuke  and
      Inui, Kentaro  and
      Sekine, Satoshi  and
      Abzianidze, Lasha  and
      Bos, Johan",
    editor = "Linzen, Tal  and
      Chrupa{\l}a, Grzegorz  and
      Belinkov, Yonatan  and
      Hupkes, Dieuwke",
    booktitle = "Proceedings of the 2019 ACL Workshop BlackboxNLP: Analyzing and Interpreting Neural Networks for NLP",
    month = aug,
    year = "2019",
    address = "Florence, Italy",
    publisher = "Association for Computational Linguistics",
    url = "https://aclanthology.org/W19-4804/",
    doi = "10.18653/v1/W19-4804",
    pages = "31--40"
}

@inproceedings{wijnholds_assessing_2023,
	address = {Dubrovnik, Croatia},
	title = {Assessing {Monotonicity} {Reasoning} in {Dutch} through {Natural} {Language} {Inference}},
	url = {https://aclanthology.org/2023.findings-eacl.110/},
	doi = {10.18653/v1/2023.findings-eacl.110},
	urldate = {2025-08-15},
	booktitle = {Findings of the {Association} for {Computational} {Linguistics}: {EACL} 2023},
	publisher = {Association for Computational Linguistics},
	author = {Wijnholds, Gijs},
	editor = {Vlachos, Andreas and Augenstein, Isabelle},
	month = may,
	year = {2023},
	pages = {1494--1500},
}

@inproceedings{alt_probing_2020,
	address = {Online},
	title = {Probing {Linguistic} {Features} of {Sentence}-{Level} {Representations} in {Neural} {Relation} {Extraction}},
	url = {https://aclanthology.org/2020.acl-main.140/},
	doi = {10.18653/v1/2020.acl-main.140},
	urldate = {2025-08-15},
	booktitle = {Proceedings of the 58th {Annual} {Meeting} of the {Association} for {Computational} {Linguistics}},
	publisher = {Association for Computational Linguistics},
	author = {Alt, Christoph and Gabryszak, Aleksandra and Hennig, Leonhard},
	editor = {Jurafsky, Dan and Chai, Joyce and Schluter, Natalie and Tetreault, Joel},
	month = jul,
	year = {2020},
	pages = {1534--1545},
}

@inproceedings{artetxe_cross-lingual_2020,
	address = {Online},
	title = {On the {Cross}-lingual {Transferability} of {Monolingual} {Representations}},
	url = {https://aclanthology.org/2020.acl-main.421/},
	doi = {10.18653/v1/2020.acl-main.421},
	urldate = {2025-08-15},
	booktitle = {Proceedings of the 58th {Annual} {Meeting} of the {Association} for {Computational} {Linguistics}},
	publisher = {Association for Computational Linguistics},
	author = {Artetxe, Mikel and Ruder, Sebastian and Yogatama, Dani},
	editor = {Jurafsky, Dan and Chai, Joyce and Schluter, Natalie and Tetreault, Joel},
	month = jul,
	year = {2020},
	pages = {4623--4637},
}

@inproceedings{he_large_2025,
    title = "Large Language Models as Neurolinguistic Subjects: Discrepancy between Performance and Competence",
    author = "He, Linyang  and
      Nie, Ercong  and
      Schmid, Helmut  and
      Schuetze, Hinrich  and
      Mesgarani, Nima  and
      Brennan, Jonathan",
    editor = "Che, Wanxiang  and
      Nabende, Joyce  and
      Shutova, Ekaterina  and
      Pilehvar, Mohammad Taher",
    booktitle = "Findings of the Association for Computational Linguistics: ACL 2025",
    month = jul,
    year = "2025",
    address = "Vienna, Austria",
    publisher = "Association for Computational Linguistics",
    url = "https://aclanthology.org/2025.findings-acl.986/",
    doi = "10.18653/v1/2025.findings-acl.986",
    pages = "19284--19302",
    ISBN = "979-8-89176-256-5"
}

@inproceedings{diehl_martinez_mitigating_2024,
	address = {Miami, Florida, USA},
	title = {Mitigating {Frequency} {Bias} and {Anisotropy} in {Language} {Model} {Pre}-{Training} with {Syntactic} {Smoothing}},
	url = {https://aclanthology.org/2024.emnlp-main.344/},
	doi = {10.18653/v1/2024.emnlp-main.344},
	urldate = {2025-08-15},
	booktitle = {Proceedings of the 2024 {Conference} on {Empirical} {Methods} in {Natural} {Language} {Processing}},
	publisher = {Association for Computational Linguistics},
	author = {Diehl Martinez, Richard and Goriely, Zébulon and Caines, Andrew and Buttery, Paula and Beinborn, Lisa},
	editor = {Al-Onaizan, Yaser and Bansal, Mohit and Chen, Yun-Nung},
	month = nov,
	year = {2024},
	pages = {5999--6011},
}

@inproceedings{hewitt-etal-2023-backpack,
    title = "Backpack Language Models",
    author = "Hewitt, John  and
      Thickstun, John  and
      Manning, Christopher  and
      Liang, Percy",
    editor = "Rogers, Anna  and
      Boyd-Graber, Jordan  and
      Okazaki, Naoaki",
    booktitle = "Proceedings of the 61st Annual Meeting of the Association for Computational Linguistics (Volume 1: Long Papers)",
    month = jul,
    year = "2023",
    address = "Toronto, Canada",
    publisher = "Association for Computational Linguistics",
    url = "https://aclanthology.org/2023.acl-long.506/",
    doi = "10.18653/v1/2023.acl-long.506",
    pages = "9103--9125",
}

@inproceedings{georges_gabriel_charpentier_not_2023,
	address = {Singapore},
	title = {Not all layers are equally as important: {Every} {Layer} {Counts} {BERT}},
	shorttitle = {Not all layers are equally as important},
	url = {https://aclanthology.org/2023.conll-babylm.20/},
	doi = {10.18653/v1/2023.conll-babylm.20},
	urldate = {2025-08-15},
	booktitle = {Proceedings of the {BabyLM} {Challenge} at the 27th {Conference} on {Computational} {Natural} {Language} {Learning}},
	publisher = {Association for Computational Linguistics},
	author = {Georges Gabriel Charpentier, Lucas and Samuel, David},
	editor = {Warstadt, Alex and Mueller, Aaron and Choshen, Leshem and Wilcox, Ethan and Zhuang, Chengxu and Ciro, Juan and Mosquera, Rafael and Paranjabe, Bhargavi and Williams, Adina and Linzen, Tal and Cotterell, Ryan},
	month = dec,
	year = {2023},
	pages = {238--252},
}

@inproceedings{ferrando_similarity_2024,
	address = {Miami, Florida, USA},
	title = {On the {Similarity} of {Circuits} across {Languages}: a {Case} {Study} on the {Subject}-verb {Agreement} {Task}},
	shorttitle = {On the {Similarity} of {Circuits} across {Languages}},
	url = {https://aclanthology.org/2024.findings-emnlp.591/},
	doi = {10.18653/v1/2024.findings-emnlp.591},
	urldate = {2025-08-15},
	booktitle = {Findings of the {Association} for {Computational} {Linguistics}: {EMNLP} 2024},
	publisher = {Association for Computational Linguistics},
	author = {Ferrando, Javier and Costa-Jussà, Marta R.},
	editor = {Al-Onaizan, Yaser and Bansal, Mohit and Chen, Yun-Nung},
	month = nov,
	year = {2024},
	pages = {10115--10125},
}

@inproceedings{arora_causalgym_2024,
	address = {Bangkok, Thailand},
	title = {{CausalGym}: {Benchmarking} causal interpretability methods on linguistic tasks},
	shorttitle = {{CausalGym}},
	url = {https://aclanthology.org/2024.acl-long.785/},
	doi = {10.18653/v1/2024.acl-long.785},
	urldate = {2025-08-15},
	booktitle = {Proceedings of the 62nd {Annual} {Meeting} of the {Association} for {Computational} {Linguistics} ({Volume} 1: {Long} {Papers})},
	publisher = {Association for Computational Linguistics},
	author = {Arora, Aryaman and Jurafsky, Dan and Potts, Christopher},
	editor = {Ku, Lun-Wei and Martins, Andre and Srikumar, Vivek},
	month = aug,
	year = {2024},
	pages = {14638--14663},
}

@inproceedings{wilcox_targeted_2021,
	address = {Online},
	title = {A {Targeted} {Assessment} of {Incremental} {Processing} in {Neural} {Language} {Models} and {Humans}},
	url = {https://aclanthology.org/2021.acl-long.76/},
	doi = {10.18653/v1/2021.acl-long.76},
	urldate = {2025-08-15},
	booktitle = {Proceedings of the 59th {Annual} {Meeting} of the {Association} for {Computational} {Linguistics} and the 11th {International} {Joint} {Conference} on {Natural} {Language} {Processing} ({Volume} 1: {Long} {Papers})},
	publisher = {Association for Computational Linguistics},
	author = {Wilcox, Ethan and Vani, Pranali and Levy, Roger},
	editor = {Zong, Chengqing and Xia, Fei and Li, Wenjie and Navigli, Roberto},
	month = aug,
	year = {2021},
	pages = {939--952},
}

@inproceedings{eisape_probing_2022,
	address = {Abu Dhabi, United Arab Emirates},
	title = {Probing for {Incremental} {Parse} {States} in {Autoregressive} {Language} {Models}},
	url = {https://aclanthology.org/2022.findings-emnlp.203/},
	doi = {10.18653/v1/2022.findings-emnlp.203},
	urldate = {2025-08-15},
	booktitle = {Findings of the {Association} for {Computational} {Linguistics}: {EMNLP} 2022},
	publisher = {Association for Computational Linguistics},
	author = {Eisape, Tiwalayo and Gangireddy, Vineet and Levy, Roger and Kim, Yoon},
	editor = {Goldberg, Yoav and Kozareva, Zornitsa and Zhang, Yue},
	month = dec,
	year = {2022},
	pages = {2801--2813},
}

@article{patil_filtered_2024,
	title = {Filtered {Corpus} {Training} ({FiCT}) {Shows} that {Language} {Models} {Can} {Generalize} from {Indirect} {Evidence}},
	volume = {12},
	url = {https://aclanthology.org/2024.tacl-1.87/},
	doi = {10.1162/tacl_a_00720},
	urldate = {2025-08-15},
	journal = {Transactions of the Association for Computational Linguistics},
	author = {Patil, Abhinav and Jumelet, Jaap and Chiu, Yu Ying and Lapastora, Andy and Shen, Peter and Wang, Lexie and Willrich, Clevis and Steinert-Threlkeld, Shane},
	year = {2024},
	pages = {1597--1615},
}

@inproceedings{jumelet_feature_2023,
	address = {Toronto, Canada},
	title = {Feature {Interactions} {Reveal} {Linguistic} {Structure} in {Language} {Models}},
	url = {https://aclanthology.org/2023.findings-acl.554/},
	doi = {10.18653/v1/2023.findings-acl.554},
	urldate = {2025-08-14},
	booktitle = {Findings of the {Association} for {Computational} {Linguistics}: {ACL} 2023},
	publisher = {Association for Computational Linguistics},
	author = {Jumelet, Jaap and Zuidema, Willem},
	editor = {Rogers, Anna and Boyd-Graber, Jordan and Okazaki, Naoaki},
	month = jul,
	year = {2023},
	pages = {8697--8712},
}

@article{ahuja_learning_2025,
	title = {Learning {Syntax} {Without} {Planting} {Trees}: {Understanding} {Hierarchical} {Generalization} in {Transformers}},
	volume = {13},
	shorttitle = {Learning {Syntax} {Without} {Planting} {Trees}},
	url = {https://aclanthology.org/2025.tacl-1.6/},
	doi = {10.1162/tacl_a_00733},
	urldate = {2025-08-14},
	journal = {Transactions of the Association for Computational Linguistics},
	author = {Ahuja, Kabir and Balachandran, Vidhisha and Panwar, Madhur and He, Tianxing and Smith, Noah A. and Goyal, Navin and Tsvetkov, Yulia},
	year = {2025},
	pages = {121--141},
}

@inproceedings{kim_testing_2021,
	address = {Online},
	title = {Testing for {Grammatical} {Category} {Abstraction} in {Neural} {Language} {Models}},
	url = {https://aclanthology.org/2021.scil-1.59/},
	urldate = {2025-08-14},
	booktitle = {Proceedings of the {Society} for {Computation} in {Linguistics} 2021},
	publisher = {Association for Computational Linguistics},
	author = {Kim, Najoung and Smolensky, Paul},
	editor = {Ettinger, Allyson and Pavlick, Ellie and Prickett, Brandon},
	month = feb,
	year = {2021},
	pages = {467--470},
}

@inproceedings{yun_transformer_2021,
	address = {Online},
	title = {Transformer visualization via dictionary learning: contextualized embedding as a linear superposition of transformer factors},
	shorttitle = {Transformer visualization via dictionary learning},
	url = {https://aclanthology.org/2021.deelio-1.1/},
	doi = {10.18653/v1/2021.deelio-1.1},
	urldate = {2025-08-14},
	booktitle = {Proceedings of {Deep} {Learning} {Inside} {Out} ({DeeLIO}): {The} 2nd {Workshop} on {Knowledge} {Extraction} and {Integration} for {Deep} {Learning} {Architectures}},
	publisher = {Association for Computational Linguistics},
	author = {Yun, Zeyu and Chen, Yubei and Olshausen, Bruno and LeCun, Yann},
	editor = {Agirre, Eneko and Apidianaki, Marianna and Vulić, Ivan},
	month = jun,
	year = {2021},
	pages = {1--10},
}

@article{turner_steering_2024,
	title = {Steering {Language} {Models} with {Activation} {Engineering}},
	url = {https://openreview.net/forum?id=2XBPdPIcFK},
	language = {en},
	urldate = {2025-08-14},
	author = {Turner, Alexander Matt and Thiergart, Lisa and Leech, Gavin and Udell, David and Vazquez, Juan J. and Mini, Ulisse and MacDiarmid, Monte},
	month = oct,
	year = {2024},
}

@article{gurnee_finding_2023,
	title = {Finding {Neurons} in a {Haystack}: {Case} {Studies} with {Sparse} {Probing}},
	issn = {2835-8856},
	shorttitle = {Finding {Neurons} in a {Haystack}},
	url = {https://openreview.net/forum?id=JYs1R9IMJr},
	language = {en},
	urldate = {2025-08-14},
	journal = {Transactions on Machine Learning Research},
	author = {Gurnee, Wes and Nanda, Neel and Pauly, Matthew and Harvey, Katherine and Troitskii, Dmitrii and Bertsimas, Dimitris},
	month = jun,
	year = {2023},
}

@inproceedings{yin_interpreting_2022,
	address = {Abu Dhabi, United Arab Emirates},
	title = {Interpreting {Language} {Models} with {Contrastive} {Explanations}},
	url = {https://aclanthology.org/2022.emnlp-main.14/},
	doi = {10.18653/v1/2022.emnlp-main.14},
	urldate = {2025-08-14},
	booktitle = {Proceedings of the 2022 {Conference} on {Empirical} {Methods} in {Natural} {Language} {Processing}},
	publisher = {Association for Computational Linguistics},
	author = {Yin, Kayo and Neubig, Graham},
	editor = {Goldberg, Yoav and Kozareva, Zornitsa and Zhang, Yue},
	month = dec,
	year = {2022},
	pages = {184--198},
}

@inproceedings{blevins_prompting_2023,
	address = {Toronto, Canada},
	title = {Prompting {Language} {Models} for {Linguistic} {Structure}},
	url = {https://aclanthology.org/2023.acl-long.367/},
	doi = {10.18653/v1/2023.acl-long.367},
	urldate = {2025-08-13},
	booktitle = {Proceedings of the 61st {Annual} {Meeting} of the {Association} for {Computational} {Linguistics} ({Volume} 1: {Long} {Papers})},
	publisher = {Association for Computational Linguistics},
	author = {Blevins, Terra and Gonen, Hila and Zettlemoyer, Luke},
	editor = {Rogers, Anna and Boyd-Graber, Jordan and Okazaki, Naoaki},
	month = jul,
	year = {2023},
	pages = {6649--6663},
}

@inproceedings{elgaar_ling-cl_2023,
	address = {Singapore},
	title = {Ling-{CL}: {Understanding} {NLP} {Models} through {Linguistic} {Curricula}},
	shorttitle = {Ling-{CL}},
	url = {https://aclanthology.org/2023.emnlp-main.834/},
	doi = {10.18653/v1/2023.emnlp-main.834},
	urldate = {2025-08-13},
	booktitle = {Proceedings of the 2023 {Conference} on {Empirical} {Methods} in {Natural} {Language} {Processing}},
	publisher = {Association for Computational Linguistics},
	author = {Elgaar, Mohamed and Amiri, Hadi},
	editor = {Bouamor, Houda and Pino, Juan and Bali, Kalika},
	month = dec,
	year = {2023},
	pages = {13526--13542},
}

@article{ettinger_what_2020,
	title = {What {BERT} {Is} {Not}: {Lessons} from a {New} {Suite} of {Psycholinguistic} {Diagnostics} for {Language} {Models}},
	volume = {8},
	shorttitle = {What {BERT} {Is} {Not}},
	url = {https://aclanthology.org/2020.tacl-1.3/},
	doi = {10.1162/tacl_a_00298},
	urldate = {2025-08-13},
	journal = {Transactions of the Association for Computational Linguistics},
	author = {Ettinger, Allyson},
	editor = {Johnson, Mark and Roark, Brian and Nenkova, Ani},
	year = {2020},
	pages = {34--48},
}

@inproceedings{bunzeck_subword_2025,
	address = {Vienna, Austria},
	title = {Subword models struggle with word learning, but surprisal hides it},
	isbn = {979-8-89176-252-7},
	url = {https://aclanthology.org/2025.acl-short.24/},
	doi = {10.18653/v1/2025.acl-short.24},
	urldate = {2025-08-13},
	booktitle = {Proceedings of the 63rd {Annual} {Meeting} of the {Association} for {Computational} {Linguistics} ({Volume} 2: {Short} {Papers})},
	publisher = {Association for Computational Linguistics},
	author = {Bunzeck, Bastian and Zarrieß, Sina},
	editor = {Che, Wanxiang and Nabende, Joyce and Shutova, Ekaterina and Pilehvar, Mohammad Taher},
	month = jul,
	year = {2025},
	pages = {286--300},
}

@inproceedings{ravishankar_effects_2022,
	address = {Abu Dhabi, United Arab Emirates},
	title = {The {Effects} of {Corpus} {Choice} and {Morphosyntax} on {Multilingual} {Space} {Induction}},
	url = {https://aclanthology.org/2022.findings-emnlp.304/},
	doi = {10.18653/v1/2022.findings-emnlp.304},
	urldate = {2025-08-13},
	booktitle = {Findings of the {Association} for {Computational} {Linguistics}: {EMNLP} 2022},
	publisher = {Association for Computational Linguistics},
	author = {Ravishankar, Vinit and Nivre, Joakim},
	editor = {Goldberg, Yoav and Kozareva, Zornitsa and Zhang, Yue},
	month = dec,
	year = {2022},
	pages = {4130--4139},
}

@inproceedings{papadimitriou_injecting_2023,
	address = {Singapore},
	title = {Injecting structural hints: {Using} language models to study inductive biases in language learning},
	shorttitle = {Injecting structural hints},
	url = {https://aclanthology.org/2023.findings-emnlp.563/},
	doi = {10.18653/v1/2023.findings-emnlp.563},
	urldate = {2025-08-13},
	booktitle = {Findings of the {Association} for {Computational} {Linguistics}: {EMNLP} 2023},
	publisher = {Association for Computational Linguistics},
	author = {Papadimitriou, Isabel and Jurafsky, Dan},
	editor = {Bouamor, Houda and Pino, Juan and Bali, Kalika},
	month = dec,
	year = {2023},
	pages = {8402--8413},
}

@inproceedings{misra_language_2024,
	address = {Miami, Florida, USA},
	title = {Language {Models} {Learn} {Rare} {Phenomena} from {Less} {Rare} {Phenomena}: {The} {Case} of the {Missing} {AANNs}},
	shorttitle = {Language {Models} {Learn} {Rare} {Phenomena} from {Less} {Rare} {Phenomena}},
	url = {https://aclanthology.org/2024.emnlp-main.53/},
	doi = {10.18653/v1/2024.emnlp-main.53},
	urldate = {2025-08-13},
	booktitle = {Proceedings of the 2024 {Conference} on {Empirical} {Methods} in {Natural} {Language} {Processing}},
	publisher = {Association for Computational Linguistics},
	author = {Misra, Kanishka and Mahowald, Kyle},
	editor = {Al-Onaizan, Yaser and Bansal, Mohit and Chen, Yun-Nung},
	month = nov,
	year = {2024},
	pages = {913--929},
}

@inproceedings{someya_jblimp_2023,
	address = {Dubrovnik, Croatia},
	title = {{JBLiMP}: {Japanese} {Benchmark} of {Linguistic} {Minimal} {Pairs}},
	shorttitle = {{JBLiMP}},
	url = {https://aclanthology.org/2023.findings-eacl.117/},
	doi = {10.18653/v1/2023.findings-eacl.117},
	urldate = {2025-08-13},
	booktitle = {Findings of the {Association} for {Computational} {Linguistics}: {EACL} 2023},
	publisher = {Association for Computational Linguistics},
	author = {Someya, Taiga and Oseki, Yohei},
	editor = {Vlachos, Andreas and Augenstein, Isabelle},
	month = may,
	year = {2023},
	pages = {1581--1594},
}

@misc{xu2025languagemodelslearntypologically,
 arxiv = {https://www.arxiv.org/pdf/2502.12317},
 author = {Xu, Tianyang and 
Kuribayashi, Tatsuki and 
Oseki, Yohei and 
Cotterell, Ryan and 
Warstadt, Alex},
 code = {https://github.com/sally-xu-42/Typological_Universals},
 slides = {N/A},
 title = {Can Language Models Learn Typologically Implausible Languages?},
 url = {https://www.arxiv.org/pdf/2502.12317},
 venue = {TACL},
 video = {N/A},
 year = {2025}
}

@article{clark_cross-linguistic_2023,
	title = {A {Cross}-{Linguistic} {Pressure} for {Uniform} {Information} {Density} in {Word} {Order}},
	volume = {11},
	url = {https://aclanthology.org/2023.tacl-1.59/},
	doi = {10.1162/tacl_a_00589},
	urldate = {2025-08-13},
	journal = {Transactions of the Association for Computational Linguistics},
	author = {Clark, Thomas Hikaru and Meister, Clara and Pimentel, Tiago and Hahn, Michael and Cotterell, Ryan and Futrell, Richard and Levy, Roger},
	year = {2023},
	pages = {1048--1065},
}

@inproceedings{kuribayashi_emergent_2024,
	address = {Bangkok, Thailand},
	title = {Emergent {Word} {Order} {Universals} from {Cognitively}-{Motivated} {Language} {Models}},
	url = {https://aclanthology.org/2024.acl-long.781/},
	doi = {10.18653/v1/2024.acl-long.781},
	urldate = {2025-08-13},
	booktitle = {Proceedings of the 62nd {Annual} {Meeting} of the {Association} for {Computational} {Linguistics} ({Volume} 1: {Long} {Papers})},
	publisher = {Association for Computational Linguistics},
	author = {Kuribayashi, Tatsuki and Ueda, Ryo and Yoshida, Ryo and Oseki, Yohei and Briscoe, Ted and Baldwin, Timothy},
	editor = {Ku, Lun-Wei and Martins, Andre and Srikumar, Vivek},
	month = aug,
	year = {2024},
	pages = {14522--14543},
}

@article{constantinescu_investigating_2025,
	title = {Investigating {Critical} {Period} {Effects} in {Language} {Acquisition} through {Neural} {Language} {Models}},
	volume = {13},
	url = {https://aclanthology.org/2025.tacl-1.5/},
	doi = {10.1162/tacl_a_00725},
	urldate = {2025-08-13},
	journal = {Transactions of the Association for Computational Linguistics},
	author = {Constantinescu, Ionut and Pimentel, Tiago and Cotterell, Ryan and Warstadt, Alex},
	year = {2025},
	pages = {96--120},
}

@inproceedings{kallini_mission_2024,
	address = {Bangkok, Thailand},
	title = {Mission: {Impossible} {Language} {Models}},
	shorttitle = {Mission},
	url = {https://aclanthology.org/2024.acl-long.787/},
	doi = {10.18653/v1/2024.acl-long.787},
	urldate = {2025-08-13},
	booktitle = {Proceedings of the 62nd {Annual} {Meeting} of the {Association} for {Computational} {Linguistics} ({Volume} 1: {Long} {Papers})},
	publisher = {Association for Computational Linguistics},
	author = {Kallini, Julie and Papadimitriou, Isabel and Futrell, Richard and Mahowald, Kyle and Potts, Christopher},
	editor = {Ku, Lun-Wei and Martins, Andre and Srikumar, Vivek},
	month = aug,
	year = {2024},
	pages = {14691--14714},
}

@inproceedings{hu_between_2025,
	address = {Vienna, Austria},
	title = {Between {Circuits} and {Chomsky}: {Pre}-pretraining on {Formal} {Languages} {Imparts} {Linguistic} {Biases}},
	isbn = {979-8-89176-251-0},
	shorttitle = {Between {Circuits} and {Chomsky}},
	url = {https://aclanthology.org/2025.acl-long.478/},
	doi = {10.18653/v1/2025.acl-long.478},
	urldate = {2025-08-13},
	booktitle = {Proceedings of the 63rd {Annual} {Meeting} of the {Association} for {Computational} {Linguistics} ({Volume} 1: {Long} {Papers})},
	publisher = {Association for Computational Linguistics},
	author = {Hu, Michael Y. and Petty, Jackson and Shi, Chuan and Merrill, William and Linzen, Tal},
	editor = {Che, Wanxiang and Nabende, Joyce and Shutova, Ekaterina and Pilehvar, Mohammad Taher},
	month = jul,
	year = {2025},
	pages = {9691--9709},
}

@misc{ginn_tree_2024,
	title = {Tree {Transformers} are an {Ineffective} {Model} of {Syntactic} {Constituency}},
	url = {http://arxiv.org/abs/2411.16993},
	doi = {10.48550/arXiv.2411.16993},
	urldate = {2025-08-13},
	publisher = {arXiv},
	author = {Ginn, Michael},
	month = nov,
	year = {2024},
	note = {arXiv:2411.16993},
}

@misc{zimmerman_tokens_2025,
	title = {Tokens, the oft-overlooked appetizer: {Large} language models, the distributional hypothesis, and meaning},
	shorttitle = {Tokens, the oft-overlooked appetizer},
	url = {http://arxiv.org/abs/2412.10924},
	doi = {10.48550/arXiv.2412.10924},
	urldate = {2025-08-13},
	publisher = {arXiv},
	author = {Zimmerman, Julia Witte and Hudon, Denis and Cramer, Kathryn and Ruiz, Alejandro J. and Beauregard, Calla and Fehr, Ashley and Fudolig, Mikaela Irene and Demarest, Bradford and Bird, Yoshi Meke and Trujillo, Milo Z. and Danforth, Christopher M. and Dodds, Peter Sheridan},
	month = apr,
	year = {2025},
	note = {arXiv:2412.10924},
}

@inproceedings{swarup-etal-2025-syntax,
    title = "From Syntax to Semantics: Evaluating the Impact of Linguistic Structures on {LLM}-Based Information Extraction",
    author = "Swarup, Anushka  and
      Bhandarkar, Avanti  and
      Wilson, Ronald  and
      Pan, Tianyu  and
      Woodard, Damon",
    editor = "Fei, Hao  and
      Tu, Kewei  and
      Zhang, Yuhui  and
      Hu, Xiang  and
      Han, Wenjuan  and
      Jia, Zixia  and
      Zheng, Zilong  and
      Cao, Yixin  and
      Zhang, Meishan  and
      Lu, Wei  and
      Siddharth, N.  and
      {\O}vrelid, Lilja  and
      Xue, Nianwen  and
      Zhang, Yue",
    booktitle = "Proceedings of the 1st Joint Workshop on Large Language Models and Structure Modeling (XLLM 2025)",
    month = aug,
    year = "2025",
    address = "Vienna, Austria",
    publisher = "Association for Computational Linguistics",
    url = "https://aclanthology.org/2025.xllm-1.5/",
    doi = "10.18653/v1/2025.xllm-1.5",
    pages = "36--48",
    ISBN = "979-8-89176-286-2",
}

@article{dentella_language_2025,
	title = {Language in vivo vs. in silico: {Size} matters but {Larger} {Language} {Models} still do not comprehend language on a par with humans due to impenetrable semantic reference},
	volume = {20},
	issn = {1932-6203},
	shorttitle = {Language in vivo vs. in silico},
	url = {https://journals.plos.org/plosone/article?id=10.1371/journal.pone.0327794},
	doi = {10.1371/journal.pone.0327794},
	language = {en},
	number = {7},
	urldate = {2025-08-12},
	journal = {PLOS ONE},
	author = {Dentella, Vittoria and Günther, Fritz and Leivada, Evelina},
	month = jul,
	year = {2025},
	pages = {e0327794},
}

@inproceedings{yang_anything_2025,
	address = {Vienna, Austria},
	title = {Anything {Goes}? {A} {Crosslinguistic} {Study} of ({Im})possible {Language} {Learning} in {LMs}},
	isbn = {979-8-89176-251-0},
	shorttitle = {Anything {Goes}?},
	url = {https://aclanthology.org/2025.acl-long.1264/},
	doi = {10.18653/v1/2025.acl-long.1264},
	urldate = {2025-08-12},
	booktitle = {Proceedings of the 63rd {Annual} {Meeting} of the {Association} for {Computational} {Linguistics} ({Volume} 1: {Long} {Papers})},
	publisher = {Association for Computational Linguistics},
	author = {Yang, Xiulin and Aoyama, Tatsuya and Yao, Yuekun and Wilcox, Ethan},
	editor = {Che, Wanxiang and Nabende, Joyce and Shutova, Ekaterina and Pilehvar, Mohammad Taher},
	month = jul,
	year = {2025},
	pages = {26058--26077},
}

@article{WILCOX2025104650,
title = {Bigger is not always better: The importance of human-scale language modeling for psycholinguistics},
journal = {Journal of Memory and Language},
volume = {144},
pages = {104650},
year = {2025},
issn = {0749-596X},
doi = {https://doi.org/10.1016/j.jml.2025.104650},
url = {https://www.sciencedirect.com/science/article/pii/S0749596X25000439},
author = {Ethan Gotlieb Wilcox and Michael Y. Hu and Aaron Mueller and Alex Warstadt and Leshem Choshen and Chengxu Zhuang and Adina Williams and Ryan Cotterell and Tal Linzen},
}

@inproceedings{xefteri-etal-2025-syntactic,
    title = "Syntactic Control of Language Models by Posterior Inference",
    author = "Xefteri, Vicky  and
      Vieira, Tim  and
      Cotterell, Ryan  and
      Amini, Afra",
    editor = "Che, Wanxiang  and
      Nabende, Joyce  and
      Shutova, Ekaterina  and
      Pilehvar, Mohammad Taher",
    booktitle = "Findings of the Association for Computational Linguistics: ACL 2025",
    month = jul,
    year = "2025",
    address = "Vienna, Austria",
    publisher = "Association for Computational Linguistics",
    url = "https://aclanthology.org/2025.findings-acl.1300/",
    doi = "10.18653/v1/2025.findings-acl.1300",
    pages = "25350--25365",
    ISBN = "979-8-89176-256-5"
}

@misc{li_incremental_2024,
	title = {Incremental {Comprehension} of {Garden}-{Path} {Sentences} by {Large} {Language} {Models}: {Semantic} {Interpretation}, {Syntactic} {Re}-{Analysis}, and {Attention}},
	shorttitle = {Incremental {Comprehension} of {Garden}-{Path} {Sentences} by {Large} {Language} {Models}},
	url = {http://arxiv.org/abs/2405.16042},
	doi = {10.48550/arXiv.2405.16042},
	urldate = {2025-08-12},
	publisher = {arXiv},
	author = {Li, Andrew and Feng, Xianle and Narang, Siddhant and Peng, Austin and Cai, Tianle and Shah, Raj Sanjay and Varma, Sashank},
	month = may,
	year = {2024},
	note = {arXiv:2405.16042},
}

@article{huang_large-scale_2024,
	title = {Large-scale benchmark yields no evidence that language model surprisal explains syntactic disambiguation difficulty},
	volume = {137},
	issn = {0749-596X},
	url = {https://www.sciencedirect.com/science/article/pii/S0749596X24000135},
	doi = {10.1016/j.jml.2024.104510},
	urldate = {2025-08-12},
	journal = {Journal of Memory and Language},
	author = {Huang, Kuan-Jung and Arehalli, Suhas and Kugemoto, Mari and Muxica, Christian and Prasad, Grusha and Dillon, Brian and Linzen, Tal},
	month = aug,
	year = {2024},
	pages = {104510},
}

@inproceedings{sinha_language_2023,
	address = {Toronto, Canada},
	title = {Language model acceptability judgements are not always robust to context},
	url = {https://aclanthology.org/2023.acl-long.333/},
	doi = {10.18653/v1/2023.acl-long.333},
	urldate = {2025-08-12},
	booktitle = {Proceedings of the 61st {Annual} {Meeting} of the {Association} for {Computational} {Linguistics} ({Volume} 1: {Long} {Papers})},
	publisher = {Association for Computational Linguistics},
	author = {Sinha, Koustuv and Gauthier, Jon and Mueller, Aaron and Misra, Kanishka and Fuentes, Keren and Levy, Roger and Williams, Adina},
	editor = {Rogers, Anna and Boyd-Graber, Jordan and Okazaki, Naoaki},
	month = jul,
	year = {2023},
	pages = {6043--6063},
}

@inproceedings{michaelov_structural_2023,
	address = {Singapore},
	title = {Structural {Priming} {Demonstrates} {Abstract} {Grammatical} {Representations} in {Multilingual} {Language} {Models}},
	url = {https://aclanthology.org/2023.emnlp-main.227/},
	doi = {10.18653/v1/2023.emnlp-main.227},
	urldate = {2025-08-12},
	booktitle = {Proceedings of the 2023 {Conference} on {Empirical} {Methods} in {Natural} {Language} {Processing}},
	publisher = {Association for Computational Linguistics},
	author = {Michaelov, James A. and Arnett, Catherine and Chang, Tyler A. and Bergen, Benjamin K.},
	editor = {Bouamor, Houda and Pino, Juan and Bali, Kalika},
	month = dec,
	year = {2023},
	pages = {3703--3720},
}

@article{sinclair_structural_2022,
	title = {Structural {Persistence} in {Language} {Models}: {Priming} as a {Window} into {Abstract} {Language} {Representations}},
	volume = {10},
	shorttitle = {Structural {Persistence} in {Language} {Models}},
	url = {https://aclanthology.org/2022.tacl-1.60/},
	doi = {10.1162/tacl_a_00504},
	urldate = {2025-08-12},
	journal = {Transactions of the Association for Computational Linguistics},
	author = {Sinclair, Arabella and Jumelet, Jaap and Zuidema, Willem and Fernández, Raquel},
	editor = {Roark, Brian and Nenkova, Ani},
	year = {2022},
	pages = {1031--1050},
}

@inproceedings{weber_interpretability_2024,
	address = {Bangkok, Thailand},
	title = {Interpretability of {Language} {Models} via {Task} {Spaces}},
	url = {https://aclanthology.org/2024.acl-long.248/},
	doi = {10.18653/v1/2024.acl-long.248},
	urldate = {2025-08-12},
	booktitle = {Proceedings of the 62nd {Annual} {Meeting} of the {Association} for {Computational} {Linguistics} ({Volume} 1: {Long} {Papers})},
	publisher = {Association for Computational Linguistics},
	author = {Weber, Lucas and Jumelet, Jaap and Bruni, Elia and Hupkes, Dieuwke},
	editor = {Ku, Lun-Wei and Martins, Andre and Srikumar, Vivek},
	month = aug,
	year = {2024},
	pages = {4522--4538},
}

@inproceedings{alkhamissi-etal-2025-language,
    title = "From Language to Cognition: How {LLM}s Outgrow the Human Language Network",
    author = "AlKhamissi, Badr  and
      Tuckute, Greta  and
      Tang, Yingtian  and
      Binhuraib, Taha Osama A  and
      Bosselut, Antoine  and
      Schrimpf, Martin",
    editor = "Christodoulopoulos, Christos  and
      Chakraborty, Tanmoy  and
      Rose, Carolyn  and
      Peng, Violet",
    booktitle = "Proceedings of the 2025 Conference on Empirical Methods in Natural Language Processing",
    month = nov,
    year = "2025",
    address = "Suzhou, China",
    publisher = "Association for Computational Linguistics",
    url = "https://aclanthology.org/2025.emnlp-main.1237/",
    doi = "10.18653/v1/2025.emnlp-main.1237",
    pages = "24332--24350",
    ISBN = "979-8-89176-332-6",
}

@inproceedings{wang_extracting_2025,
	address = {Abu Dhabi, UAE},
	title = {Extracting structure from an {LLM} - how to improve on surprisal-based models of {Human} {Language} {Processing}},
	url = {https://aclanthology.org/2025.coling-main.329/},
	urldate = {2025-08-12},
	booktitle = {Proceedings of the 31st {International} {Conference} on {Computational} {Linguistics}},
	publisher = {Association for Computational Linguistics},
	author = {Wang, Daphne P. and Sadrzadeh, Mehrnoosh and Stanojević, Miloš and Chow, Wing-Yee and Breheny, Richard},
	editor = {Rambow, Owen and Wanner, Leo and Apidianaki, Marianna and Al-Khalifa, Hend and Eugenio, Barbara Di and Schockaert, Steven},
	month = jan,
	year = {2025},
	pages = {4938--4944},
}

@article{kuncoro-etal-2020-syntactic,
    title = "Syntactic Structure Distillation Pretraining for Bidirectional Encoders",
    author = "Kuncoro, Adhiguna  and
      Kong, Lingpeng  and
      Fried, Daniel  and
      Yogatama, Dani  and
      Rimell, Laura  and
      Dyer, Chris  and
      Blunsom, Phil",
    editor = "Johnson, Mark  and
      Roark, Brian  and
      Nenkova, Ani",
    journal = "Transactions of the Association for Computational Linguistics",
    volume = "8",
    year = "2020",
    address = "Cambridge, MA",
    publisher = "MIT Press",
    url = "https://aclanthology.org/2020.tacl-1.50/",
    doi = "10.1162/tacl_a_00345",
    pages = "776--794"
}

@article{sartran_transformer_2022,
	title = {Transformer {Grammars}: {Augmenting} {Transformer} {Language} {Models} with {Syntactic} {Inductive} {Biases} at {Scale}},
	volume = {10},
	shorttitle = {Transformer {Grammars}},
	url = {https://aclanthology.org/2022.tacl-1.81/},
	doi = {10.1162/tacl_a_00526},
	urldate = {2025-08-11},
	journal = {Transactions of the Association for Computational Linguistics},
	author = {Sartran, Laurent and Barrett, Samuel and Kuncoro, Adhiguna and Stanojević, Miloš and Blunsom, Phil and Dyer, Chris},
	editor = {Roark, Brian and Nenkova, Ani},
	year = {2022},
	pages = {1423--1439},
}

@inproceedings{kuribayashi_lower_2021,
	address = {Online},
	title = {Lower {Perplexity} is {Not} {Always} {Human}-{Like}},
	url = {https://aclanthology.org/2021.acl-long.405/},
	doi = {10.18653/v1/2021.acl-long.405},
	urldate = {2025-08-11},
	booktitle = {Proceedings of the 59th {Annual} {Meeting} of the {Association} for {Computational} {Linguistics} and the 11th {International} {Joint} {Conference} on {Natural} {Language} {Processing} ({Volume} 1: {Long} {Papers})},
	publisher = {Association for Computational Linguistics},
	author = {Kuribayashi, Tatsuki and Oseki, Yohei and Ito, Takumi and Yoshida, Ryo and Asahara, Masayuki and Inui, Kentaro},
	editor = {Zong, Chengqing and Xia, Fei and Li, Wenjie and Navigli, Roberto},
	month = aug,
	year = {2021},
	pages = {5203--5217},
}

@inproceedings{salvatore_logical-based_2019,
	address = {Hong Kong, China},
	title = {A logical-based corpus for cross-lingual evaluation},
	url = {https://aclanthology.org/D19-6103/},
	doi = {10.18653/v1/D19-6103},
	urldate = {2025-08-11},
	booktitle = {Proceedings of the 2nd {Workshop} on {Deep} {Learning} {Approaches} for {Low}-{Resource} {NLP} ({DeepLo} 2019)},
	publisher = {Association for Computational Linguistics},
	author = {Salvatore, Felipe and Finger, Marcelo and Hirata Jr, Roberto},
	editor = {Cherry, Colin and Durrett, Greg and Foster, George and Haffari, Reza and Khadivi, Shahram and Peng, Nanyun and Ren, Xiang and Swayamdipta, Swabha},
	month = nov,
	year = {2019},
	pages = {22--30},
}

@inproceedings{clouatre_local_2022,
	address = {Online only},
	title = {Local {Structure} {Matters} {Most} in {Most} {Languages}},
	url = {https://aclanthology.org/2022.aacl-short.35/},
	doi = {10.18653/v1/2022.aacl-short.35},
	urldate = {2025-08-11},
	booktitle = {Proceedings of the 2nd {Conference} of the {Asia}-{Pacific} {Chapter} of the {Association} for {Computational} {Linguistics} and the 12th {International} {Joint} {Conference} on {Natural} {Language} {Processing} ({Volume} 2: {Short} {Papers})},
	publisher = {Association for Computational Linguistics},
	author = {Clouatre, Louis and Parthasarathi, Prasanna and Zouaq, Amal and Chandar, Sarath},
	editor = {He, Yulan and Ji, Heng and Li, Sujian and Liu, Yang and Chang, Chua-Hui},
	month = nov,
	year = {2022},
	pages = {285--294},
}

@inproceedings{kim_does_2024,
	title = {Does {Incomplete} {Syntax} {Influence} {Korean} {Language} {Model}? {Focusing} on {Word} {Order} and {Case} {Markers}},
	shorttitle = {Does {Incomplete} {Syntax} {Influence} {Korean} {Language} {Model}?},
	url = {https://openreview.net/forum?id=yfyHxvVzZT},
	language = {en},
	urldate = {2025-08-11},
	author = {Kim, Jong Myoung and Lee, Young-Jun and Han, Yong-Jin and Choi, Ho-Jin and Jung, Sangkeun},
	month = aug,
	year = {2024},
}

@inproceedings{kervadec_unnatural_2023,
	address = {Singapore},
	title = {Unnatural language processing: {How} do language models handle machine-generated prompts?},
	shorttitle = {Unnatural language processing},
	url = {https://aclanthology.org/2023.findings-emnlp.959/},
	doi = {10.18653/v1/2023.findings-emnlp.959},
	urldate = {2025-08-11},
	booktitle = {Findings of the {Association} for {Computational} {Linguistics}: {EMNLP} 2023},
	publisher = {Association for Computational Linguistics},
	author = {Kervadec, Corentin and Franzon, Francesca and Baroni, Marco},
	editor = {Bouamor, Houda and Pino, Juan and Bali, Kalika},
	month = dec,
	year = {2023},
	pages = {14377--14392},
}

@misc{chen_when_2024,
	title = {When does word order matter and when doesn't it?},
	url = {http://arxiv.org/abs/2402.18838},
	doi = {10.48550/arXiv.2402.18838},
	urldate = {2025-08-11},
	publisher = {arXiv},
	author = {Chen, Xuanda and O'Donnell, Timothy and Reddy, Siva},
	month = mar,
	year = {2024},
	note = {arXiv:2402.18838},
}

@inproceedings{cheng_linguistic_2025,
	address = {Albuquerque, New Mexico, USA},
	title = {Linguistic {Blind} {Spots} of {Large} {Language} {Models}},
	isbn = {979-8-89176-227-5},
	url = {https://aclanthology.org/2025.cmcl-1.3/},
	doi = {10.18653/v1/2025.cmcl-1.3},
	urldate = {2025-08-11},
	booktitle = {Proceedings of the {Workshop} on {Cognitive} {Modeling} and {Computational} {Linguistics}},
	publisher = {Association for Computational Linguistics},
	author = {Cheng, Jiali and Amiri, Hadi},
	editor = {Kuribayashi, Tatsuki and Rambelli, Giulia and Takmaz, Ece and Wicke, Philipp and Li, Jixing and Oh, Byung-Doh},
	month = may,
	year = {2025},
	pages = {1--17},
}

@inproceedings{hou_detecting_2023,
	address = {Singapore},
	title = {Detecting {Syntactic} {Change} with {Pre}-trained {Transformer} {Models}},
	url = {https://aclanthology.org/2023.findings-emnlp.230/},
	doi = {10.18653/v1/2023.findings-emnlp.230},
	urldate = {2025-08-11},
	booktitle = {Findings of the {Association} for {Computational} {Linguistics}: {EMNLP} 2023},
	publisher = {Association for Computational Linguistics},
	author = {Hou, Liwen and Smith, David},
	editor = {Bouamor, Houda and Pino, Juan and Bali, Kalika},
	month = dec,
	year = {2023},
	pages = {3564--3574},
}

@inproceedings{taktasheva_shaking_2021,
	address = {Punta Cana, Dominican Republic},
	title = {Shaking {Syntactic} {Trees} on the {Sesame} {Street}: {Multilingual} {Probing} with {Controllable} {Perturbations}},
	shorttitle = {Shaking {Syntactic} {Trees} on the {Sesame} {Street}},
	url = {https://aclanthology.org/2021.mrl-1.17/},
	doi = {10.18653/v1/2021.mrl-1.17},
	urldate = {2025-08-11},
	booktitle = {Proceedings of the 1st {Workshop} on {Multilingual} {Representation} {Learning}},
	publisher = {Association for Computational Linguistics},
	author = {Taktasheva, Ekaterina and Mikhailov, Vladislav and Artemova, Ekaterina},
	editor = {Ataman, Duygu and Birch, Alexandra and Conneau, Alexis and Firat, Orhan and Ruder, Sebastian and Sahin, Gozde Gul},
	month = nov,
	year = {2021},
	pages = {191--210},
}

@inproceedings{duan_unnatural_2025,
	title = {Unnatural {Languages} {Are} {Not} {Bugs} but {Features} for {LLMs}},
	url = {https://openreview.net/forum?id=LxRvQwO95c&referrer=%5Bthe%20profile%20of%20Qian%20Liu%5D(%2Fprofile%3Fid%3D~Qian_Liu2)},
	language = {en},
	urldate = {2025-08-11},
	author = {Duan, Keyu and Zhao, Yiran and Feng, Zhili and Ni, Jinjie and Pang, Tianyu and Liu, Qian and Cai, Tianle and Dou, Longxu and Kawaguchi, Kenji and Goyal, Anirudh and Kolter, J. Zico and Shieh, Michael Qizhe},
	month = mar,
	year = {2025},
}

@inproceedings{yanaka_assessing_2021,
	address = {Punta Cana, Dominican Republic},
	title = {Assessing the {Generalization} {Capacity} of {Pre}-trained {Language} {Models} through {Japanese} {Adversarial} {Natural} {Language} {Inference}},
	url = {https://aclanthology.org/2021.blackboxnlp-1.26/},
	doi = {10.18653/v1/2021.blackboxnlp-1.26},
	urldate = {2025-08-11},
	booktitle = {Proceedings of the {Fourth} {BlackboxNLP} {Workshop} on {Analyzing} and {Interpreting} {Neural} {Networks} for {NLP}},
	publisher = {Association for Computational Linguistics},
	author = {Yanaka, Hitomi and Mineshima, Koji},
	editor = {Bastings, Jasmijn and Belinkov, Yonatan and Dupoux, Emmanuel and Giulianelli, Mario and Hupkes, Dieuwke and Pinter, Yuval and Sajjad, Hassan},
	month = nov,
	year = {2021},
	pages = {337--349},
}

@inproceedings{asher_limits_2023,
	address = {Toronto, Canada},
	title = {Limits for learning with language models},
	url = {https://aclanthology.org/2023.starsem-1.22/},
	doi = {10.18653/v1/2023.starsem-1.22},
	urldate = {2025-08-10},
	booktitle = {Proceedings of the 12th {Joint} {Conference} on {Lexical} and {Computational} {Semantics} (*{SEM} 2023)},
	publisher = {Association for Computational Linguistics},
	author = {Asher, Nicholas and Bhar, Swarnadeep and Chaturvedi, Akshay and Hunter, Julie and Paul, Soumya},
	editor = {Palmer, Alexis and Camacho-collados, Jose},
	month = jul,
	year = {2023},
	pages = {236--248},
}

@inproceedings{ri_pretraining_2022,
	address = {Dublin, Ireland},
	title = {Pretraining with {Artificial} {Language}: {Studying} {Transferable} {Knowledge} in {Language} {Models}},
	shorttitle = {Pretraining with {Artificial} {Language}},
	url = {https://aclanthology.org/2022.acl-long.504/},
	doi = {10.18653/v1/2022.acl-long.504},
	urldate = {2025-08-10},
	booktitle = {Proceedings of the 60th {Annual} {Meeting} of the {Association} for {Computational} {Linguistics} ({Volume} 1: {Long} {Papers})},
	publisher = {Association for Computational Linguistics},
	author = {Ri, Ryokan and Tsuruoka, Yoshimasa},
	editor = {Muresan, Smaranda and Nakov, Preslav and Villavicencio, Aline},
	month = may,
	year = {2022},
	pages = {7302--7315},
}

@inproceedings{ravishankar_attention_2021,
	address = {Online},
	title = {Attention {Can} {Reflect} {Syntactic} {Structure} ({If} {You} {Let} {It})},
	url = {https://aclanthology.org/2021.eacl-main.264/},
	doi = {10.18653/v1/2021.eacl-main.264},
	urldate = {2025-08-10},
	booktitle = {Proceedings of the 16th {Conference} of the {European} {Chapter} of the {Association} for {Computational} {Linguistics}: {Main} {Volume}},
	publisher = {Association for Computational Linguistics},
	author = {Ravishankar, Vinit and Kulmizev, Artur and Abdou, Mostafa and Søgaard, Anders and Nivre, Joakim},
	editor = {Merlo, Paola and Tiedemann, Jorg and Tsarfaty, Reut},
	month = apr,
	year = {2021},
	pages = {3031--3045},
}

@inproceedings{thrush_investigating_2020,
	address = {Online},
	title = {Investigating {Novel} {Verb} {Learning} in {BERT}: {Selectional} {Preference} {Classes} and {Alternation}-{Based} {Syntactic} {Generalization}},
	shorttitle = {Investigating {Novel} {Verb} {Learning} in {BERT}},
	url = {https://aclanthology.org/2020.blackboxnlp-1.25/},
	doi = {10.18653/v1/2020.blackboxnlp-1.25},
	urldate = {2025-08-10},
	booktitle = {Proceedings of the {Third} {BlackboxNLP} {Workshop} on {Analyzing} and {Interpreting} {Neural} {Networks} for {NLP}},
	publisher = {Association for Computational Linguistics},
	author = {Thrush, Tristan and Wilcox, Ethan and Levy, Roger},
	editor = {Alishahi, Afra and Belinkov, Yonatan and Chrupała, Grzegorz and Hupkes, Dieuwke and Pinter, Yuval and Sajjad, Hassan},
	month = nov,
	year = {2020},
	pages = {265--275},
}

@inproceedings{warstadt_learning_2020,
	address = {Online},
	title = {Learning {Which} {Features} {Matter}: {RoBERTa} {Acquires} a {Preference} for {Linguistic} {Generalizations} ({Eventually})},
	shorttitle = {Learning {Which} {Features} {Matter}},
	url = {https://aclanthology.org/2020.emnlp-main.16/},
	doi = {10.18653/v1/2020.emnlp-main.16},
	urldate = {2025-08-10},
	booktitle = {Proceedings of the 2020 {Conference} on {Empirical} {Methods} in {Natural} {Language} {Processing} ({EMNLP})},
	publisher = {Association for Computational Linguistics},
	author = {Warstadt, Alex and Zhang, Yian and Li, Xiaocheng and Liu, Haokun and Bowman, Samuel R.},
	editor = {Webber, Bonnie and Cohn, Trevor and He, Yulan and Liu, Yang},
	month = nov,
	year = {2020},
	pages = {217--235},
}

@article{mccoy_embers_2023,
	author = {R. Thomas McCoy  and Shunyu Yao  and Dan Friedman  and Mathew D. Hardy  and Thomas L. Griffiths },
title = {Embers of autoregression show how large language models are shaped by the problem they are trained to solve},
journal = {Proceedings of the National Academy of Sciences},
volume = {121},
number = {41},
pages = {e2322420121},
year = {2024},
doi = {10.1073/pnas.2322420121},
URL = {https://www.pnas.org/doi/abs/10.1073/pnas.2322420121},
eprint = {https://www.pnas.org/doi/pdf/10.1073/pnas.2322420121},
}

@inproceedings{petersen_lexical_2023,
	address = {Dubrovnik, Croatia},
	title = {Lexical {Semantics} with {Large} {Language} {Models}: {A} {Case} {Study} of {English} “break”},
	shorttitle = {Lexical {Semantics} with {Large} {Language} {Models}},
	url = {https://aclanthology.org/2023.findings-eacl.36/},
	doi = {10.18653/v1/2023.findings-eacl.36},
	urldate = {2025-08-08},
	booktitle = {Findings of the {Association} for {Computational} {Linguistics}: {EACL} 2023},
	publisher = {Association for Computational Linguistics},
	author = {Petersen, Erika and Potts, Christopher},
	editor = {Vlachos, Andreas and Augenstein, Isabelle},
	month = may,
	year = {2023},
	pages = {490--511},
}

@inproceedings{li_neural_2022,
	address = {Dublin, Ireland},
	title = {Neural reality of argument structure constructions},
	url = {https://aclanthology.org/2022.acl-long.512/},
	doi = {10.18653/v1/2022.acl-long.512},
	urldate = {2025-08-08},
	booktitle = {Proceedings of the 60th {Annual} {Meeting} of the {Association} for {Computational} {Linguistics} ({Volume} 1: {Long} {Papers})},
	publisher = {Association for Computational Linguistics},
	author = {Li, Bai and Zhu, Zining and Thomas, Guillaume and Rudzicz, Frank and Xu, Yang},
	editor = {Muresan, Smaranda and Nakov, Preslav and Villavicencio, Aline},
	month = may,
	year = {2022},
	pages = {7410--7423},
}

@inproceedings{wang_cross-lingual_2019,
	address = {Hong Kong, China},
	title = {Cross-{Lingual} {BERT} {Transformation} for {Zero}-{Shot} {Dependency} {Parsing}},
	url = {https://aclanthology.org/D19-1575/},
	doi = {10.18653/v1/D19-1575},
	urldate = {2025-08-08},
	booktitle = {Proceedings of the 2019 {Conference} on {Empirical} {Methods} in {Natural} {Language} {Processing} and the 9th {International} {Joint} {Conference} on {Natural} {Language} {Processing} ({EMNLP}-{IJCNLP})},
	publisher = {Association for Computational Linguistics},
	author = {Wang, Yuxuan and Che, Wanxiang and Guo, Jiang and Liu, Yijia and Liu, Ting},
	editor = {Inui, Kentaro and Jiang, Jing and Ng, Vincent and Wan, Xiaojun},
	month = nov,
	year = {2019},
	pages = {5721--5727},
}

@inproceedings{vig_analyzing_2019,
	address = {Florence, Italy},
	title = {Analyzing the {Structure} of {Attention} in a {Transformer} {Language} {Model}},
	url = {https://aclanthology.org/W19-4808/},
	doi = {10.18653/v1/W19-4808},
	urldate = {2025-08-08},
	booktitle = {Proceedings of the 2019 {ACL} {Workshop} {BlackboxNLP}: {Analyzing} and {Interpreting} {Neural} {Networks} for {NLP}},
	publisher = {Association for Computational Linguistics},
	author = {Vig, Jesse and Belinkov, Yonatan},
	editor = {Linzen, Tal and Chrupała, Grzegorz and Belinkov, Yonatan and Hupkes, Dieuwke},
	month = aug,
	year = {2019},
	pages = {63--76},
}

@inproceedings{perez-mayos_evolution_2021,
	address = {Online},
	title = {On the evolution of syntactic information encoded by {BERT}'s contextualized representations},
	url = {https://aclanthology.org/2021.eacl-main.191/},
	doi = {10.18653/v1/2021.eacl-main.191},
	urldate = {2025-08-07},
	booktitle = {Proceedings of the 16th {Conference} of the {European} {Chapter} of the {Association} for {Computational} {Linguistics}: {Main} {Volume}},
	publisher = {Association for Computational Linguistics},
	author = {Pérez-Mayos, Laura and Carlini, Roberto and Ballesteros, Miguel and Wanner, Leo},
	editor = {Merlo, Paola and Tiedemann, Jorg and Tsarfaty, Reut},
	month = apr,
	year = {2021},
	pages = {2243--2258},
}

@inproceedings{mccoy_right_2019,
	address = {Florence, Italy},
	title = {Right for the {Wrong} {Reasons}: {Diagnosing} {Syntactic} {Heuristics} in {Natural} {Language} {Inference}},
	shorttitle = {Right for the {Wrong} {Reasons}},
	url = {https://aclanthology.org/P19-1334/},
	doi = {10.18653/v1/P19-1334},
	urldate = {2025-08-07},
	booktitle = {Proceedings of the 57th {Annual} {Meeting} of the {Association} for {Computational} {Linguistics}},
	publisher = {Association for Computational Linguistics},
	author = {McCoy, R. Thomas and Pavlick, Ellie and Linzen, Tal},
	editor = {Korhonen, Anna and Traum, David and Màrquez, Lluís},
	month = jul,
	year = {2019},
	pages = {3428--3448},
}

@article{wu_infusing_2021,
	title = {Infusing {Finetuning} with {Semantic} {Dependencies}},
	volume = {9},
	issn = {2307-387X},
	url = {https://direct.mit.edu/tacl/article/doi/10.1162/tacl_a_00363/98087/Infusing-Finetuning-with-Semantic-Dependencies},
	doi = {10.1162/tacl_a_00363},
	language = {en},
	urldate = {2025-08-07},
	journal = {Transactions of the Association for Computational Linguistics},
	author = {Wu, Zhaofeng and Peng, Hao and Smith, Noah A.},
	month = mar,
	year = {2021},
	pages = {226--242},
}

@inproceedings{clouatre_local_2022-1,
	address = {Dublin, Ireland},
	title = {Local {Structure} {Matters} {Most}: {Perturbation} {Study} in {NLU}},
	shorttitle = {Local {Structure} {Matters} {Most}},
	url = {https://aclanthology.org/2022.findings-acl.293/},
	doi = {10.18653/v1/2022.findings-acl.293},
	urldate = {2025-08-07},
	booktitle = {Findings of the {Association} for {Computational} {Linguistics}: {ACL} 2022},
	publisher = {Association for Computational Linguistics},
	author = {Clouatre, Louis and Parthasarathi, Prasanna and Zouaq, Amal and Chandar, Sarath},
	editor = {Muresan, Smaranda and Nakov, Preslav and Villavicencio, Aline},
	month = may,
	year = {2022},
	pages = {3712--3731},
}

@inproceedings{sinha_unnatural_2021,
	address = {Online},
	title = {{UnNatural} {Language} {Inference}},
	url = {https://aclanthology.org/2021.acl-long.569/},
	doi = {10.18653/v1/2021.acl-long.569},
	urldate = {2025-08-07},
	booktitle = {Proceedings of the 59th {Annual} {Meeting} of the {Association} for {Computational} {Linguistics} and the 11th {International} {Joint} {Conference} on {Natural} {Language} {Processing} ({Volume} 1: {Long} {Papers})},
	publisher = {Association for Computational Linguistics},
	author = {Sinha, Koustuv and Parthasarathi, Prasanna and Pineau, Joelle and Williams, Adina},
	editor = {Zong, Chengqing and Xia, Fei and Li, Wenjie and Navigli, Roberto},
	month = aug,
	year = {2021},
	pages = {7329--7346},
}

@inproceedings{wei-etal-2021-frequency,
    title = "Frequency Effects on Syntactic Rule Learning in Transformers",
    author = "Wei, Jason  and
      Garrette, Dan  and
      Linzen, Tal  and
      Pavlick, Ellie",
    editor = "Moens, Marie-Francine  and
      Huang, Xuanjing  and
      Specia, Lucia  and
      Yih, Scott Wen-tau",
    booktitle = "Proceedings of the 2021 Conference on Empirical Methods in Natural Language Processing",
    month = nov,
    year = "2021",
    address = "Online and Punta Cana, Dominican Republic",
    publisher = "Association for Computational Linguistics",
    url = "https://aclanthology.org/2021.emnlp-main.72/",
    doi = "10.18653/v1/2021.emnlp-main.72",
    pages = "932--948"
}

@inproceedings{wang-etal-2021-controlled,
    title = "Controlled Evaluation of Grammatical Knowledge in {M}andarin {C}hinese Language Models",
    author = "Wang, Yiwen  and
      Hu, Jennifer  and
      Levy, Roger  and
      Qian, Peng",
    editor = "Moens, Marie-Francine  and
      Huang, Xuanjing  and
      Specia, Lucia  and
      Yih, Scott Wen-tau",
    booktitle = "Proceedings of the 2021 Conference on Empirical Methods in Natural Language Processing",
    month = nov,
    year = "2021",
    address = "Online and Punta Cana, Dominican Republic",
    publisher = "Association for Computational Linguistics",
    url = "https://aclanthology.org/2021.emnlp-main.454/",
    doi = "10.18653/v1/2021.emnlp-main.454",
    pages = "5604--5620"
}

@inproceedings{pham_out_2021,
	address = {Online},
	title = {Out of {Order}: {How} important is the sequential order of words in a sentence in {Natural} {Language} {Understanding} tasks?},
	shorttitle = {Out of {Order}},
	url = {https://aclanthology.org/2021.findings-acl.98/},
	doi = {10.18653/v1/2021.findings-acl.98},
	urldate = {2025-08-07},
	booktitle = {Findings of the {Association} for {Computational} {Linguistics}: {ACL}-{IJCNLP} 2021},
	publisher = {Association for Computational Linguistics},
	author = {Pham, Thang and Bui, Trung and Mai, Long and Nguyen, Anh},
	editor = {Zong, Chengqing and Xia, Fei and Li, Wenjie and Navigli, Roberto},
	month = aug,
	year = {2021},
	pages = {1145--1160},
}

@inproceedings{gupta_bert_2021,
  title={BERT \& Family Eat Word Salad: Experiments with Text Understanding},
  author={Ashim Gupta and Giorgi Kvernadze and Vivek Srikumar},
  booktitle={AAAI Conference on Artificial Intelligence},
  year={2021},
  url={https://api.semanticscholar.org/CorpusID:231573267}
}

@article{manning_emergent_2020,
	title = {Emergent linguistic structure in artificial neural networks trained by self-supervision},
	volume = {117},
	issn = {0027-8424, 1091-6490},
	url = {https://pnas.org/doi/full/10.1073/pnas.1907367117},
	doi = {10.1073/pnas.1907367117},
	language = {en},
	number = {48},
	urldate = {2025-08-07},
	journal = {Proceedings of the National Academy of Sciences},
	author = {Manning, Christopher D. and Clark, Kevin and Hewitt, John and Khandelwal, Urvashi and Levy, Omer},
	month = dec,
	year = {2020},
	pages = {30046--30054},
}

@inproceedings{wang_superglue_2019,
	title = {{SuperGLUE}: {A} {Stickier} {Benchmark} for {General}-{Purpose} {Language} {Understanding} {Systems}},
	volume = {32},
	shorttitle = {{SuperGLUE}},
	url = {https://proceedings.neurips.cc/paper/2019/hash/4496bf24afe7fab6f046bf4923da8de6-Abstract.html},
	urldate = {2025-08-07},
	booktitle = {Advances in {Neural} {Information} {Processing} {Systems}},
	publisher = {Curran Associates, Inc.},
	author = {Wang, Alex and Pruksachatkun, Yada and Nangia, Nikita and Singh, Amanpreet and Michael, Julian and Hill, Felix and Levy, Omer and Bowman, Samuel},
	year = {2019},
}

@inproceedings{van_schijndel_quantity_2019,
	address = {Hong Kong, China},
	title = {Quantity doesn't buy quality syntax with neural language models},
	url = {https://aclanthology.org/D19-1592/},
	doi = {10.18653/v1/D19-1592},
	urldate = {2025-08-07},
	booktitle = {Proceedings of the 2019 {Conference} on {Empirical} {Methods} in {Natural} {Language} {Processing} and the 9th {International} {Joint} {Conference} on {Natural} {Language} {Processing} ({EMNLP}-{IJCNLP})},
	publisher = {Association for Computational Linguistics},
	author = {van Schijndel, Marten and Mueller, Aaron and Linzen, Tal},
	editor = {Inui, Kentaro and Jiang, Jing and Ng, Vincent and Wan, Xiaojun},
	month = nov,
	year = {2019},
	pages = {5831--5837},
}

@misc{bacon_does_2019,
	title = {Does {BERT} agree? {Evaluating} knowledge of structure dependence through agreement relations},
	shorttitle = {Does {BERT} agree?},
	url = {http://arxiv.org/abs/1908.09892},
	doi = {10.48550/arXiv.1908.09892},
	urldate = {2025-08-07},
	publisher = {arXiv},
	author = {Bacon, Geoff and Regier, Terry},
	month = aug,
	year = {2019},
	note = {arXiv:1908.09892},
}

@inproceedings{koyama_targeted_2025,
	address = {Vienna, Austria},
	title = {Targeted {Syntactic} {Evaluation} for {Grammatical} {Error} {Correction}},
	isbn = {979-8-89176-251-0},
	url = {https://aclanthology.org/2025.acl-long.1026/},
	doi = {10.18653/v1/2025.acl-long.1026},
	urldate = {2025-08-27},
	booktitle = {Proceedings of the 63rd {Annual} {Meeting} of the {Association} for {Computational} {Linguistics} ({Volume} 1: {Long} {Papers})},
	publisher = {Association for Computational Linguistics},
	author = {Koyama, Aomi and Mita, Masato and Yoon, Su-Youn and Takama, Yasufumi and Komachi, Mamoru},
	editor = {Che, Wanxiang and Nabende, Joyce and Shutova, Ekaterina and Pilehvar, Mohammad Taher},
	month = jul,
	year = {2025},
	pages = {21108--21125},
}

@inproceedings{kennedy_evidence_2025,
	address = {Vienna, Austria},
	title = {Evidence of {Generative} {Syntax} in {LLMs}},
	isbn = {979-8-89176-271-8},
	url = {https://aclanthology.org/2025.conll-1.25/},
	doi = {10.18653/v1/2025.conll-1.25},
	urldate = {2025-08-27},
	booktitle = {Proceedings of the 29th {Conference} on {Computational} {Natural} {Language} {Learning}},
	publisher = {Association for Computational Linguistics},
	author = {Kennedy, Mary},
	editor = {Boleda, Gemma and Roth, Michael},
	month = jul,
	year = {2025},
	pages = {377--396},
}

@inproceedings{ju_domain_2025,
	address = {Vienna, Austria},
	title = {Domain {Regeneration}: {How} well do {LLMs} match syntactic properties of text domains?},
	isbn = {979-8-89176-256-5},
	shorttitle = {Domain {Regeneration}},
	url = {https://aclanthology.org/2025.findings-acl.120/},
	doi = {10.18653/v1/2025.findings-acl.120},
	urldate = {2025-08-27},
	booktitle = {Findings of the {Association} for {Computational} {Linguistics}: {ACL} 2025},
	publisher = {Association for Computational Linguistics},
	author = {Ju, Da and Blix, Hagen and Williams, Adina},
	editor = {Che, Wanxiang and Nabende, Joyce and Shutova, Ekaterina and Pilehvar, Mohammad Taher},
	month = jul,
	year = {2025},
	pages = {2367--2388},
}

@inproceedings{vandermeerschen_supervised_2025,
	address = {Vienna, Austria},
	title = {Supervised and {Unsupervised} {Probing} of {Shortcut} {Learning}: {Case} {Study} on the {Emergence} and {Evolution} of {Syntactic} {Heuristics} in {BERT}},
	isbn = {979-8-89176-256-5},
	shorttitle = {Supervised and {Unsupervised} {Probing} of {Shortcut} {Learning}},
	url = {https://aclanthology.org/2025.findings-acl.499/},
	doi = {10.18653/v1/2025.findings-acl.499},
	urldate = {2025-08-27},
	booktitle = {Findings of the {Association} for {Computational} {Linguistics}: {ACL} 2025},
	publisher = {Association for Computational Linguistics},
	author = {Vandermeerschen, Elke and de Lhoneux, Miryam},
	editor = {Che, Wanxiang and Nabende, Joyce and Shutova, Ekaterina and Pilehvar, Mohammad Taher},
	month = jul,
	year = {2025},
	pages = {9592--9604},
}

@inproceedings{zhao_systematic_2025,
	address = {Vienna, Austria},
	title = {A {Systematic} {Study} of {Compositional} {Syntactic} {Transformer} {Language} {Models}},
	isbn = {979-8-89176-251-0},
	url = {https://aclanthology.org/2025.acl-long.350/},
	doi = {10.18653/v1/2025.acl-long.350},
	urldate = {2025-08-26},
	booktitle = {Proceedings of the 63rd {Annual} {Meeting} of the {Association} for {Computational} {Linguistics} ({Volume} 1: {Long} {Papers})},
	publisher = {Association for Computational Linguistics},
	author = {Zhao, Yida and Xve, Hao and Hu, Xiang and Tu, Kewei},
	editor = {Che, Wanxiang and Nabende, Joyce and Shutova, Ekaterina and Pilehvar, Mohammad Taher},
	month = jul,
	year = {2025},
	pages = {7070--7083},
}

@inproceedings{wijesiriwardene_relationship_2024,
    title = "On the Relationship between Sentence Analogy Identification and Sentence Structure Encoding in Large Language Models",
    author = "Wijesiriwardene, Thilini  and
      Wickramarachchi, Ruwan  and
      Reganti, Aishwarya Naresh  and
      Jain, Vinija  and
      Chadha, Aman  and
      Sheth, Amit  and
      Das, Amitava",
    editor = "Graham, Yvette  and
      Purver, Matthew",
    booktitle = "Findings of the Association for Computational Linguistics: EACL 2024",
    month = mar,
    year = "2024",
    address = "St. Julian{'}s, Malta",
    publisher = "Association for Computational Linguistics",
    url = "https://aclanthology.org/2024.findings-eacl.31/",
    pages = "451--457",
}

@inproceedings{ide_how_2025,
	address = {Albuquerque, New Mexico},
	title = {How to {Make} the {Most} of {LLMs}' {Grammatical} {Knowledge} for {Acceptability} {Judgments}},
	isbn = {979-8-89176-189-6},
	url = {https://aclanthology.org/2025.naacl-long.380/},
	doi = {10.18653/v1/2025.naacl-long.380},
	urldate = {2025-08-26},
	booktitle = {Proceedings of the 2025 {Conference} of the {Nations} of the {Americas} {Chapter} of the {Association} for {Computational} {Linguistics}: {Human} {Language} {Technologies} ({Volume} 1: {Long} {Papers})},
	publisher = {Association for Computational Linguistics},
	author = {Ide, Yusuke and Nishida, Yuto and Vasselli, Justin and Oba, Miyu and Sakai, Yusuke and Kamigaito, Hidetaka and Watanabe, Taro},
	editor = {Chiruzzo, Luis and Ritter, Alan and Wang, Lu},
	month = apr,
	year = {2025},
	pages = {7416--7432},
}

@misc{aljaafari_interpreting_2025,
	title = {Interpreting token compositionality in {LLMs}: {A} robustness analysis},
	shorttitle = {Interpreting token compositionality in {LLMs}},
	url = {http://arxiv.org/abs/2410.12924},
	doi = {10.48550/arXiv.2410.12924},
	urldate = {2025-08-26},
	publisher = {arXiv},
	author = {Aljaafari, Nura and Carvalho, Danilo S. and Freitas, André},
	month = may,
	year = {2025},
	note = {arXiv:2410.12924},
}

@article{n_atox_evaluating_nodate,
	title = {Evaluating {Large} {Language} {Models} through the {Lens} of {Linguistic} {Proficiency} and {World} {Knowledge}: {A} {Comparative} {Study}},
	doi = {10.22541/au.172479372.22580887/v1},
	number = {August 27, 2024},
	journal = {Authorea},
	author = {{N. Atox} and {M. Clark}},
}

@inproceedings{miaschi-etal-2020-linguistic,
    title = "Linguistic Profiling of a Neural Language Model",
    author = "Miaschi, Alessio  and
      Brunato, Dominique  and
      Dell{'}Orletta, Felice  and
      Venturi, Giulia",
    editor = "Scott, Donia  and
      Bel, Nuria  and
      Zong, Chengqing",
    booktitle = "Proceedings of the 28th International Conference on Computational Linguistics",
    month = dec,
    year = "2020",
    address = "Barcelona, Spain (Online)",
    publisher = "International Committee on Computational Linguistics",
    url = "https://aclanthology.org/2020.coling-main.65/",
    doi = "10.18653/v1/2020.coling-main.65",
    pages = "745--756"
}

@inproceedings{miaschi_evaluating_2024,
	address = {Miami, Florida, USA},
	title = {Evaluating {Large} {Language} {Models} via {Linguistic} {Profiling}},
	url = {https://aclanthology.org/2024.emnlp-main.166/},
	doi = {10.18653/v1/2024.emnlp-main.166},
	urldate = {2025-08-26},
	booktitle = {Proceedings of the 2024 {Conference} on {Empirical} {Methods} in {Natural} {Language} {Processing}},
	publisher = {Association for Computational Linguistics},
	author = {Miaschi, Alessio and Dell'Orletta, Felice and Venturi, Giulia},
	editor = {Al-Onaizan, Yaser and Bansal, Mohit and Chen, Yun-Nung},
	month = nov,
	year = {2024},
	pages = {2835--2848},
}

@inproceedings{xu_are_2023,
	address = {Singapore},
	title = {Are {Structural} {Concepts} {Universal} in {Transformer} {Language} {Models}? {Towards} {Interpretable} {Cross}-{Lingual} {Generalization}},
	shorttitle = {Are {Structural} {Concepts} {Universal} in {Transformer} {Language} {Models}?},
	url = {https://aclanthology.org/2023.findings-emnlp.931/},
	doi = {10.18653/v1/2023.findings-emnlp.931},
	urldate = {2025-08-26},
	booktitle = {Findings of the {Association} for {Computational} {Linguistics}: {EMNLP} 2023},
	publisher = {Association for Computational Linguistics},
	author = {Xu, Ningyu and Zhang, Qi and Ye, Jingting and Zhang, Menghan and Huang, Xuanjing},
	editor = {Bouamor, Houda and Pino, Juan and Bali, Kalika},
	month = dec,
	year = {2023},
	pages = {13951--13976},
}

@misc{bai_constituency_2023,
	title = {Constituency {Parsing} using {LLMs}},
	url = {http://arxiv.org/abs/2310.19462},
	doi = {10.48550/arXiv.2310.19462},
	urldate = {2025-08-23},
	publisher = {arXiv},
	author = {Bai, Xuefeng and Wu, Jialong and Chen, Yulong and Wang, Zhongqing and Zhang, Yue},
	month = oct,
	year = {2023},
	note = {arXiv:2310.19462},
}

@inproceedings{perez-mayos_how_2021,
	address = {Online and Punta Cana, Dominican Republic},
	title = {How much pretraining data do language models need to learn syntax?},
	url = {https://aclanthology.org/2021.emnlp-main.118/},
	doi = {10.18653/v1/2021.emnlp-main.118},
	urldate = {2025-08-23},
	booktitle = {Proceedings of the 2021 {Conference} on {Empirical} {Methods} in {Natural} {Language} {Processing}},
	publisher = {Association for Computational Linguistics},
	author = {Pérez-Mayos, Laura and Ballesteros, Miguel and Wanner, Leo},
	editor = {Moens, Marie-Francine and Huang, Xuanjing and Specia, Lucia and Yih, Scott Wen-tau},
	month = nov,
	year = {2021},
	pages = {1571--1582},
}

@inproceedings{limisiewicz_introducing_2021,
	address = {Online},
	title = {Introducing {Orthogonal} {Constraint} in {Structural} {Probes}},
	url = {https://aclanthology.org/2021.acl-long.36/},
	doi = {10.18653/v1/2021.acl-long.36},
	urldate = {2025-08-23},
	booktitle = {Proceedings of the 59th {Annual} {Meeting} of the {Association} for {Computational} {Linguistics} and the 11th {International} {Joint} {Conference} on {Natural} {Language} {Processing} ({Volume} 1: {Long} {Papers})},
	publisher = {Association for Computational Linguistics},
	author = {Limisiewicz, Tomasz and Mareček, David},
	editor = {Zong, Chengqing and Xia, Fei and Li, Wenjie and Navigli, Roberto},
	month = aug,
	year = {2021},
	pages = {428--442},
}

@article{arehalli_neural_2024,
	title = {Neural {Networks} as {Cognitive} {Models} of the {Processing} of {Syntactic} {Constraints}},
	volume = {8},
	issn = {2470-2986},
	doi = {10.1162/opmi_a_00137},
	language = {eng},
	journal = {Open Mind: Discoveries in Cognitive Science},
	author = {Arehalli, Suhas and Linzen, Tal},
	year = {2024},
	pmid = {38746852},
	pmcid = {PMC11093404},
	pages = {558--614},
}

@inproceedings{marks_geometry_2024,
	title = {The {Geometry} of {Truth}: {Emergent} {Linear} {Structure} in {Large} {Language} {Model} {Representations} of {True}/{False} {Datasets}},
	shorttitle = {The {Geometry} of {Truth}},
	url = {https://openreview.net/forum?id=aajyHYjjsk},
	language = {en},
	urldate = {2025-08-23},
	author = {Marks, Samuel and Tegmark, Max},
	month = aug,
	year = {2024},
}

@inproceedings{tayyar_madabushi_cxgbert_2020,
	address = {Barcelona, Spain (Online)},
	title = {{CxGBERT}: {BERT} meets {Construction} {Grammar}},
	shorttitle = {{CxGBERT}},
	url = {https://aclanthology.org/2020.coling-main.355/},
	doi = {10.18653/v1/2020.coling-main.355},
	urldate = {2025-08-23},
	booktitle = {Proceedings of the 28th {International} {Conference} on {Computational} {Linguistics}},
	publisher = {International Committee on Computational Linguistics},
	author = {Tayyar Madabushi, Harish and Romain, Laurence and Divjak, Dagmar and Milin, Petar},
	editor = {Scott, Donia and Bel, Nuria and Zong, Chengqing},
	month = dec,
	year = {2020},
	pages = {4020--4032},
}

@inproceedings{hale_llms_2024,
	address = {Miami, Florida, USA},
	title = {Do {LLMs} learn a true syntactic universal?},
	url = {https://aclanthology.org/2024.emnlp-main.950/},
	doi = {10.18653/v1/2024.emnlp-main.950},
	urldate = {2025-08-23},
	booktitle = {Proceedings of the 2024 {Conference} on {Empirical} {Methods} in {Natural} {Language} {Processing}},
	publisher = {Association for Computational Linguistics},
	author = {Hale, John T. and Stanojević, Miloš},
	editor = {Al-Onaizan, Yaser and Bansal, Mohit and Chen, Yun-Nung},
	month = nov,
	year = {2024},
	pages = {17106--17119},
}

@misc{zhou_how_2023,
	title = {How {Well} {Do} {Large} {Language} {Models} {Understand} {Syntax}? {An} {Evaluation} by {Asking} {Natural} {Language} {Questions}},
	shorttitle = {How {Well} {Do} {Large} {Language} {Models} {Understand} {Syntax}?},
	url = {http://arxiv.org/abs/2311.08287},
	doi = {10.48550/arXiv.2311.08287},
	urldate = {2025-08-23},
	publisher = {arXiv},
	author = {Zhou, Houquan and Hou, Yang and Li, Zhenghua and Wang, Xuebin and Wang, Zhefeng and Duan, Xinyu and Zhang, Min},
	month = nov,
	year = {2023},
	note = {arXiv:2311.08287},
}

@inproceedings{kunz_test_2021,
	address = {Punta Cana, Dominican Republic},
	title = {Test {Harder} than {You} {Train}: {Probing} with {Extrapolation} {Splits}},
	shorttitle = {Test {Harder} than {You} {Train}},
	url = {https://aclanthology.org/2021.blackboxnlp-1.2/},
	doi = {10.18653/v1/2021.blackboxnlp-1.2},
	urldate = {2025-08-23},
	booktitle = {Proceedings of the {Fourth} {BlackboxNLP} {Workshop} on {Analyzing} and {Interpreting} {Neural} {Networks} for {NLP}},
	publisher = {Association for Computational Linguistics},
	author = {Kunz, Jenny and Kuhlmann, Marco},
	editor = {Bastings, Jasmijn and Belinkov, Yonatan and Dupoux, Emmanuel and Giulianelli, Mario and Hupkes, Dieuwke and Pinter, Yuval and Sajjad, Hassan},
	month = nov,
	year = {2021},
	pages = {15--25},
}

@inproceedings{nikolaev_universe_2023,
	address = {Nancy, France},
	title = {The {Universe} of {Utterances} {According} to {BERT}},
	url = {https://aclanthology.org/2023.iwcs-1.12/},
	urldate = {2025-08-23},
	booktitle = {Proceedings of the 15th {International} {Conference} on {Computational} {Semantics}},
	publisher = {Association for Computational Linguistics},
	author = {Nikolaev, Dmitry and Padó, Sebastian},
	editor = {Amblard, Maxime and Breitholtz, Ellen},
	month = jun,
	year = {2023},
	pages = {99--105},
}

@inproceedings{kryvosheieva_controlled_2025,
	address = {Abu Dhabi, United Arab Emirates},
	title = {Controlled {Evaluation} of {Syntactic} {Knowledge} in {Multilingual} {Language} {Models}},
	url = {https://aclanthology.org/2025.loreslm-1.30/},
	urldate = {2025-08-22},
	booktitle = {Proceedings of the {First} {Workshop} on {Language} {Models} for {Low}-{Resource} {Languages}},
	publisher = {Association for Computational Linguistics},
	author = {Kryvosheieva, Daria and Levy, Roger},
	editor = {Hettiarachchi, Hansi and Ranasinghe, Tharindu and Rayson, Paul and Mitkov, Ruslan and Gaber, Mohamed and Premasiri, Damith and Tan, Fiona Anting and Uyangodage, Lasitha},
	month = jan,
	year = {2025},
	pages = {402--413},
}

@inproceedings{zhang_how_2023,
	address = {Singapore},
	title = {How {Well} {Do} {Text} {Embedding} {Models} {Understand} {Syntax}?},
	url = {https://aclanthology.org/2023.findings-emnlp.650/},
	doi = {10.18653/v1/2023.findings-emnlp.650},
	urldate = {2025-08-22},
	booktitle = {Findings of the {Association} for {Computational} {Linguistics}: {EMNLP} 2023},
	publisher = {Association for Computational Linguistics},
	author = {Zhang, Yan and Feng, Zhaopeng and Teng, Zhiyang and Liu, Zuozhu and Li, Haizhou},
	editor = {Bouamor, Houda and Pino, Juan and Bali, Kalika},
	month = dec,
	year = {2023},
	pages = {9717--9728},
}

@inproceedings{zhao_transformers_2023,
	address = {Singapore},
	title = {Do {Transformers} {Parse} while {Predicting} the {Masked} {Word}?},
	url = {https://aclanthology.org/2023.emnlp-main.1029/},
	doi = {10.18653/v1/2023.emnlp-main.1029},
	urldate = {2025-08-22},
	booktitle = {Proceedings of the 2023 {Conference} on {Empirical} {Methods} in {Natural} {Language} {Processing}},
	publisher = {Association for Computational Linguistics},
	author = {Zhao, Haoyu and Panigrahi, Abhishek and Ge, Rong and Arora, Sanjeev},
	editor = {Bouamor, Houda and Pino, Juan and Bali, Kalika},
	month = dec,
	year = {2023},
	pages = {16513--16542},
}

@inproceedings{zhang_unveiling_2024,
	address = {Bangkok, Thailand},
	title = {Unveiling {Linguistic} {Regions} in {Large} {Language} {Models}},
	url = {https://aclanthology.org/2024.acl-long.338/},
	doi = {10.18653/v1/2024.acl-long.338},
	urldate = {2025-08-22},
	booktitle = {Proceedings of the 62nd {Annual} {Meeting} of the {Association} for {Computational} {Linguistics} ({Volume} 1: {Long} {Papers})},
	publisher = {Association for Computational Linguistics},
	author = {Zhang, Zhihao and Zhao, Jun and Zhang, Qi and Gui, Tao and Huang, Xuanjing},
	editor = {Ku, Lun-Wei and Martins, Andre and Srikumar, Vivek},
	month = aug,
	year = {2024},
	pages = {6228--6247},
}

@article{zhang_dependency-based_2021,
	title = {Dependency-based syntax-aware word representations},
	volume = {292},
	issn = {0004-3702},
	url = {https://www.sciencedirect.com/science/article/pii/S0004370220301764},
	doi = {10.1016/j.artint.2020.103427},
	urldate = {2025-08-22},
	journal = {Artificial Intelligence},
	author = {Zhang, Meishan and Li, Zhenghua and Fu, Guohong and Zhang, Min},
	month = mar,
	year = {2021},
	pages = {103427},
}

@inproceedings{xu_syntax-enhanced_2021,
	address = {Online},
	title = {Syntax-{Enhanced} {Pre}-trained {Model}},
	url = {https://aclanthology.org/2021.acl-long.420/},
	doi = {10.18653/v1/2021.acl-long.420},
	urldate = {2025-08-21},
	booktitle = {Proceedings of the 59th {Annual} {Meeting} of the {Association} for {Computational} {Linguistics} and the 11th {International} {Joint} {Conference} on {Natural} {Language} {Processing} ({Volume} 1: {Long} {Papers})},
	publisher = {Association for Computational Linguistics},
	author = {Xu, Zenan and Guo, Daya and Tang, Duyu and Su, Qinliang and Shou, Linjun and Gong, Ming and Zhong, Wanjun and Quan, Xiaojun and Jiang, Daxin and Duan, Nan},
	editor = {Zong, Chengqing and Xia, Fei and Li, Wenjie and Navigli, Roberto},
	month = aug,
	year = {2021},
	pages = {5412--5422},
}

@inproceedings{wu_are_2020,
	address = {Online},
	title = {Are {All} {Languages} {Created} {Equal} in {Multilingual} {BERT}?},
	url = {https://aclanthology.org/2020.repl4nlp-1.16/},
	doi = {10.18653/v1/2020.repl4nlp-1.16},
	urldate = {2025-08-21},
	booktitle = {Proceedings of the 5th {Workshop} on {Representation} {Learning} for {NLP}},
	publisher = {Association for Computational Linguistics},
	author = {Wu, Shijie and Dredze, Mark},
	editor = {Gella, Spandana and Welbl, Johannes and Rei, Marek and Petroni, Fabio and Lewis, Patrick and Strubell, Emma and Seo, Minjoon and Hajishirzi, Hannaneh},
	month = jul,
	year = {2020},
	pages = {120--130},
}

@inproceedings{wu_beto_2019,
	address = {Hong Kong, China},
	title = {Beto, {Bentz}, {Becas}: {The} {Surprising} {Cross}-{Lingual} {Effectiveness} of {BERT}},
	shorttitle = {Beto, {Bentz}, {Becas}},
	url = {https://aclanthology.org/D19-1077/},
	doi = {10.18653/v1/D19-1077},
	urldate = {2025-08-21},
	booktitle = {Proceedings of the 2019 {Conference} on {Empirical} {Methods} in {Natural} {Language} {Processing} and the 9th {International} {Joint} {Conference} on {Natural} {Language} {Processing} ({EMNLP}-{IJCNLP})},
	publisher = {Association for Computational Linguistics},
	author = {Wu, Shijie and Dredze, Mark},
	editor = {Inui, Kentaro and Jiang, Jing and Ng, Vincent and Wan, Xiaojun},
	month = nov,
	year = {2019},
	pages = {833--844},
}

@misc{tian_large_2025,
	title = {Large {Language} {Models} {Enhanced} by {Plug} and {Play} {Syntactic} {Knowledge} for {Aspect}-based {Sentiment} {Analysis}},
	url = {http://arxiv.org/abs/2506.12991},
	doi = {10.48550/arXiv.2506.12991},
	urldate = {2025-08-21},
	publisher = {arXiv},
	author = {Tian, Yuanhe and Li, Xu and Wang, Wei and Jin, Guoqing and Cheng, Pengsen and Song, Yan},
	month = jun,
	year = {2025},
	note = {arXiv:2506.12991},
}


%% file: nora-import.bib
@preamble{ " \newcommand{\noop}[1]{} " }
